\documentclass{article} % For LaTeX2e
\usepackage{iclr2027_conference,times}

\usepackage{amsmath,amsfonts,bm}

\def\eqref#1{equation~\ref{#1}}
\def\1{\bm{1}}

\DeclareMathAlphabet{\mathsfit}{\encodingdefault}{\sfdefault}{m}{sl}
\SetMathAlphabet{\mathsfit}{bold}{\encodingdefault}{\sfdefault}{bx}{n}

\newcommand{\abl}[1]{\ensuremath{-\mathrm{#1}}}
\newcommand{\appref}[1]{Appendix~\ref{#1}}
\newcommand{\cmark}{\ensuremath{\checkmark}}

\usepackage{listings}
\definecolor{promptbg}{gray}{0.95}
\lstdefinestyle{promptstyle}{
  basicstyle=\ttfamily\scriptsize,
  backgroundcolor=\color{promptbg},
  xleftmargin=6pt, xrightmargin=6pt,
  framexleftmargin=4pt, framexrightmargin=4pt,
  framextopmargin=3pt, framexbottommargin=3pt,
  columns=fullflexible, keepspaces=true,
  breaklines=false,
}

\usepackage{longtable}
\usepackage{algorithm}
\usepackage{algorithmic}
\usepackage{float}
\usepackage{hyperref}
\hypersetup{
  pdftitle={OpenKAGE: Open-Ended Lifelong Learning in the Open World via Knowledge-Sharing Agent-Group Evolution},
  pdfauthor={Wenhao Yuan, Longhui Yuan, Zexi Jia, Ningzhe Shi, Jingwei Li, Shipeng Cen, Pengcheng Luo, Jie Zhou, Jinchao Zhang},
  pdfsubject={Open-ended lifelong learning in Minecraft},
  pdfkeywords={open-ended learning, lifelong learning, LLM agents, knowledge sharing, Minecraft}
}
\usepackage{url}
\usepackage{tikz}
\usepackage{graphicx}
\graphicspath{{./}{fig/}}
\usepackage{booktabs}
\usepackage{array}
\usetikzlibrary{arrows.meta, positioning, fit, backgrounds, calc}
\renewcommand{\floatpagefraction}{.85}
\title{\SYSNAME: Open-Ended Lifelong Learning in the Open World via Knowledge-Sharing Agent-Group Evolution}

\author{Wenhao Yuan \And Longhui Yuan \And Zexi Jia \AND
Ningzhe Shi \And Jingwei Li \And Shipeng Cen \AND
Pengcheng Luo \And Jie Zhou \And Jinchao Zhang}

\iclrfinalcopy % arXiv build: disables the review ruler; headers are cleared in the .sty
\begin{document}

% System name. OpenKAGE = Open Knowledge-sharing Agent-Group Evolution.
% "Open" 一词三义：开放世界 / 开放式学习 / 知识对全群开放（vs 锁死在单一智能体）。
% 改名只改这一处，全文 \SYSNAME 自动替换。
\newcommand{\SYSNAME}{OpenKAGE}

\maketitle

\begin{abstract}
Building autonomous agents that continually explore open worlds and acquire
new capabilities through experience is a long-standing aspiration in
artificial intelligence. Existing research often pursues lifelong learning
within predefined tasks, curricula, or objectives, with self-improvement
largely confined to individual agents. We introduce \SYSNAME{}, an
agent-group evolution framework for open-ended lifelong learning. Within
this framework, agents autonomously explore open worlds and continually
acquire new capabilities by accumulating and reusing knowledge from their
own interactions and the experience of others. \SYSNAME{} builds on two
key mechanisms: (1)~a two-layer knowledge representation, with a memory
layer for situated memories and an experience layer for transferable
experience and skills that retains the applicability conditions needed
for reuse; and (2)~a knowledge-sharing mechanism that semantically
consolidates textual experience into a continuously evolving shared
knowledge graph, while executable skills are shared verbatim. In
Minecraft, \SYSNAME{} supports open-ended lifelong learning, enabling
agent groups to sustain exploration and reach advanced technological
milestones. Furthermore, the accumulated knowledge can be transferred to
a newly initialized agent group on a different backbone model, improving
its survival and exploration progress. Code and data are available at
\url{https://github.com/zhuzhuxia5678/OpenKAGE} and
\url{https://github.com/zhuzhuxia5678/OpenKAGE_results}.
\end{abstract}

\section{Introduction}
\label{sec:introduction}

We study \emph{open-ended lifelong learning} as a continuing process in
which agents autonomously generate goals, accumulate knowledge from
interaction, and reuse that knowledge to acquire new capabilities. Unlike
learning toward a fixed terminal task, this process is not organized
around a predefined endpoint. A suggestive image comes from \emph{The Wild
Robot}: cast ashore on an unfamiliar island with only her built-in
directives, Roz learns to coexist with the wildlife and, by raising an
orphaned gosling, comes to belong to the island. Future autonomous agents
may likewise enter physical or digital worlds carrying only system-level
directives, and must confront unfamiliar phenomena, discover new things to
do, and devise new strategies without human intervention.

Toward this vision, substantial progress has been made in enabling agents
to learn and act in open worlds. Reinforcement and imitation learning
support long-horizon control \citep{baker2022vpt,hafner2023dreamerv3}, while
LLM-based agents combine planning, memory, and reflection to tackle
increasingly complex tasks
\citep{zhu2023gitm,wang2023jarvis1,shinn2023reflexion}.
Systems such as Voyager, Optimus-1, and AriGraph further demonstrate how
reusable skills, structured experience, and knowledge graphs can support
knowledge accumulation and reuse
\citep{wang2023voyager,li2024optimus1,anokhin2024arigraph}.
Despite these advances, much of the research remains organized around
predefined tasks, curricula, or objectives, with self-improvement largely
confined to individual agents. Although systems such as Voyager support
autonomous exploration, an individual agent still learns primarily from
its own interactions, limiting the range of experience available to guide
its development. Open-endedness research does generate new challenges rather
than fixing them, by co-evolving environments with their solvers or by asking
a foundation model which tasks are interesting
\citep{wang2019poet,zhang2023omni}, but this concerns what a single learner
pursues next. Knowledge acquired by one agent does not automatically
become available to others. This motivates moving beyond individual
learning toward agent-group evolution, where experience acquired through
diverse interactions can contribute to a shared, growing body of knowledge.

% ═══ Figure 1: motivation (fig1.svg, generated) ═══
\begin{figure}[htbp]
\centering
\includegraphics[width=\linewidth]{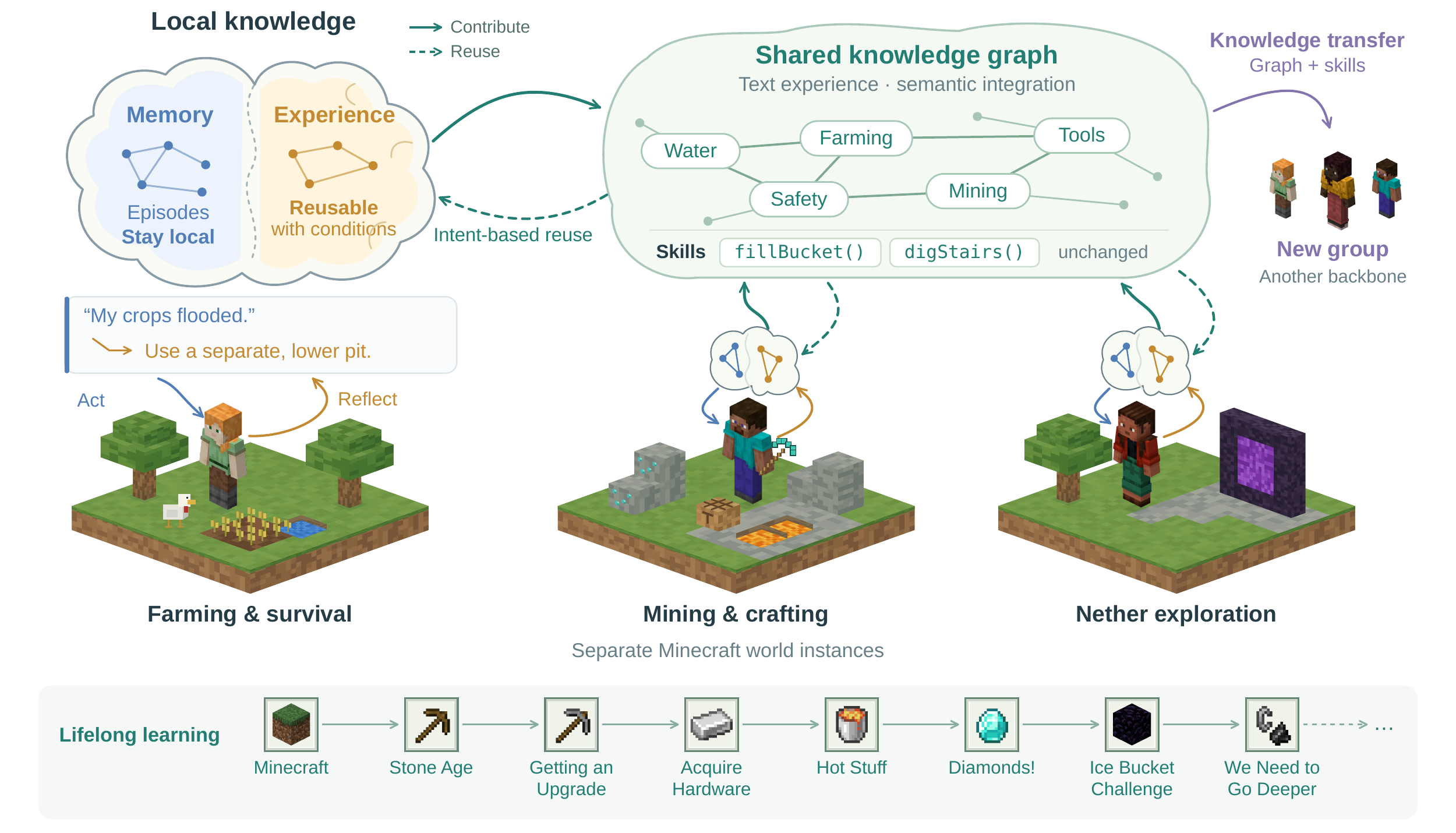}
\caption{Open-ended lifelong learning through shared knowledge in \SYSNAME{}. Agents
explore separate Minecraft worlds and keep situated memories local; transferable experience,
with its applicability conditions, is consolidated into a shared knowledge graph, and
executable skills are shared verbatim. The bottom strip illustrates milestones, not a
prescribed curriculum.}
\label{fig:motivation}
\end{figure}

Learning from others offers a natural route from individual learning to collective improvement.
CAMEL and MetaGPT enable agents to collaborate through role-based communication and structured workflows
\citep{li2023camel,hong2023metagpt}, while Generative Agents combines memory, reflection, and social interaction
to support information diffusion and coordination \citep{park2023generative}.
More directly, Group-Evolving Agents (GEA) shares evolutionary traces across agents and turns them into
framework-level improvements through reflection \citep{weng2026gea}.
Memory systems such as G-Memory and LLMA-Mem keep shared memory across LLM agents
\citep{zhang2025gmemory,wu2026llmamem}, and selective-sharing methods pass only chosen
experiences or policy modules between agents \citep{gerstgrasser2023seps,nath2026policy}.
These approaches demonstrate different ways for agents to benefit from one another, from coordinating
actions to improving their own implementations or sharing memory and experience, though most are
evaluated with tasks given in advance.
For open-ended lifelong learning, we focus on a complementary challenge: how to turn knowledge acquired
through individual interactions into a shared resource that supports continued exploration and capability
acquisition across the group.
This calls for distinguishing memories tied to an agent's particular circumstances from experience that
can be reused by others, and continually consolidating the latter as agents explore and learn.

To this end, we introduce \SYSNAME{}, an agent-group evolution framework for open-ended lifelong learning
(Figure~\ref{fig:motivation}).
\SYSNAME{} enables agents to autonomously explore open worlds and continually acquire capabilities by
accumulating and reusing knowledge from their own interactions and the experience of others.
Two mechanisms underpin this process.
First, a two-layer knowledge representation separates private situated memories from transferable
experience and skills, retaining the applicability conditions needed for reuse across agents.
Second, a knowledge-sharing mechanism semantically consolidates textual experience into a continuously
evolving shared knowledge graph, while executable skills are shared verbatim and remain maintained by
their authors.
Situated memories remain private, while transferable knowledge becomes a cumulative resource for the group
rather than remaining isolated within individual agents.
To bring this accumulated knowledge into subsequent decisions, intent-aware subgraph retrieval draws
relevant knowledge from both the agent's local graph and the shared graph to support its ongoing exploration.

We evaluate \SYSNAME{} in Minecraft, where agents must survive, explore, and develop increasingly
advanced capabilities through sustained interaction with an open world.
Starting with empty knowledge stores, agents accumulate and share knowledge acquired through
their own exploration.
Across four backbone models, \SYSNAME{} supports sustained exploration and progression to advanced
technological milestones, obtaining obsidian on all four and entering the Nether on three, while
reducing deaths relative to the planner-only control; on the default backbone it also outperforms
Voyager and an Echo baseline, neither of which reaches obsidian or the Nether within the
evaluated horizon.
Furthermore, knowledge accumulated by one group can be transferred to a newly initialized group using
a different backbone model, substantially improving its survival and exploration progress.
Together, these results show that knowledge accumulated through individual interactions supports
continued learning within a group and capability acquisition by new agents.

\section{\SYSNAME}
\label{sec:method}

% ═══ Figure 2: system overview (fig2.svg, generated) ═══
\begin{figure}[htbp]
\centering
\includegraphics[width=\linewidth]{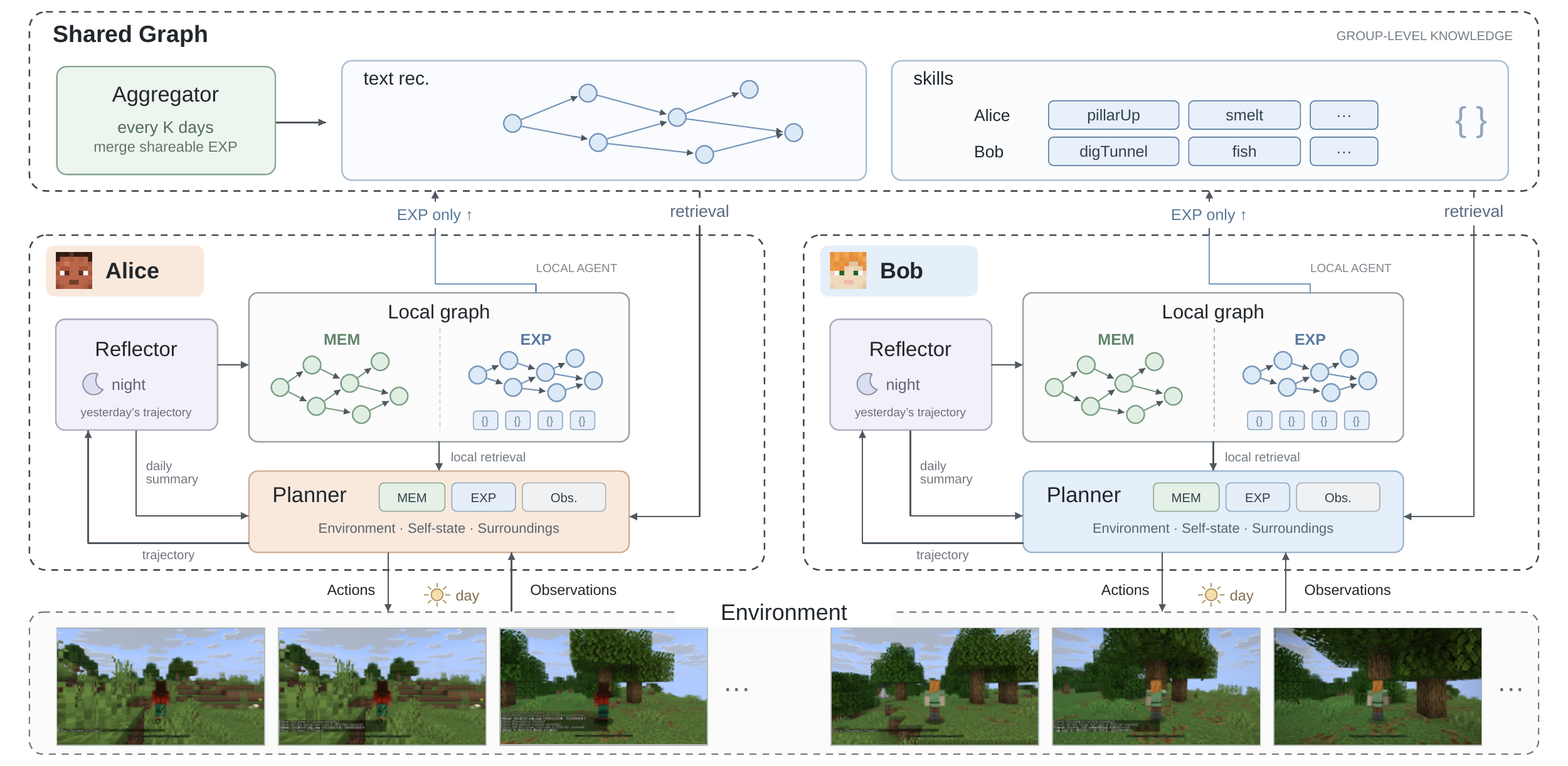}
\caption{\SYSNAME{} overview. Each agent's local graph has a private memory layer and a
shareable experience layer; only experience-layer records leave it. Textual records are
consolidated into the shared graph every $K$ days, and skills enter it verbatim under their
authors' names. An intent-aware retriever serves the planner. Schematic; agents receive
text observations, not images.}
\label{fig:framework}
\end{figure}

\definecolor{kgeagentblue}{HTML}{4477AA}
\definecolor{kgeagentgreen}{HTML}{44AA77}
\definecolor{kgeagentpurple}{HTML}{8844AA}
\definecolor{kgeagentolive}{HTML}{777733}
\definecolor{kgeagentmaroon}{HTML}{AA4455}
\definecolor{kgefunnel}{HTML}{F5F0DC}
\definecolor{kgefunneledge}{HTML}{6B6B2F}
\definecolor{kgeshared}{HTML}{E8F5EE}
\definecolor{kgesharededge}{HTML}{2E7D54}
\definecolor{kgeretrieve}{HTML}{992233}
\definecolor{kgeskill}{HTML}{C77F2E}
\definecolor{soft}{HTML}{6B6B6B}
\definecolor{ink}{HTML}{1A1A1A}
% GraphRAG-style retrieval figure (fig:retrieval) palette
\definecolor{qedge}{HTML}{4477AA}
\definecolor{entryred}{HTML}{C94F5A}
\definecolor{hopgray}{HTML}{9AA3AD}
\definecolor{takenblue}{HTML}{5B9BD5}
\definecolor{outgreen}{HTML}{4CAF7D}
\definecolor{pcblue}{HTML}{2E6DA4}
\definecolor{amber}{HTML}{E0A458}

\subsection{Overview}
\label{sec:agents}

\SYSNAME{} enables a group of agents to accumulate and reuse knowledge
from their individual interactions with an open world
(Figure~\ref{fig:framework}). Each agent interacts with a separate
Minecraft world instance, and agents exchange knowledge rather than
in-world messages or resources. Each agent runs a ReAct loop organized
into game-day cycles \citep{yao2023react}, using a \emph{planner} to act
and a \emph{reflector} to separate situated memories from transferable
experience; an \emph{aggregator} merges experience-layer knowledge
into the shared graph, semantically consolidating textual experience
records and adding executable skills verbatim as isolated nodes; an
intent-aware \emph{retriever} supplies relevant subgraphs to the planner. This
subsection describes how these pieces operate over time.

\paragraph{Planner.}
During each game-day cycle, the agent alternates between thinking,
acting, and observing. Each morning it sets a plan for the day (a goal and
ordered steps); planning with a goal triggers retrieval from the agent's
own graph and the shared graph, so the day's work is informed by
previously accumulated group knowledge. The agent then acts by writing
short asynchronous JavaScript programs, one program per step, over the
environment's low-level native Mineflayer API (pathfind-and-move, dig,
place, craft, equip, \ldots). The API provides no task-level routines for gathering,
smelting, or shelter construction; packaging them into skills is precisely
the agents' job. Executed programs return structured observations
(position, health, inventory, nearby entities and blocks, time,
advancements). Following \citet{wang2023voyager}, code is the action space:
programs are temporally extended, compositional, and directly executable.

% ═══ 图3（日循环 fig3.svg）：左 panel 观察/反馈示例 + 右 panel 日循环时序 ═══
\begin{figure}[htbp]
\centering
\includegraphics[width=\textwidth]{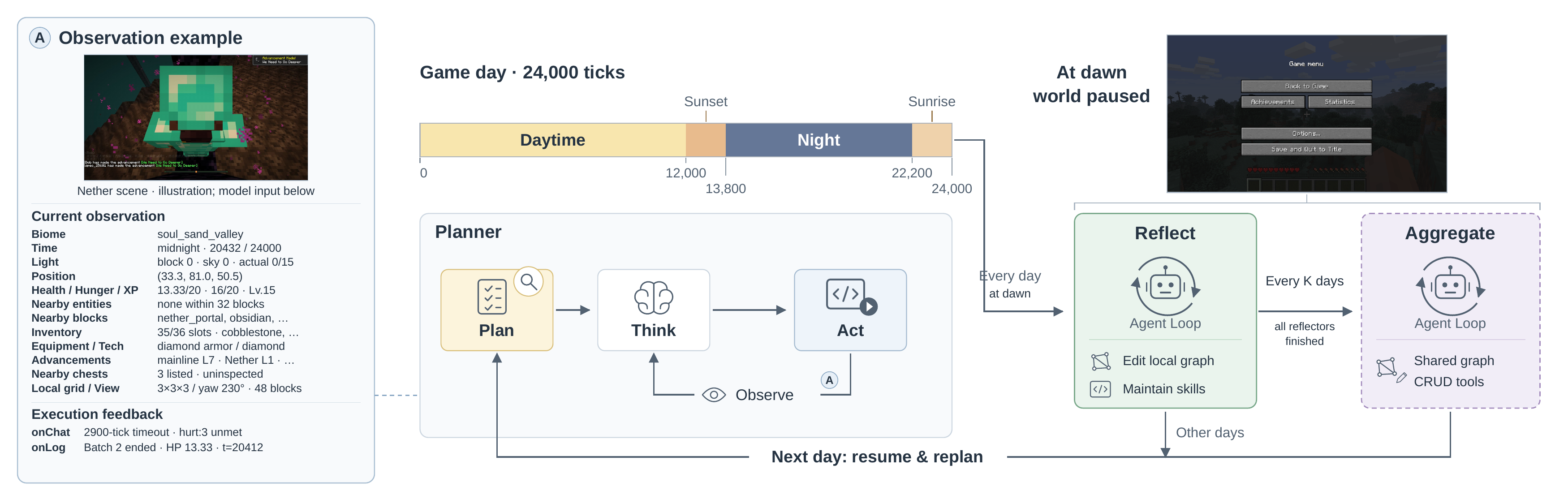}
\caption{Operation in game-day cycles. \textbf{Left:} structured observation and
execution feedback (schematic). \textbf{Right:} within each 24{,}000-tick day the planner
alternates between thinking, acting, and observing; at dawn the world pauses for
reflection, and aggregation runs every $K$ days.}
\label{fig:day_cycle}
\end{figure}

\paragraph{Reflector.}
At the end of each game-day cycle (at dawn), the world pauses and the
agent reflects on the elapsed day in two phases. The first re-reads the
day's trajectory and writes both layers of its local graph. It extends the
timeline, updates locations and possessions, and proposes
decontextualized experience-layer concepts supported by the day's events.
The second updates its skills based on per-skill success and failure
counts. The reflector also emits a one-line factual summary of the day's
progress. Apart from the knowledge stores, this summary and the previous
day's goal are the only free-form text carried from one day to the next;
everything else lives in the graphs.

\paragraph{Aggregator.}
Skills written or revised in a dawn reflection enter the shared graph
immediately; every $K$ days, after all reflectors finish, the textual
experience records are consolidated as well, so the agents enter the next
$K$-day aggregation window with the shared graph already updated. The
loop of act, reflect, aggregate, and retrieve continues for the whole run:
competence that any single agent acquires is consolidated where every agent
can find it, and the group's knowledge compounds. No step of the loop
involves human feedback, a human-authored curriculum, or parameter
updates.

\subsection{Two-layer knowledge graphs}
\label{sec:two-layer}

\SYSNAME{} represents world knowledge as
directed labeled graphs $G=(V,E)$: nodes $v\in V$ are \emph{concepts},
each carrying typed \emph{facets} (short textual statements describing one
aspect of the concept), and edges $E$ assert typed relations $u \xrightarrow{r} v$. Each
agent $i$ owns a \emph{local graph} $G_i$, maintained by its reflector and
consisting of a private memory layer and a shareable experience layer; the group maintains one
\emph{shared graph} $G^{\ast}$ that consolidates the agents' experience
over the run.

The two layers follow the complementary
learning systems account of memory \citep{kumaran2016cls}: $V_i = V_i^{\mathrm{mem}} \cup
V_i^{\mathrm{exp}}$, with each concept carrying its layer
$\ell \in \{\mathrm{mem}, \mathrm{exp}\}$:

\begin{itemize}
    \item \textbf{Memory layer.} Situated, first-person records of what
    happened to \emph{this} agent: a dated, agent-bound episode, illustrated
    with a concrete concept identifier and its episode text in
    \appref{sec:app_schema} (Figure~\ref{fig:twolayer}). Nodes are bound
    to the agent's coordinate frame and timeline, forming a day-chained
    backbone with locations, possessions, and events attached.
    \item \textbf{Experience layer.} Transferable knowledge
    abstracted from the agent's interactions. Records in this layer abstract
    away agent-specific locations, timestamps, and identities while
    preserving task-relevant preconditions and applicability conditions.
    The layer contains two forms of transferable knowledge:
    \emph{textual experience records}, which state a reusable procedure or
    failure condition, and \emph{executable skills}, which are parameterized
    programs produced by the reflector and invoked by the planner. Nodes
    use agent-independent concept names, so they are useful to any agent
    facing a similar situation. Concrete record texts, skill names and
    concept identifiers are given in \appref{sec:app_schema}.
\end{itemize}

% ═══ 图4（两层私有图，五个真实画面时刻）已移至附录 A.2（fig:twolayer，
% 紧跟 Layers 段，单栏下用 0.9\textwidth；正文此处仅保留引用）═══

The two forms influence behavior through different channels. Textual
experience records are retrieved into context and inform planning;
executable skills are injected into the execution environment and
implement behavior directly.

The separation is enforced at the level of access: the reflector declares
the layer of every concept, and the aggregator is given only the
experience layers as input, $V^{\mathrm{in}} \subseteq \bigcup_{i}
V_i^{\mathrm{exp}}$, so memory-layer records are not provided to the
aggregator; the reflector is instructed to omit
agent-specific coordinates and timestamps from shareable records. The
two layers have different access channels: the experience layer is
consulted \emph{by intent} (goal-conditioned retrieval), while the memory
layer is \emph{injected} each morning as a bounded slice of persistent
context, with per-category budgets (episodic, spatial, lesson, status;
\appref{sec:app_schema}) so
that high-volume daily records cannot displace slowly changing spatial
records (camps, chests, mines) that may not be retrieved by
goal-directed queries.

\subsection{Knowledge sharing}
\label{sec:sharing}

Sharing operates on both forms of knowledge in the experience layer, and
the two call for different treatment. \emph{Textual experience records}
reach an agent only through its context, so the group is free to rewrite
them into whatever consolidated wording serves best: they are semantically
consolidated into the shared graph every $K$ days. \emph{Executable skills}
are executed rather than read, so their code must not be rewritten: each
skill enters the shared graph verbatim as an isolated node, without
semantic consolidation, in the same dawn reflection in which it is written
or revised.

\paragraph{Textual-experience consolidation.}
Every $K$ days, the aggregator reads the agents' experience-layer
text and updates the shared graph (\appref{sec:app_sharing}). It consolidates
textual experience records rather than merely appending them: \textbf{merging
synonyms}, where several agents' wordings of one insight become a single
concept: a vector-similarity pre-screen nominates candidate duplicates,
and names already in use are reused, so a concept named once keeps its
name; \textbf{recording cross-agent corroboration}, where a facet states
explicitly when experience is confirmed by multiple agents, an
unconfirmed claim still being admitted to await corroboration;
\textbf{reconciling contradictions}, where conflicts are first tested for
differing applicability conditions (one observation holds by day, the
other fails at night) and corrected in place when one side is simply
wrong; and \textbf{aligning with prior knowledge}, where new experience
is folded into existing concepts wherever they match.

Let $V_{i,\mathrm{text}}^{\mathrm{exp}} \subseteq V_i^{\mathrm{exp}}$
denote the textual nodes in agent $i$'s experience layer. After a
successful aggregation, these local textual records are
cleared, $V_{i,\mathrm{text}}^{\mathrm{exp}} \leftarrow \emptyset$, so each
agent's shareable textual output stays bounded at $K$ days of reflection
while the shared graph grows across windows.

\paragraph{Skill sharing.}
Skills bypass semantic consolidation. A skill node in the shared
graph carries its author's code unchanged, together with its author and
its cumulative success and failure record; it has no edges, so retrieval
reaches it by matching the goal against the skill's stated intent (a
one-line description of when to use it), as in vector
anchoring, rather than by relational expansion. A skill is
propagated \emph{verbatim} and selected by its record; the author alone
revises or retires it, so what circulates in the group is exactly the set
of skills their authors still stand behind. The skill nodes are implemented
as a read-only registry over the agents' own skill libraries, so code is
neither copied nor rewritten in transit
(\appref{sec:app_skills}). For skills, code delivery and
context presentation use different channels: the code is injected
wholesale into the execution environment, while only the signatures of
the top-$n_{\mathrm{skill}}{=}20$ skills with the strongest execution records
occupy the context, supplemented by
signatures retrieved for the current goal.

\subsection{Intent-aware subgraph retrieval}
\label{sec:retrieval}

The \emph{retriever} is the component bringing the graphs back to
agents: it is triggered by intent, i.e., the agent's current goal or a
query it issues, and during the day the agent may query
or expand at will. Unlike skill-library or RAG memory
\citep{wang2023voyager,lewis2020rag}, it exploits graph structure for
intent-aware retrieval, in three stages.

% ═══ Figure 4: intent-aware subgraph retrieval (Sec 2.3) = fig6.svg ═══
% Generated replacement of the hand-built TikZ version (still in git history;
% comment block preserved below). Schematic example, not experimental data:
% LEFT = query box + entry column (hop 0 anchors); RIGHT = hop1 / hop2 node
% columns. Edge language: blue solid = the sampled two-hop path, gray dashed
% = the other candidate relations at each frontier. Scoring per hop = joint
% normalization over the frontier, top-p pruning, renormalization, then
% independent sampling. Prompt spec: fig/method_figures_prompts.md.
\begin{figure}[htbp]
\centering
\includegraphics[width=\textwidth]{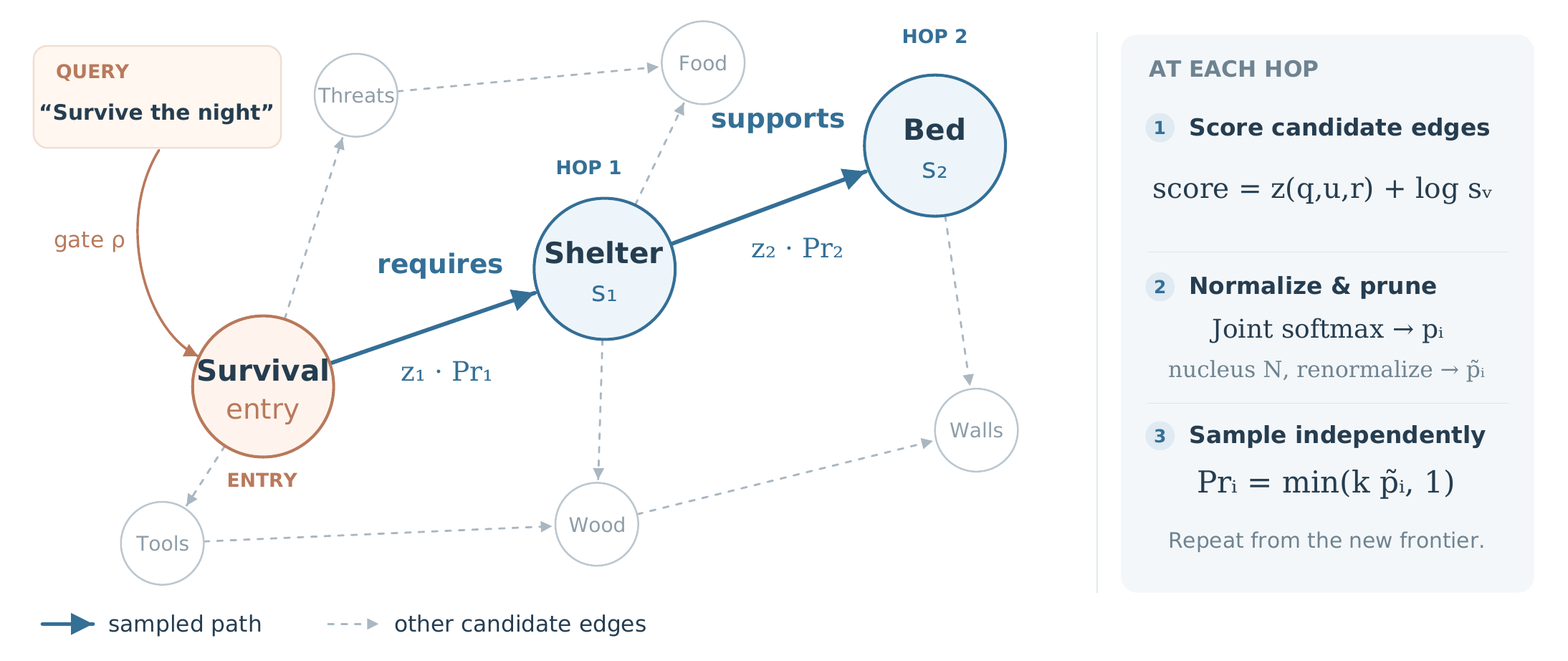}
\caption{Intent-aware subgraph retrieval (schematic). The retriever selects
anchors by relative similarity, then expands along relations scored by
query-conditioned relevance and retrieval-use strength; blue solid arrows
trace one sampled path, dashed arrows the other candidates. A structured
subgraph is returned, not a ranked list.}
\label{fig:retrieval}
\end{figure}

\paragraph{Vector anchoring.}
The query is embedded and matched against concept embeddings; among the
top-$k_{\mathrm{anc}}$ candidates, an anchor is admitted only if its
similarity is within a ratio $\rho$ of the best match, i.e., its cosine
similarity is at least $\rho$ times the best candidate's. The gate is
relative because absolute cosine similarities are not comparable across
embedding models and corpora. The number of anchors is set
by the query's semantics, not by a fixed count or an absolute threshold.

\paragraph{Relational expansion.}
From the anchors, retrieval walks the graph's edges outward for up to $h$
hops. For a frontier concept $u$ with an edge of type $r$ to a neighbor
$v$, the edge score is
\begin{equation}
    \mathrm{score}(q, u \xrightarrow{r} v) = z_{q,u,r} + \log s_v,
    \qquad
    z_{q,u,r} = \frac{\cos\!\big(e_q,\, e_{u,r}\big) - \mu_u}{\sigma_u},
    \label{eq:edge-score}
\end{equation}
where $e_q$ is the query embedding, $e_{u,r}$ is the embedding of the
relation's text prefixed with $u$'s name, and $\mu_u, \sigma_u$ are the
mean and standard deviation of raw cosine similarities across the
relations at $u$ (when the standard deviation is zero, as at a node with a
single relation, the query-relevance terms $z_{q,u,r}$ are set to zero).
The cosine-similarity standardization is performed within each node
rather than globally. Here $s_v$ is the neighbor's retrieval-use strength, a
monotone counter of how often retrieval has used $v$; it is initialized
at $1$ and updated only through retrieval
(\appref{sec:app_retrieval}).

The unnormalized sampling weight of an edge is thus proportional to
$\exp(z_{q,u,r}/\tau)\, s_v^{1/\tau}$, combining query relevance with
retrieval-use strength. The candidate edges of a hop,
$\mathcal{C}$, are softmax-normalized jointly at temperature $\tau$, so
that an anchor with many good edges is not outvoted by a one-edge
anchor, then pruned to a nucleus $\mathcal{N} \subseteq \mathcal{C}$ of
cumulative mass $p_{\mathrm{nuc}}$ and renormalized; each surviving edge is then taken
independently with probability
\begin{equation}
  \Pr[\text{follow } i] = \min\big(k\,\tilde{p}_i,\ 1\big),
  \qquad
  p_i = \frac{\exp(\mathrm{score}_i / \tau)}
             {\sum_{j \in \mathcal{C}} \exp(\mathrm{score}_j / \tau)},
  \label{eq:edge-sample}
\end{equation}
where $\mathrm{score}_i$ is the score of edge $i$ from \eqref{eq:edge-score},
$\tilde{p}_i$ is $p_i$ renormalized over the nucleus, and $k$ is the number
of frontier nodes. This rule bounds the expected number of selected edges
per hop by the frontier size; it does not enforce equal expansion across
frontier nodes, and multiple selected edges may lead to the same node.
Lower-probability edges retained in the nucleus may still be sampled:
retrieval explores the graph rather than looking it up.

\paragraph{Subgraph recall.}
The returned representation contains the anchors, the nodes reached by
expansion, all their facets, the edges internal to this set, and the
edges leading outward (directions available for further
expansion, reported as outgoing-relation metadata). The agent receives a
structured subgraph, potentially containing multiple connected
components, rather than a flat ranked list.

\section{Experiments}
\label{sec:experiments}

\subsection{Setup}
\label{sec:setup}

\paragraph{Experimental setup.}
Runs last 30 Minecraft game days on one fixed world seed (seed 7) at hard
difficulty with inventory retained after death; game time advances only during
action execution. Each group has $N=3$ agents in separate worlds with distinct
personas, aggregating every $K=2$ days. DeepSeek-V4.1-Flash (DS-V4.1F) is the
default backbone, GLM-5.3, hy4-preview and Kimi-K3 are additional backbones,
and GLM-5.3-Flash is the target of the transfer experiment
(\S\ref{sec:transfer}); one backbone serves all roles of a run. Every run
starts from empty knowledge stores, each method--backbone cell has 10
independent group runs, and reported values are averages over them. Full
configuration and sources of randomness are in \appref{sec:setup_details}.

\paragraph{Baselines.}
The planner-only control has the same backbone, personas, action API, world
seed and horizon but no persistent memory, reflection or skill library
(\appref{sec:app_baselines}), so it measures the combined effect of the knowledge
components, not sharing alone. On the default backbone we also re-implement two
external methods under the same world, interface and horizon: Voyager
\citep{wang2023voyager} (automatic curriculum, iterative code generation,
self-verification, skill library) and Echo \citep{li2026echo} (successful tasks
stored as structured experience, new tasks induced by analogy). Differences from both
originals are in \appref{sec:app_baselines}.

\subsection{Main results}
\label{sec:results}

We evaluate \SYSNAME{} on two long-horizon capabilities over 30 game
days: exploration progress and survival. Every cell contains 10
independent group runs of three agents on world seed 7.
Table~\ref{tab:main} compares methods on the default backbone, and
Table~\ref{tab:backbones} compares \SYSNAME{} with its planner-only control
across backbones. Metrics are defined in \appref{sec:app_metrics};
per-run values are in \appref{sec:app_full_results}.

\begin{table}[htbp]
\centering
\caption{Comparison with baselines on DeepSeek-V4.1-Flash (30 game days, 10
runs of three agents per method). Deaths, Repeat (deaths repeating an
earlier one of the same agent) and XP/life are per-agent means. Tool era
(dmd.\ = diamond), Obsidian and Nether give the mean first-attainment day
over the agents that reached it, with their count out of 30; --- means none.
Definitions: \appref{sec:app_metrics}.}
\label{tab:main}
\scriptsize
\setlength{\tabcolsep}{1.5pt}
\begin{tabular}{lccc|ccc}
\toprule
 & \multicolumn{3}{c|}{Survival} & \multicolumn{3}{c}{Exploration} \\
\cmidrule(lr){2-4}\cmidrule(lr){5-7}
Method & Deaths\,$\downarrow$ & Repeat\,$\downarrow$ & XP/life\,$\uparrow$ & Tool era & Obsidian & Nether \\
\midrule
Voyager & $228.9$ & $203.1$ & $1.0$ & $dmd.\,(3.0\,\mathrm{day},\,1/30)$ & --- & --- \\
Echo & $73.6$ & $53.1$ & $4.2$ & $dmd.\,(15.8\,\mathrm{day},\,26/30)$ & --- & --- \\
Planner-only & $27.9$ & $15.1$ & $14.0$ & $dmd.\,(15.2\,\mathrm{day},\,16/30)$ & --- & --- \\
\SYSNAME{} & $\mathbf{16.2}$ & $\mathbf{7.4}$ & $\mathbf{18.2}$ & $dmd.\,(13.7\,\mathrm{day},\,24/30)$ & $22.8\,\mathrm{day}\,(6/30)$ & $23.8\,\mathrm{day}\,(4/30)$ \\
\bottomrule
\end{tabular}
\end{table}

\begin{table}[htbp]
\centering
\caption{\SYSNAME{} and its planner-only control across backbones (30
game days, 10 runs of three agents per cell). All metrics and conventions
are as in Table~\ref{tab:main}; --- means none attained.}
\label{tab:backbones}
\scriptsize
\setlength{\tabcolsep}{1.2pt}
\begin{tabular}{llccc|ccc}
\toprule
 & & \multicolumn{3}{c|}{Survival} & \multicolumn{3}{c}{Exploration} \\
\cmidrule(lr){3-5}\cmidrule(lr){6-8}
Backbone & Method & Deaths\,$\downarrow$ & Repeat\,$\downarrow$ & XP/life\,$\uparrow$ & Tool era & Obsidian & Nether \\
\midrule
\raisebox{-4pt}[0pt][0pt]{DS-V4.1F} & Planner-only & $27.9$ & $15.1$ & $14.0$ & $dmd.\,(15.2\,\mathrm{day},\,16/30)$ & --- & --- \\
 & \SYSNAME{} & $16.2$ & $7.4$ & $18.2$ & $dmd.\,(13.7\,\mathrm{day},\,24/30)$ & $22.8\,\mathrm{day}\,(6/30)$ & $23.8\,\mathrm{day}\,(4/30)$ \\
\midrule
\raisebox{-4pt}[0pt][0pt]{GLM-5.3} & Planner-only & $33.4$ & $17.4$ & $8.1$ & $dmd.\,(15.2\,\mathrm{day},\,21/30)$ & --- & --- \\
 & \SYSNAME{} & $28.6$ & $16.9$ & $24.1$ & $dmd.\,(13.8\,\mathrm{day},\,28/30)$ & $23.9\,\mathrm{day}\,(10/30)$ & $27.0\,\mathrm{day}\,(2/30)$ \\
\midrule
\raisebox{-4pt}[0pt][0pt]{hy4-preview} & Planner-only & $47.0$ & $28.8$ & $8.2$ & $dmd.\,(16.3\,\mathrm{day},\,29/30)$ & --- & --- \\
 & \SYSNAME{} & $15.6$ & $8.3$ & $32.4$ & $dmd.\,(18.8\,\mathrm{day},\,13/30)$ & $28.0\,\mathrm{day}\,(1/30)$ & $29.0\,\mathrm{day}\,(1/30)$ \\
\midrule
\raisebox{-4pt}[0pt][0pt]{Kimi-K3} & Planner-only & $64.9$ & $48.7$ & $7.0$ & $dmd.\,(21.7\,\mathrm{day},\,15/30)$ & --- & --- \\
 & \SYSNAME{} & $42.1$ & $27.0$ & $6.5$ & $dmd.\,(19.1\,\mathrm{day},\,10/30)$ & $22.0\,\mathrm{day}\,(2/30)$ & --- \\
\bottomrule
\end{tabular}
\end{table}

\paragraph*{Better exploration: completing the Nether route.}
We find that the hardest part within 30 days is obtaining obsidian, building a
portal and entering the Nether, a route through a diamond pickaxe, an iron bucket, a
lava source doused with water and a frame lit with flint and steel. Only
\SYSNAME{} completes it (Table~\ref{tab:main}): $6$ of its $30$ agents obtain
obsidian, on day $22.8$ on average, and $4$ enter the Nether, on day $23.8$,
whereas no agent of any baseline does either. Diamond-tier tools are
reached by $24$ \SYSNAME{} agents, against $16$ for the planner-only control,
$26$ for Echo and $1$ for Voyager. Across backbones
(Table~\ref{tab:backbones}), \SYSNAME{} obtains obsidian on all four and
enters the Nether on three, whereas none of the $120$ planner-only agents
does either. Voyager's authors note that abstract exploration goals are
``challenging to execute without an appropriate curriculum'' and add a
dedicated curriculum module \citep{wang2023voyager}; \SYSNAME{} has none, and
the acting agent sets its own goal each day from its state and accumulated
knowledge. On three of the four backbones, agents following that route reach
its end within 30 days with no externally assigned task, whereas Voyager
does not get past iron tools in $29$ of its $30$ agents.

\begin{figure}[htbp]
\centering
\includegraphics[width=\linewidth]{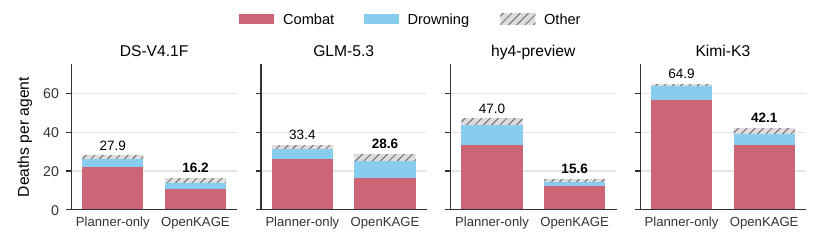}
\caption{Deaths per agent over 30 days by cause (combat: hostile mobs,
arrows, explosions); bar-end numbers are totals.}
\label{fig:death_causes}
\end{figure}

\paragraph*{Better survival: fewer repeated mistakes.}
Mean deaths per agent fall on every backbone (Table~\ref{tab:backbones}).
Two readouts locate the reduction. Deaths that repeat an earlier one (same
cause, within 16 blocks of the agent's previous death) fall from $15.1$ to
$7.4$ per agent on the default backbone, and also on hy4-preview and
Kimi-K3, and make up a smaller share of all deaths ($45\%$ under \SYSNAME{}
against $52\%$ for the planner-only control, $70\%$ for Echo and $88\%$ for
Voyager, whose isolated agents often fall into the same lethal situation
dozens of times; mean of per-run shares). Combat deaths from hostile mobs,
arrows and explosions fall on four backbones
(Figure~\ref{fig:death_causes}; $22.2$ to $10.5$ per agent on the default
backbone), while drowning falls clearly only on hy4-preview and rises on
GLM-5.3, the backbone where the most agents obtain obsidian: obtaining
obsidian requires working beside water and lava, which is very dangerous. Agents
learn to handle threats they meet repeatedly, and the remaining deaths shift
towards the hazards of deeper exploration. This is not bought by exploring
less: mean XP per life is higher on three of the four backbones
(markedly on GLM-5.3 and hy4-preview).

\paragraph*{Shared knowledge: growing and reused across agents.}
On the default backbone the skill library available to each agent grows almost
linearly to $132.5$ skills by day 30, of which an agent writes about $44$
itself; on every backbone $62$ to $65\%$ of skill calls invoke a skill written
by another agent, and $94\%$ of those succeed on the default backbone
(success means the call returned without error; growth curves, failure rates
and skill revisions are in \appref{sec:app_full_results}). The knowledge
components cost more: per game night \SYSNAME{} makes $21\%$ more LLM calls than
the planner-only control, processes about $4.2$ times as much text, and takes
about $1.6$ times as much wall-clock time per complete run, the extra time cost
of reflection and aggregation (\appref{sec:app_cost}).

\subsection{Ablations}
\label{sec:ablation}

All conditions run on DeepSeek-V4.1-Flash with 10 group runs each; Full and B
are the \SYSNAME{} and planner-only cells of Table~\ref{tab:main}, and each
condition is defined in \appref{sec:ablation_conditions}. These comparisons
provide evidence about component contributions, but do not establish
interaction-free effects.

\begin{table}[htbp]
\centering
\caption{Ablation results (DeepSeek-V4.1-Flash, 30 game days, 10 runs per
condition; metrics as in Table~\ref{tab:main}). Conditions are named by what
they remove: \abl{S} sharing, \abl{R} subgraph retrieval, \abl{M} the memory
layer, \abl{E} the experience layer, B all graphs; Raw keeps every component
but shares experience records verbatim (\appref{sec:ablation_conditions}).}
\label{tab:ablation}
\scriptsize
\setlength{\tabcolsep}{1.5pt}
\begin{tabular}{lccc|ccc}
\toprule
 & \multicolumn{3}{c|}{Survival} & \multicolumn{3}{c}{Exploration} \\
\cmidrule(lr){2-4}\cmidrule(lr){5-7}
Condition & Deaths\,$\downarrow$ & Repeat\,$\downarrow$ & XP/life\,$\uparrow$ & Tool era & Obsidian & Nether \\
\midrule
Full & $16.2$ & $7.4$ & $18.2$ & $dmd.\,(13.7\,\mathrm{day},\,24/30)$ & $22.8\,\mathrm{day}\,(6/30)$ & $23.8\,\mathrm{day}\,(4/30)$ \\
Raw & $17.6$ & $10.8$ & $21.6$ & $dmd.\,(13.0\,\mathrm{day},\,14/30)$ & --- & --- \\
\abl{S} & $17.4$ & $8.1$ & $16.1$ & $dmd.\,(13.3\,\mathrm{day},\,21/30)$ & $21.5\,\mathrm{day}\,(2/30)$ & $21.0\,\mathrm{day}\,(1/30)$ \\
\abl{R} & $15.6$ & $7.4$ & $13.3$ & $dmd.\,(13.3\,\mathrm{day},\,12/30)$ & --- & --- \\
\abl{M} & $11.6$ & $4.1$ & $37.7$ & $dmd.\,(17.9\,\mathrm{day},\,19/30)$ & --- & --- \\
\abl{E} & $14.5$ & $7.4$ & $21.7$ & $dmd.\,(14.0\,\mathrm{day},\,15/30)$ & --- & --- \\
B & $27.9$ & $15.1$ & $14.0$ & $dmd.\,(15.2\,\mathrm{day},\,16/30)$ & --- & --- \\
\bottomrule
\end{tabular}
\end{table}

\begin{itemize}
\item \textbf{Obsidian and the Nether are delicate indicators.} Open-ended
long-horizon exploration does not let us quantify the contribution of each
component at fine granularity. Full reaches obsidian with $6$ of $30$ agents (in
$4$ of $10$ runs) and the Nether with $4$ (in $3$ runs). Removing any one of
retrieval, the memory layer or the experience layer (\abl{R}, \abl{M},
\abl{E}) leaves no agent with obsidian or the Nether in 30 days, and
removing sharing (\abl{S}) cuts the numbers to two agents and one. This suggests
that sharing promotes the flow of knowledge within the group: agents able to
reach higher achievements pass on knowledge that lifts others too.
\item \textbf{Direct sharing without consolidation.} Raw appends records
verbatim with its skill channel active ($60\%$ of skill calls), yet reaches
neither deep milestone: $16$ of $30$ agents never get a diamond pickaxe ($6$
under Full), and of the $14$ that do, $8$ set an obsidian goal (Full: $18/24$)
but only $2$ obtain an iron bucket (Full: $13/24$) and none obtains obsidian.
Nor is missing route knowledge the whole story: two Raw agents had the complete
pouring procedure and matching skills in both their shared records and
retrieved blocks, followed them for days, and still failed. One possible attribution is
low effective information: verbatim records keep events one by one, so a
recurring failure reappears as several similar entries while the actionable
rule takes up little of the retrieved block, and agents read what happened
rather than what to do next. The same errors then recur and agents die or run
out of time before the prerequisites are in place, whereas consolidation
merges repeated failures into one rule. This fits the gap appearing at the
earlier stages.
\end{itemize}

\subsection{Cross-group and cross-backbone knowledge transfer}
\label{sec:transfer}

\begin{figure}[htbp]
\centering
\includegraphics[width=\linewidth]{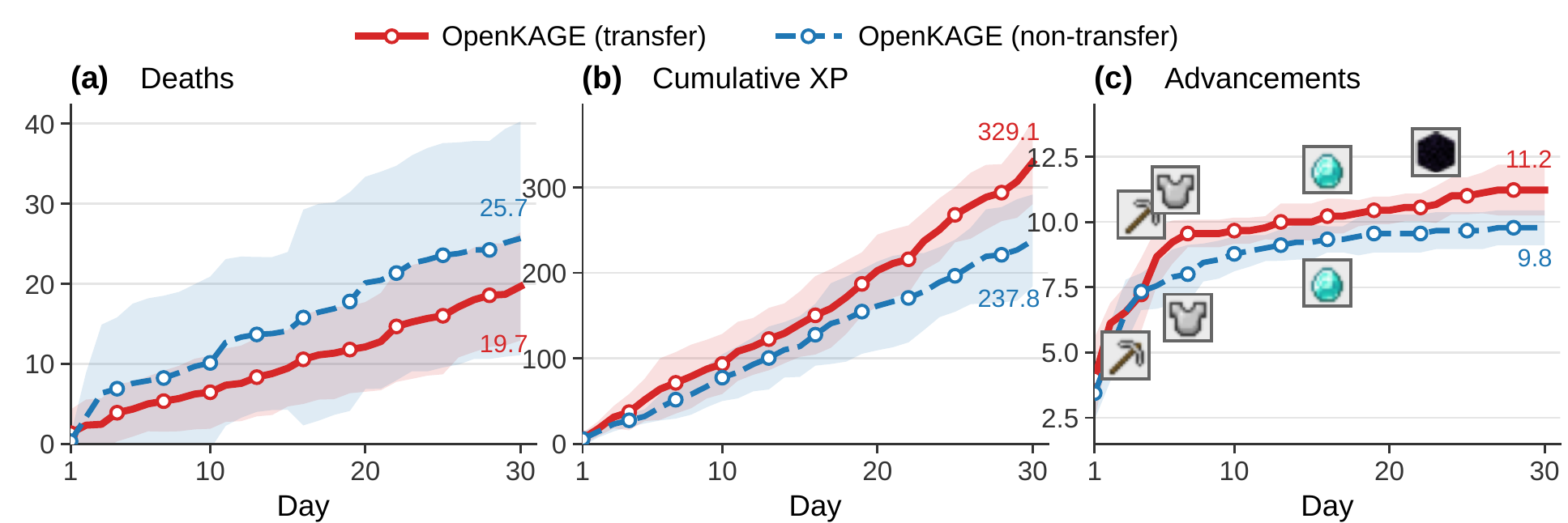}
\caption{Knowledge transfer (transfer solid, non-transfer dashed; three runs
per group, nine-agent means, bands $\pm 1$ sd across agents). Icons in (c) mark
the median first-reach night of the iron pickaxe, first armor, first diamond
and obsidian; their height is arbitrary.}
\label{fig:transfer}
\end{figure}

Does the value of the accumulated knowledge stores survive the group, and the
backbone, that produced them? The \textbf{transfer group} is a fresh
GLM-5.3-Flash group whose stores are initialized with the shared graph and skill
libraries that a DeepSeek-V4.1-Flash \SYSNAME{} group accumulated in 30 days;
the \textbf{non-transfer group} starts from empty stores. Both follow the same
30-day protocol, three runs each. GLM-5.3-Flash is a weak backbone for this
world: without transferred knowledge, larger groups do not reach obsidian or
the Nether within 30 days, and doubling the horizon reaches obsidian only on
day $54$ (\appref{sec:app_flash}).

The two groups separate (Figure~\ref{fig:transfer}): deaths fall from 25.7 to
19.7 per agent, final cumulative XP rises from 237.8 to 329.1 ($+38\%$),
diamonds are reached by seven of nine transfer agents against three of nine,
and obsidian by one of nine against none. In this configuration, initializing a
new group with the accumulated stores improves survival and exploration
over empty-store initialization.

\section{Related Work}
\label{sec:related}

\paragraph{Open-world learning and agent memory.}
\label{sec:related-openworld}
Voyager combines an automatic curriculum with a growing executable skill
library for Minecraft exploration \citep{wang2023voyager}; \SYSNAME{}
instead places goal generation inside the acting agent's own loop
(\S\ref{sec:results}); this concerns how goals are produced, not whether
exploration is autonomous. Open-endedness research likewise generates the
next challenge outside the learner, by co-evolving environments with their
solvers \citep{wang2019poet} or by asking a foundation model which learnable
tasks are interesting \citep{zhang2023omni}; both concern what a single
learner pursues next, whereas \SYSNAME{} concerns what a group keeps and
shares. We use ``open-ended'' to mean no predefined endpoint, with progress
measured by milestones and no novelty criterion. Other agents organize
long-horizon behavior around injected task knowledge \citep{zhu2023gitm,wang2023jarvis1}, a hybrid
graph-and-pool memory \citep{li2024optimus1}, an episodic What-Where-When
memory \citep{park2024mrsteve}, a knowledge-graph world model
\citep{anokhin2024arigraph} or a causal graph \citep{yu2024adam,tang2024mars}.
In \SYSNAME{} the stores start empty, and the distinction is the unit of
accumulation: the memory--experience split becomes a sharing boundary, so
situated records stay local while transferable knowledge is consolidated.

\paragraph{Multi-agent experience sharing.}
\label{sec:related-sharing}
Multi-agent frameworks coordinate through communication and role division
\citep{li2023camel,hong2023metagpt}; GEA treats the group as an evolutionary
unit and pools members' trajectories \citep{weng2026gea}. Parameter-level
sharing includes federated continual learning, which aggregates knowledge
through parameters and leaves no inspectable record \citep{dong2023fcl}, and
selective decentralized reuse: agents relay a few high-priority transitions
\citep{gerstgrasser2023seps} or retrieve and compose policy modules by task
similarity \citep{nath2026policy}, both with tasks given in advance. Memory
systems for multi-agent LLMs are closer in form. G-Memory retrieves related
past task queries with their insights and trajectories, filtered by role
\citep{zhang2025gmemory}; its hierarchy is keyed by task queries, which an
open-ended setting does not provide. LLMA-Mem keeps episodic, procedural and
transactive memory under local, shared or hybrid topologies on assigned tasks
\citep{wu2026llmamem}; its hybrid topology likewise keeps episodes local and
shares consolidated procedures, so that design choice is not new to us. DAMCS
gives each LLM agent a hierarchical knowledge graph and exchanges structured
messages with chosen partners toward a fixed goal \citep{yang2025damcs},
without consolidating them into a group store.
\SYSNAME{} differs in drawing the boundary by kind of knowledge rather than
priority or partner, in sharing a structured graph whose applicability
conditions stay with executable skills, in returning subgraphs rather than a
ranked list, and in running without assigned tasks. A controlled comparison
with verbatim sharing (\appref{sec:app_raw_sharing}) isolates consolidation
inside an active sharing channel, and GEA, G-Memory and LLMA-Mem are compared
with \SYSNAME{} at the level of design (\appref{sec:app_comparison}).

\paragraph{Retrieval-based memory.}
\label{sec:related-memory}
Retrieval-augmented generation retrieves external knowledge
\citep{lewis2020rag}, and GraphRAG and AriGraph use graph
representations \citep{edge2024graphrag,anokhin2024arigraph}. In agent memory
the same pattern usually yields a flat ranked list \citep{wang2023voyager}:
nearest neighbors of the query enter the context ranked by similarity, without
the dependencies, preconditions and failure modes that bind them to the
situation. \SYSNAME{} instead returns a subgraph, via vector anchoring and
query-conditioned relational expansion, over knowledge consolidated across
agents. \appref{sec:app_comparison} compares systems in detail.

\section{Conclusion}
\label{sec:conclusion}

We presented \SYSNAME{}, a framework that drives lifelong learning and
collective evolution of agents: it separates private situated memories from
shareable experience, consolidates textual experience and shares executable
skills across the group, and returns this knowledge to decisions through
intent-aware subgraph retrieval. Across
four backbones and 30-day Minecraft runs, it reached milestones that a
planner-only control, Voyager and Echo did not, and reduced deaths, mostly
repeated ones; its shared skills were reused across agents, and transferred
knowledge helped a new group on another backbone.

\paragraph{Limitations and outlook.}
Our evidence is limited to Minecraft, one world seed and a 30-day horizon (60
days only in the scaling runs), so it speaks to open-ended lifelong learning
only to a limited extent, although to our knowledge it is the longest-running
study in an open-ended world to date. At 10 runs per condition the ablations do not resolve
which component they depend on.

\subsection*{AI use statement}
Generative AI tools were used in three capacities during the preparation of
this work, each subject to human oversight: (i) to aid and polish the writing
of the manuscript; (ii) for retrieval and discovery, e.g., locating and
surveying related work; and (iii) for research ideation and execution,
including assistance with code development and experiment orchestration. All
experimental design, data interpretation, and conclusions are the
responsibility of the authors.

\subsection*{Ethics statement}
This work is conducted in the Minecraft simulation environment and involves
no human subjects or real-world intervention. The agents' knowledge-sharing
mechanism involves no user data.

\subsection*{Reproducibility statement}
Experimental settings, prompts, interfaces, and metric definitions are
documented in the main text and appendices; \appref{sec:setup_details}
tabulates the full run configuration. Model configurations,
interactions, and daily environment snapshots are logged to support result
verification. The framework, baselines, prompts, metric and figure scripts
are available at \url{https://github.com/zhuzhuxia5678/OpenKAGE}, and the
per-run records at \url{https://github.com/zhuzhuxia5678/OpenKAGE_results}.
These artifacts support recomputing the reported metrics and regenerating
the result figures.

\bibliography{iclr2027_conference}

@article{wang2023voyager,
  title   = {Voyager: An Open-Ended Embodied Agent with Large Language Models},
  author  = {Wang, Guanzhi and Xie, Yuqi and Jiang, Yunfan and Mandlekar, Ajay and
             Xiao, Chaowei and Zhu, Yuke and Fan, Linxi and Anandkumar, Anima},
  year    = {2024},
  journal = {Transactions on Machine Learning Research},
  issn    = {2835-8856},
  url     = {https://openreview.net/forum?id=ehfRiF0R3a}
}

@inproceedings{li2026echo,
  title     = {Experience Transfer for Multimodal {LLM} Agents in {Minecraft} Game},
  author    = {Li, Chenghao and Liu, Jun and Zhang, Songbo and Jian, Huadong and
               Ni, Hao and Lee, Lik-Hang and Bae, Sung-Ho and Wang, Guoqing and
               Yang, Yang and Zhang, Chaoning},
  booktitle = {IEEE/CVF Conference on Computer Vision and Pattern Recognition (CVPR)},
  year      = {2026},
  url       = {https://arxiv.org/abs/2604.05533}
}

@inproceedings{yu2024adam,
  title     = {{ADAM}: An Embodied Causal Agent in Open-World Environments},
  author    = {Yu, Shu and Lu, Chaochao},
  booktitle = {International Conference on Learning Representations (ICLR)},
  year      = {2025},
  pages     = {22886--22905},
  url       = {https://arxiv.org/abs/2410.22194}
}

@inproceedings{tang2024mars,
  title     = {Mars: Situated Inductive Reasoning in an Open-World Environment},
  author    = {Tang, Xiaojuan and Li, Jiaqi and Liang, Yitao and Zhu, Song-Chun and
               Zhang, Muhan and Zheng, Zilong},
  booktitle = {Advances in Neural Information Processing Systems (NeurIPS),
               Datasets and Benchmarks Track},
  year      = {2024},
  volume    = {37},
  pages     = {17830--17869},
  url       = {https://arxiv.org/abs/2410.08126}
}

@inproceedings{li2024optimus1,
  title     = {{Optimus-1}: Hybrid Multimodal Memory Empowered Agents Excel in Long-Horizon Tasks},
  author    = {Li, Zaijing and Xie, Yuquan and Shao, Rui and Chen, Gongwei and
               Jiang, Dongmei and Nie, Liqiang},
  booktitle = {Advances in Neural Information Processing Systems (NeurIPS)},
  year      = {2024},
  volume    = {37},
  pages     = {49881--49913},
  url       = {https://arxiv.org/abs/2408.03615}
}

@inproceedings{park2024mrsteve,
  title     = {{MrSteve}: Instruction-Following Agents in {Minecraft} with What-Where-When Memory},
  author    = {Park, Junyeong and Cho, Junmo and Ahn, Sungjin},
  booktitle = {International Conference on Learning Representations (ICLR)},
  year      = {2025},
  pages     = {16922--16960},
  url       = {https://arxiv.org/abs/2411.06736}
}

@article{weng2026gea,
  title   = {Group-Evolving Agents: Open-Ended Self-Improvement via Experience Sharing},
  author  = {Weng, Zhaotian and Antoniades, Antonis and Nathani, Deepak and
             Zhang, Zhen and Pu, Xiao and Wang, Xin Eric},
  year    = {2026},
  journal = {arXiv preprint arXiv:2602.04837},
  url     = {https://arxiv.org/abs/2602.04837}
}

@article{wang2023jarvis1,
  title   = {{JARVIS-1}: Open-World Multi-Task Agents with Memory-Augmented Multimodal Language Models},
  author  = {Wang, Zihao and Cai, Shaofei and Liu, Anji and Jin, Hongyuan and Hou, Chao and
             Zhang, Ying and Lin, Xiaojian and He, Xiangyu and Zheng, Zilong and
             Yang, Yu and Ma, Xiaojian and Liang, Yitao},
  journal = {IEEE Transactions on Pattern Analysis and Machine Intelligence},
  volume  = {47},
  number  = {3},
  pages   = {1894--1907},
  year    = {2025},
  url     = {https://arxiv.org/abs/2311.05997}
}

@inproceedings{li2023camel,
  title     = {{CAMEL}: Communicative Agents for ``Mind'' Exploration of Large Language Model Society},
  author    = {Li, Guohao and Hammoud, Hasan Abed Al Kader and Itani, Hani and
               Khizbullin, Dmitrii and Ghanem, Bernard},
  booktitle = {Advances in Neural Information Processing Systems (NeurIPS)},
  year      = {2023},
  volume    = {36},
  pages     = {51991--52008},
  url       = {https://arxiv.org/abs/2303.17760}
}

@inproceedings{hong2023metagpt,
  title     = {{MetaGPT}: Meta Programming for A Multi-Agent Collaborative Framework},
  author    = {Hong, Sirui and Zhuge, Mingchen and Chen, Jiaqi and Zheng, Xiawu and
               Cheng, Yuheng and Zhang, Ceyao and Wang, Jinlin and Wang, Zili and
               Yau, Steven Ka Shing and Lin, Zijuan and Zhou, Liyang and Ran, Chenyu
               and Xiao, Lingfeng and Wu, Chenglin and Schmidhuber, J{\"u}rgen},
  booktitle = {International Conference on Learning Representations (ICLR)},
  year      = {2024},
  pages     = {23247--23275},
  url       = {https://arxiv.org/abs/2308.00352}
}

@inproceedings{baker2022vpt,
  title     = {Video PreTraining ({VPT}): Learning to Act by Watching Unlabeled Online Videos},
  author    = {Baker, Bowen and Akkaya, Ilge and Zhokhov, Peter and Huizinga, Joost and
               Tang, Jie and Ecoffet, Adrien and Houghton, Brandon and Sampedro, Raul and
               Clune, Jeff},
  booktitle = {Advances in Neural Information Processing Systems (NeurIPS)},
  year      = {2022},
  url       = {https://arxiv.org/abs/2206.11795}
}

@article{zhu2023gitm,
  title   = {Ghost in the {Minecraft}: Generally Capable Agents for Open-World Environments via Large Language Models with Text-based Knowledge and Memory},
  author  = {Zhu, Xizhou and Chen, Yuntao and Tian, Hao and Tao, Chenxin and Su, Weijie and
             Yang, Chenyu and Huang, Gao and Li, Bin and Lu, Lewei and Wang, Xiaogang and
             Qiao, Yu and Zhang, Zhaoxiang and Dai, Jifeng},
  year    = {2023},
  journal = {arXiv preprint arXiv:2305.17144},
  url     = {https://arxiv.org/abs/2305.17144}
}

@article{hafner2023dreamerv3,
  title   = {Mastering Diverse Domains through World Models},
  author  = {Hafner, Danijar and Pasukonis, Jurgis and Ba, Jimmy and Lillicrap, Timothy},
  year    = {2023},
  journal = {arXiv preprint arXiv:2301.04104},
  url     = {https://arxiv.org/abs/2301.04104}
}

@inproceedings{anokhin2024arigraph,
  title     = {{AriGraph}: Learning Knowledge Graph World Models with Episodic Memory for {LLM} Agents},
  author    = {Anokhin, Petr and Semenov, Nikita and Sorokin, Artyom and Evseev, Dmitry and
               Kravchenko, Andrey and Burtsev, Mikhail and Burnaev, Evgeny},
  booktitle = {Proceedings of the Thirty-Fourth International Joint Conference on
               Artificial Intelligence (IJCAI)},
  year      = {2025},
  pages     = {12--20},
  url       = {https://arxiv.org/abs/2407.04363}
}

@inproceedings{wang2019poet,
  title     = {Paired Open-Ended Trailblazer ({POET}): Endlessly Generating Increasingly Complex
               and Diverse Learning Environments and Their Solutions},
  author    = {Wang, Rui and Lehman, Joel and Clune, Jeff and Stanley, Kenneth O.},
  booktitle = {Proceedings of the Genetic and Evolutionary Computation Conference (GECCO)},
  year      = {2019},
  pages     = {142--151},
  url       = {https://arxiv.org/abs/1901.01753}
}

@article{dong2023fcl,
  title   = {Federated Continual Learning via Knowledge Fusion: A Survey},
  author  = {Yang, Xin and Yu, Hao and Gao, Xin and Wang, Hao and Zhang, Junbo and Li, Tianrui},
  journal = {IEEE Transactions on Knowledge and Data Engineering},
  volume  = {36},
  number  = {8},
  pages   = {3832--3850},
  year    = {2024},
  url     = {https://arxiv.org/abs/2312.16475}
}

@inproceedings{yao2023react,
  title     = {{ReAct}: Synergizing Reasoning and Acting in Language Models},
  author    = {Yao, Shunyu and Zhao, Jeffrey and Yu, Dian and Du, Nan and
               Shafran, Izhak and Narasimhan, Karthik and Cao, Yuan},
  booktitle = {International Conference on Learning Representations (ICLR)},
  year      = {2023},
  url       = {https://arxiv.org/abs/2210.03629}
}

@inproceedings{shinn2023reflexion,
  title     = {Reflexion: Language Agents with Verbal Reinforcement Learning},
  author    = {Shinn, Noah and Cassano, Federico and Gopinath, Ashwin and
               Narasimhan, Karthik and Yao, Shunyu},
  booktitle = {Advances in Neural Information Processing Systems (NeurIPS)},
  year      = {2023},
  volume    = {36},
  pages     = {8634--8652},
  url       = {https://arxiv.org/abs/2303.11366}
}

@misc{autogpt2023,
  title        = {{AutoGPT}},
  author       = {{Significant Gravitas}},
  year         = {2023},
  howpublished = {https://github.com/Significant-Gravitas/AutoGPT},
  note         = {Autonomous LLM agent loop; accessed 2026-09-25}
}

@inproceedings{lewis2020rag,
  title     = {Retrieval-Augmented Generation for Knowledge-Intensive {NLP} Tasks},
  author    = {Lewis, Patrick and Perez, Ethan and Piktus, Aleksandra and
               Petroni, Fabio and Karpukhin, Vladimir and Goyal, Naman and
               K{\"u}ttler, Heinrich and Lewis, Mike and Yih, Wen-tau and
               Rockt{\"a}schel, Tim and Riedel, Sebastian and Kiela, Douwe},
  booktitle = {Advances in Neural Information Processing Systems (NeurIPS)},
  year      = {2020},
  url       = {https://arxiv.org/abs/2005.11401}
}

@article{edge2024graphrag,
  title   = {From Local to Global: A Graph {RAG} Approach to Query-Focused Summarization},
  author  = {Edge, Darren and Trinh, Ha and Cheng, Newman and Bradley, Joshua and
             Chao, Alex and Mody, Apurva and Truitt, Steven and Larson, Jonathan},
  year    = {2024},
  journal = {arXiv preprint arXiv:2404.16130},
  url     = {https://arxiv.org/abs/2404.16130}
}

@inproceedings{park2023generative,
  title     = {Generative Agents: Interactive Simulacra of Human Behavior},
  author    = {Park, Joon Sung and O'Brien, Joseph and Cai, Carrie Jun and
               Morris, Meredith Ringel and Liang, Percy and Bernstein, Michael S.},
  booktitle = {Proceedings of the 36th Annual ACM Symposium on User Interface Software and Technology (UIST)},
  year      = {2023},
  url       = {https://arxiv.org/abs/2304.03442}
}

@article{kumaran2016cls,
  title   = {What Learning Systems do Intelligent Agents Need? Complementary
             Learning Systems Theory Updated},
  author  = {Kumaran, Dharshan and Hassabis, Demis and McClelland, James L.},
  journal = {Trends in Cognitive Sciences},
  volume  = {20},
  number  = {7},
  pages   = {512--534},
  year    = {2016}
}

@article{zhao2025kalm,
  title   = {{KaLM-Embedding-V2}: Superior Training Techniques and Data Inspire a
             Versatile Embedding Model},
  author  = {Zhao, Xinping and Hu, Xinshuo and Shan, Zifei and Huang, Shouzheng and
             Zhou, Yao and Zhang, Xin and Sun, Zetian and Liu, Zhenyu and
             Li, Dongfang and Wei, Xinyuan and Pan, Youcheng and Xiang, Yang},
  year    = {2025},
  journal = {arXiv preprint arXiv:2506.20923},
  url     = {https://arxiv.org/abs/2506.20923}
}

@inproceedings{zhang2025gmemory,
  title     = {{G-Memory}: Tracing Hierarchical Memory for Multi-Agent Systems},
  author    = {Zhang, Guibin and Fu, Muxin and Wan, Guancheng and Yu, Miao and
               Wang, Kun and Yan, Shuicheng},
  booktitle = {Advances in Neural Information Processing Systems},
  volume    = {38},
  year      = {2025},
  url       = {https://arxiv.org/abs/2506.07398}
}

@article{wu2026llmamem,
  title   = {Scaling Teams or Scaling Time? Memory Enabled Lifelong Learning in
             {LLM} Multi-Agent Systems},
  author  = {Wu, Shanglin and Luo, Yuyang and Liang, Yueqing and Shi, Kaiwen and
             Ye, Yanfang and Payani, Ali and Shu, Kai},
  year    = {2026},
  journal = {arXiv preprint arXiv:2604.03295},
  url     = {https://arxiv.org/abs/2604.03295}
}

@article{zhang2023omni,
  title   = {{OMNI}: Open-endedness via Models of human Notions of Interestingness},
  author  = {Zhang, Jenny and Lehman, Joel and Stanley, Kenneth and Clune, Jeff},
  year    = {2023},
  journal = {arXiv preprint arXiv:2306.01711},
  url     = {https://arxiv.org/abs/2306.01711}
}

@inproceedings{gerstgrasser2023seps,
  title     = {Selectively Sharing Experiences Improves Multi-Agent Reinforcement Learning},
  author    = {Gerstgrasser, Matthias and Danino, Tom and Keren, Sarah},
  booktitle = {Advances in Neural Information Processing Systems},
  volume    = {36},
  year      = {2023},
  url       = {https://arxiv.org/abs/2311.00865}
}

@inproceedings{nath2026policy,
  title     = {Policy Search, Retrieval, and Composition via Task Similarity in
               Collaborative Agentic Systems},
  author    = {Nath, Saptarshi and Peridis, Christos and Benjamin, Eseoghene and
               Liu, Xinran and Kolouri, Soheil and Kinnell, Peter and Li, Zexin and
               Liu, Cong and Dora, Shirin and Soltoggio, Andrea},
  booktitle = {Proceedings of the AAAI Conference on Artificial Intelligence},
  year      = {2026},
  url       = {https://arxiv.org/abs/2506.05577}
}

@inproceedings{yang2025damcs,
  title     = {{LLM}-Powered Decentralized Generative Agents with Adaptive Hierarchical
               Knowledge Graph for Cooperative Planning},
  author    = {Yang, Hanqing and Chen, Jingdi and Siew, Marie and
               Lorido-Botran, Tania and Joe-Wong, Carlee},
  booktitle = {Proceedings of the AAAI Conference on Artificial Intelligence},
  year      = {2025},
  url       = {https://arxiv.org/abs/2502.05453}
}
\bibliographystyle{iclr2027_conference}

% 附录抽到独立文件 appendix.tex（\input 不开新页，\appendix 与 \onecolumn 在其中声明）。
% appendix.tex — 全部附录（A 方法 + B 实验）。
% 由 iclr2027_conference.tex 末尾 \input{appendix.tex} 引入；
% \appendix 与 \onecolumn 在本文件声明（\input 不开新页，位置无所谓）。
% ⚠ 依赖主文件 preamble 的宏：\SYSNAME、\abl、\appref、\cmark、promptstyle。
\appendix
\onecolumn

% 浮动参数局部放宽（2026-11-12，修 B.8 大片空白）：正文 preamble 为逼中等图与
% 正文同页把 \floatpagefraction 抬到 .85，附录因此出现死区——A.10 的
% Table~11 约 74% 页高，高于 \topfraction(.7) 上不了页顶、低于 .85 凑不满
% 浮动页，作为队头卡住后同类 table 按先进先出全部积压（12/14/16/17/18），
% 直到文末 \clearpage 才集体 flush，B.8 文字页下半整页留空。附录正文页常被
% longtable 与整页图占满，这里放宽两参数让卡住的表能就近上页顶或合成浮动页。
% 仅影响此行之后遇到的浮动体，正文已 shipout 不受影响。
\renewcommand{\topfraction}{.9}
\renewcommand{\floatpagefraction}{.6}

\section{Method}
\label{sec:app_method}

All prompts quoted in this appendix are given in English translation; the
Chinese originals that drive the runs are released with the code.

\subsection{The \SYSNAME{} algorithm}
\label{sec:app_algorithm}

A run is a group of $N$ agents, each in its own world, playing $D$ game
days; agents never meet in-world and share only at the knowledge layer.
Each day has two parts, acting by day and reflecting at night, and every
$K$ days the group consolidates. The agents run in parallel, and a
barrier holds the group at the end of each night and again after each
consolidation, so no agent begins a day before the group's knowledge for
the previous one is final.

\begin{algorithm}[htbp]
\caption{\SYSNAME{}, one run. Stages marked \emph{barrier} block until all
$N$ agents have reached them.}
\label{alg:openkage}
\begin{algorithmic}[1]
\REQUIRE group of $N$ agents with local graphs $G_a$; shared graph $G^\ast$; $D$ days; window $K$
\FOR{$d = 1$ \TO $D$}
  \FOR{each agent $a$ \textbf{in parallel}}
    \STATE inject standing memory (per-category budgets) and the top-20 skill signatures
    \STATE plan a goal for the day; retrieval returns a subgraph of $G_a \cup G^\ast$ keyed to that goal
    \REPEAT
      \STATE observe, write and execute one program over the atomic API or any agent's shared skill, take feedback
      \STATE re-planning re-triggers retrieval
    \UNTIL{the day ends}
    \STATE night, stage 1: reflection writes $V_a^{\mathrm{mem}}$ and $V_a^{\mathrm{exp}}$ from the day's trajectory
    \STATE \emph{barrier}: every agent has finished stage 1
    \STATE night, stage 2: skill maintenance against the day's per-skill tallies; new and revised skills enter $G^\ast$ verbatim as isolated nodes
  \ENDFOR
  \STATE \emph{barrier}: night $d$ complete \COMMENT{day $d$'s knowledge is final}
  \IF{$d \bmod K = 0$}
    \STATE consolidation merges $\{V_{a,\mathrm{text}}^{\mathrm{exp}}\}_{a=1}^{N}$ into $G^\ast$; on success $V_{a,\mathrm{text}}^{\mathrm{exp}} \leftarrow \emptyset$
  \ENDIF
  \STATE \emph{barrier}: day $d{+}1$ begins with $G^\ast$ already updated \COMMENT{no lag}
\ENDFOR
\end{algorithmic}
\end{algorithm}

Two consequences of this schedule are worth stating. First, because
consolidation is synchronous rather than running in the background, every
agent entering day $K+1$ already sees the conclusions of window $K$ in the
shared graph, so the knowledge feedback loop has no lag; skills reach the
other agents sooner still, on the morning after they are written. Second,
because the textual part of the private experience layer is cleared once
per window, it holds one
window's output rather than the whole trajectory's, which keeps its size
bounded as $D$ grows. The consolidation step itself, namely what it
merges, is given in \appref{sec:app_aggregator}.

\subsection{Knowledge graph schema}
\label{sec:app_schema}

This appendix specifies the data model behind \S\ref{sec:two-layer}: what a
concept carries, what a facet and an edge are, how the layer and the memory
category are declared, the two kinds of experience node, and which component
may write each field. Each agent's local graph and the group's shared graph
use the same model; they differ in who writes them and in which layer they
carry. Concept names, facet labels, and relation names are free text in the
agents' working language (Chinese in our runs); as elsewhere in the paper we
quote the identifier and give an English gloss, and the originals are
released with the code.

\begin{figure}[h!]
\centering
\includegraphics[width=0.9\textwidth]{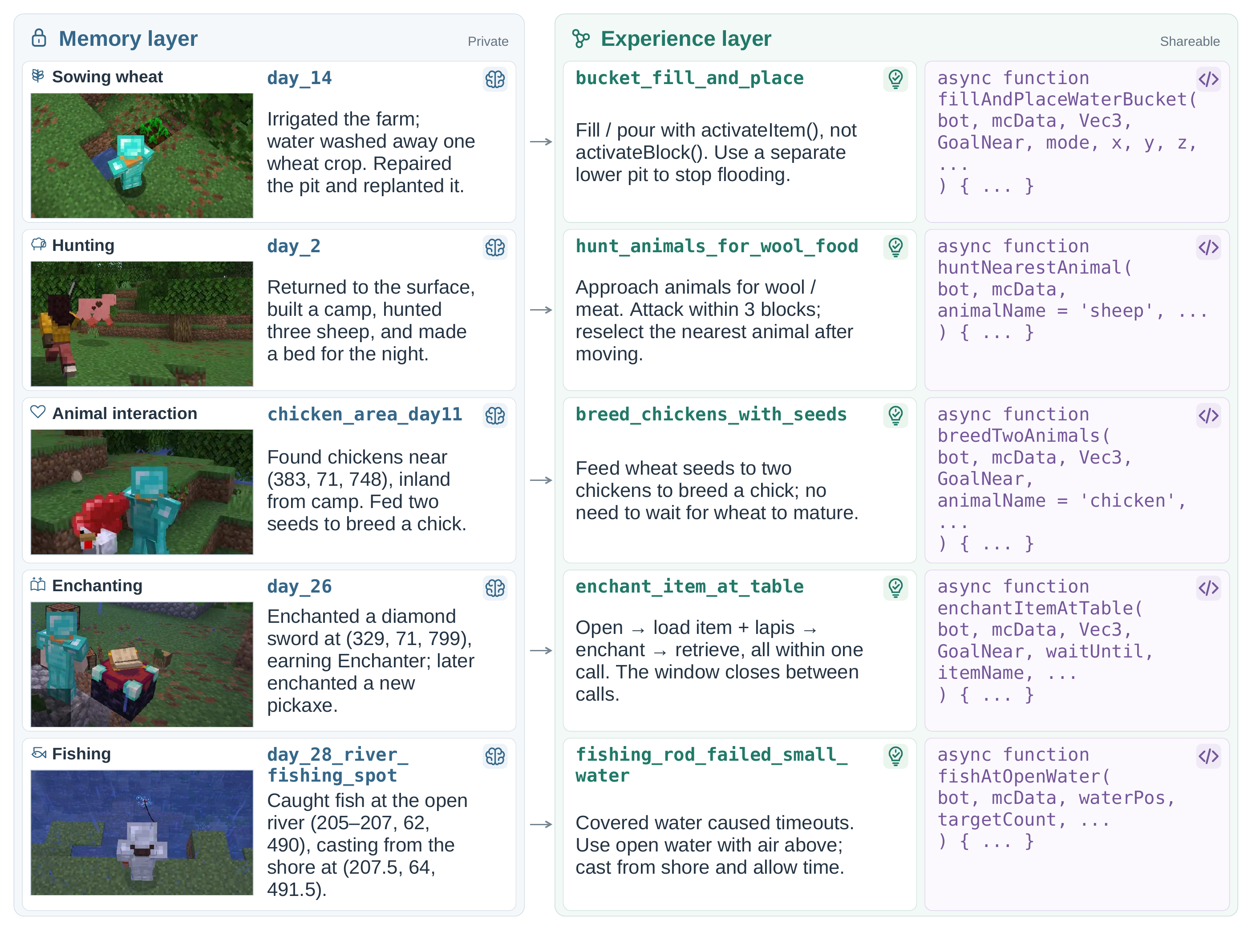}
\caption{Situated memory and transferable experience. Five examples from
different agents and runs: \textbf{left}, private memory keeps
episode-specific events and locations; \textbf{right}, the experience
layer holds reusable textual knowledge records and executable skills.
Arrows mark thematic correspondence, not graph edges.}
\label{fig:twolayer}
\end{figure}

\paragraph{Concepts.}
A concept is a stable identity for one thing the agent knows: a day, a
place, a procedure, or a hazard. It is the unit at which the graph is
addressed, retrieved, and consolidated (Table~\ref{tab:concept_fields}).
For example, \emph{day\_14 / bucket\_fill\_and\_place} is a dated,
agent-bound memory concept; Figure~\ref{fig:twolayer} shows its episode
text and coordinate frame.

\begin{table}[htbp]
\centering
\caption{Concept fields. \textbf{Who writes:} R = the reflector (private
graph), A = the aggregator (shared graph), Ret = the retrieval path of the
acting agent. No other component may write any field.}
\label{tab:concept_fields}
\small
\setlength{\tabcolsep}{4pt}
\begin{tabular}{p{2.6cm}p{7.4cm}p{2.8cm}}
\toprule
Field & Content & Written by \\
\midrule
\texttt{slug} & ASCII snake-case identifier, unique; the handle every operation addresses & R / A \\
\texttt{name} & free text; a \emph{proper} name in the memory layer, a \emph{common} name in the experience layer & R / A \\
\texttt{summary} & one-line statement of what the concept is & R / A \\
\texttt{layer} & $\mathrm{mem}$ or $\mathrm{exp}$, declared when the concept is created & R \\
\texttt{category} & \texttt{episodic}, \texttt{spatial}, \texttt{lesson} or \texttt{status}; empty for experience concepts & R (memory only) \\
\texttt{strength} & real, initialized to $1.0$; monotone, and only the retrieval path writes it & Ret only \\
timestamps & \texttt{created\_at}, \texttt{updated\_at}: game day in the local graph, window index in the shared graph & R / A, Ret \\
\bottomrule
\end{tabular}
\end{table}

Two properties of this table matter downstream. First, \texttt{layer} and
\texttt{category} are declared once, at creation, by the reflector, and are
not re-derived from content afterwards; the layer is a property of the
concept, not of any one facet, so a concept that later acquires facets of
another kind does not silently change layers. Second,
\texttt{updated\_at} advances both when the content changes and when
retrieval returns the concept. Being used counts as being active, which
is what makes it the ordering key for the morning injection.

\paragraph{Facets.}
A concept carries any number of facets, each a short statement of one
aspect: a label \texttt{kind} and one to two sentences of \texttt{content}.
At most one facet of a given kind may exist under one concept, because the
pair (concept, kind) is unique. Extending what is known about an aspect
therefore rewrites that facet in place instead of appending a second one. The kind
is a short class-label the reflector names itself, not an entry from a
fixed vocabulary; the criteria it is given (name a class, answer a question
of the form ``where is it'' or ``how is it done'', and reuse the names
visible in the graph index) are stated verbatim in the
phase-1 prompt of \appref{sec:app_reflector}.
Figure~\ref{fig:twolayer} shows the facet text of a textual experience
record derived from the irrigation episode: ``fill / pour with
\texttt{activateItem()}, not \texttt{activateBlock()}; use a separate
lower pit to stop flooding''.

\paragraph{Edges.}
An edge is typed and directed, $u \xrightarrow{r} v$, and is unique per
(from, to, relation), so the same pair may be linked by several relations.
The relation is free-named by the reflector under one constraint: the name
must read as a predicate, i.e.\ ``\emph{from} $r$ \emph{to}'' must read as
a sentence. This is why \S\ref{sec:retrieval} scores a relation by
embedding its name together with the concept it leaves from: a vague name
such as \texttt{related\_to} is semantically distant from a specific
query and therefore tends to score low. That is a property of the scores,
not a rule: no check rejects a relation name, and whether an edge is
traversed is decided by sampling
(\appref{sec:app_retrieval}), so no relation is excluded by construction.
Edges are also the one place where the two layers meet: a
local graph links an experience concept to the memory episodes that
support it, and the aggregator is shown the names of those memory-side
endpoints as labels only; its tools cannot address the memory layer
(\appref{sec:app_aggregator}).

\paragraph{Layers.}
The layer tag separates what is situated from what is transferable. The
reflector applies a single test at creation, given verbatim in
\appref{sec:app_reflector}: could someone who did not live that day take
this node and use it as it stands? Memory concepts never leave the agent;
experience concepts are what the aggregator reads, and are cleared from
the local graph once a window has been consolidated. Category is
declared for memory concepts only; it is empty for experience concepts,
which are selected by retrieval rather than injected.
Figure~\ref{fig:twolayer} shows examples of both layers from the persisted
runs: the memory concept \emph{day\_14 / bucket\_fill\_and\_place} records
``Irrigated the farm; water washed away one wheat crop. Repaired the pit
and replanted it'', while the corresponding experience record states
``fill / pour with \texttt{activateItem()}, not
\texttt{activateBlock()}; use a separate lower pit to stop flooding''
and the executable skills include \texttt{fillAndPlaceWaterBucket} and
\texttt{digStairTunnel}.

% ═══ Figure: two-layer local graph, five real moments (moved here from the
% main text §2.2, where the column width halved it) = fig4.svg ═══

\paragraph{Memory categories and the morning budget.}
The memory layer is injected each morning as a bounded slice of standing
knowledge. The slice is drawn per category rather than by a single recency
rank, because the four categories go stale at different rates, and a single
rank lets the day records written every night crowd out everything else.

\begin{table}[h!]
\centering
\caption{The four memory categories and their injection budgets. A bucket
takes the concepts of that category in decreasing order of
\texttt{updated\_at}; budgets are caps, not targets, and are not
redistributed across buckets.}
\label{tab:memory_budget}
\small
\setlength{\tabcolsep}{4pt}
\begin{tabular}{p{2.2cm}p{6.2cm}p{3.4cm}c}
\toprule
Category & Holds & Goes stale & Budget \\
\midrule
\texttt{episodic} & what is bound to a specific day: the day record, that day's blockers and findings & within a few days & 18 \\
\texttt{spatial} & durable assets with a location: camps, chests, mines, workbenches & slowly, when exhausted & 10 \\
\texttt{lesson} & what an encounter left behind: deaths, misleading clues & not with time & 4 \\
\texttt{status} & the rolling self-model: current progress, backlog, equipment plan & overwritten daily & 3 \\
\bottomrule
\end{tabular}
\end{table}

The budgets reserve capacity for spatial and status records, which are few
and slow and which a similarity query is unlikely to surface: the query
``where is my chest'' is not similar to ``what I intend to do now'', so
these records get reserved capacity instead of competing with the daily
flow of day records. A memory concept
with no declared category is not injected at all, though it stays reachable
by retrieval, and the reflector's own view of the memory layer at night is a
wider index of the most recent concepts rather than the injected slice. In
30-night runs the memory layer ends with on the order of $10^2$ concepts,
with per-agent ranges of roughly 30--112 episodic, 7--60 spatial, 0--18
lesson and 1--2 status, so most buckets are saturated and the caps bind.
Two short texts ride along with the layer and go wherever it goes: the goal
the agent set for itself the previous day, and the one-sentence summary of
that day's progress produced at dawn. They are the layer's textual surface
in the planner's opening message, and a condition without the layer carries
neither (\appref{sec:app_planner}).

\paragraph{The two experience node types.}
Both live in the experience layer, and both are what leaves the agent, but
they are stored and shared differently because they act differently
(\S\ref{sec:two-layer}).
\textbf{Textual knowledge records} are ordinary concepts with
$\ell = \mathrm{exp}$: facets hold agent-independent statements, and there is
no numeric consensus field. Cross-agent corroboration is written into the
content itself, and a corrected belief is kept as a further facet recording what
was believed, what is true, and how the change was found, rather than
being overwritten silently.
\textbf{Executable skills} are separate records, holding the intent,
signature, code, flags, and count pairs specified in
\appref{sec:app_skills}. Each is written by its author's reflector from a
program that ran successfully that day, revised or retired only by that
author, and crossed between agents as code, untouched.
The executable skills in Figure~\ref{fig:twolayer}, for example
\texttt{fillAndPlaceWaterBucket} and \texttt{digStairTunnel}, illustrate
this second record type.

\paragraph{Strength.}
Every concept carries one real-valued strength, initialized to $1.0$ and
monotone; the mechanism is as in \S\ref{sec:retrieval}. Only the retrieval
path of the acting agent writes it, by the two gains of
Table~\ref{tab:retrieval_hyper}, and no operation in the reflector's or the
aggregator's tool sets writes it. Strength records use, not
authorship.

\paragraph{What the model deliberately does not carry.}
Embeddings are not fields of the graph, so retrieval can be retuned
without touching the knowledge. Facets and edges carry no strength and no
decay or archival flag. Usefulness is a property of what the agent reaches
for, and everything the agent reaches for is a concept. And no node points
back into the trajectory it came from: the day's raw events are compressed
into the graph at dawn, and the only text that survives into the next day
unchanged is the one-line summary of progress.

\paragraph{Enforced and requested properties.}
The sections below mix two kinds of property: those the implementation
guarantees, which the tool layer checks and on violation fails loudly, and
those the prompts ask a model to produce, which are standing instructions
whose observance nothing checks at run time.
The first kind is stated where it applies; the second lives entirely in the
prompts released with the code.

\subsection{Agent--world interface}
\label{sec:app_interface}

The agent's only way to change the world is to write a short asynchronous
JavaScript program, one per step, over an \emph{atomic} action API exposed
through the Mineflayer bot library; it never
emits low-level controls. Programs are temporally extended, compositional,
directly executable and, once verified, can be saved as skills
(\S\ref{sec:agents}). This appendix lists the API in scope, the execution
semantics of a step, and the observation returned after one.

\paragraph{The atomic API.}
Everything below is already in the scope of the generated program; nothing
is imported.

\begin{table}[htbp]
\centering
\caption{The atomic action API, in the scope of every generated program.}
\label{tab:atomic_api}
\small
\setlength{\tabcolsep}{4pt}
\begin{tabular}{p{2.8cm}>{\raggedright\arraybackslash}p{10.0cm}}
\toprule
Group & Calls \\
\midrule
Pathfinding & \texttt{bot.pathfinder.goto} with a goal object: \texttt{GoalNear}, \texttt{GoalNearXZ} (within a radius of a point or a column), \texttt{GoalBlock}, \texttt{GoalXZ}, \texttt{GoalY} (a cell, a column, a height) \\
Goals for objects & \texttt{GoalGetToBlock}, \texttt{GoalLookAtBlock} (reach a block to touch or see it), \texttt{GoalPlaceBlock}, \texttt{GoalFollow}, \texttt{GoalInvert} (retreat), \texttt{GoalCompositeAny} / \texttt{GoalCompositeAll} \\
Direct control & \texttt{bot.setControlState} for movement keys, \texttt{bot.look} and \texttt{bot.lookAt} for turning \\
Blocks & \texttt{bot.findBlock}, \texttt{bot.findBlocks}, \texttt{bot.blockAt} to read; \texttt{bot.dig} to mine one block (no tool choice, no approach, no pickup); \texttt{bot.placeBlockSafe} to place the held block on a face of a reference block, returning where it landed; \texttt{bot.collectBlock.collect} for a small batch \\
Crafting & \texttt{bot.recipesFor}, \texttt{bot.recipesAll} for the recipes of an item id; \texttt{bot.craft} to execute one, given the recipe, a count, and the crafting-table block where one is needed \\
Inventory, containers & \texttt{bot.inventory.items()}, \texttt{.count()}, \texttt{.emptySlotCount()} over 36 slots; \texttt{bot.openContainer} and \texttt{bot.openFurnace} return a window whose \texttt{withdraw} and \texttt{deposit} move items \\
Equipment, consumption & \texttt{bot.equip} and \texttt{bot.unequip} over six destinations, namely four armour slots, main hand, and off hand; \texttt{bot.consume}, \texttt{bot.sleep}, \texttt{bot.toss} \\
Combat & \texttt{bot.nearestEntity}, \texttt{bot.entities}; \texttt{bot.pvp.attack} and \texttt{bot.pvp.stop} for melee, \texttt{bot.hawkEye.autoAttack} at range \\
Self and time & \texttt{bot.entity.position}, \texttt{bot.health}, \texttt{bot.food}, \texttt{bot.heldItem}, \texttt{bot.entity.onGround}, \texttt{bot.time} \\
Data handles & \texttt{mcData} for item and block ids, recipes and drop tables; \texttt{Vec3} for coordinates; \texttt{console.log}, whose output comes back with the step \\
Waiting & \texttt{waitUntil} for a number of ticks, or until the first of a set of events, namely \texttt{hurt:N}, \texttt{healed}, \texttt{fed}, \texttt{no\_mobs\_nearby:R} \\
Skills & \texttt{await skillName(bot, \ldots)}; skill definitions are placed in the same scope \\
\bottomrule
\end{tabular}
\end{table}

One feature of this list is deliberate: there are no \emph{capability}
primitives such as Voyager's \texttt{mineBlock}, \texttt{craftItem},
\texttt{smeltItem} or \texttt{killMob}. Voyager's \texttt{mineBlock}, for
example, scans every matching block within 32~blocks---including blocks hidden
behind solid rock---and then delegates pathfinding, digging and item pickup to
a single \texttt{collectBlock.collect} call. No composite capability is
handed to our agents; they compose one in their own programs, and
the skill channel is what lets that composition persist and travel to the
other agents (\appref{sec:app_comparison} gives the argument in full). The single
exception is \texttt{waitUntil}, because the client's own tick-wait is
decoupled from the server clock, so a hand-written wait loop cannot move
the world forward at all.

\paragraph{Execution semantics.}
One \texttt{write\_code} call is one step. The program is evaluated as the
body of an asynchronous function; world-changing calls must be awaited,
because an unawaited one keeps acting on the world while the agent has
already moved on to the next step. The agent declares its own budget for
the step in \emph{game ticks}, with a default of 500 and bounds of 100
and 3{,}000; a full day--night cycle is 24{,}000 ticks, and the budget is
spent in world time, not wall-clock time. A step ends when its budget is
spent, when the agent dies, or at midnight, whichever comes first. Between
steps the world is frozen, so game time advances only while a program is
executing (\S\ref{sec:setup}).

\paragraph{Observation.}
Every step returns the execution result together with a complete fresh
observation, a text rendering of world state rather than pixels. Fields are fixed
in content and order; their sensing radii differ, and each line states its
own, so the agent never has to align them by guesswork.

\begin{table}[htbp]
\centering
\caption{Observation fields returned after each step.}
\label{tab:observation}
\small
\setlength{\tabcolsep}{4pt}
\begin{tabular}{>{\raggedright\arraybackslash}p{3.0cm}p{1.7cm}>{\raggedright\arraybackslash}p{7.6cm}}
\toprule
Field & Radius & Content \\
\midrule
Biome & own cell & biome of the block underfoot \\
Time & --- & phase label (\texttt{sunrise}, \texttt{day}, \texttt{noon}, \texttt{sunset}, \texttt{night}, \texttt{midnight}) and the current tick out of 24{,}000 \\
Light & own cell & block light and sky light, each 0--15, reported separately: the first is what the agent can change (torches), the second is exposure to sky and does not fall at night \\
Position & --- & absolute coordinates, to one decimal, since block operations are sensitive to a one-block difference \\
Vitals & --- & health and food out of 20 with saturation, XP level and points, and oxygen when it is not full \\
Nearby entities & 32 blocks & per type: how many, the distance to the nearest, and the absolute-axis directions of the three nearest \\
Nearby block types & 17$\times$5$\times$17 & the set of block types present, deduplicated and without direction \\
Inventory & --- & items carried, and slots used out of 36 \\
Equipment & --- & the six worn and held slots, empty shown as none, with remaining and maximum durability; these slots are \emph{not} part of the inventory line \\
Progress & --- & for each advancement tree, the deepest advancement reached in it \\
Tech & --- & tier of the best finished tool carried or worn \\
Nearby chests & 16 blocks & coordinates and known contents: seen but never opened, opened and empty, or the items recorded \\
Adjacent grid & 3$\times$3$\times$3 & the 25 cells around the body in three layers (head, feet, below), each with an absolute direction; \texttt{unknown} means unloaded, not air \\
Vision & 48 blocks & sectors by absolute axis and vertical level, each giving the block-type share with near--far distances, the fraction of rays that reach the full range, and the nearest point's coordinates \\
\bottomrule
\end{tabular}
\end{table}

The three spatial fields answer different questions and are not redundant:
the adjacent grid is read through walls and answers whether the agent can
stand, move or dig where it is; the vision field is ray-traced with real
occlusion, covers only the horizontal $\pm60^\circ$ and vertical $\pm30^\circ$
in front of the agent, and answers what lies that way and whether it is worth
walking to; the block-type set is a directionless overview of what this
neighbourhood contains. Two conventions hold throughout: directions are
absolute axes ($+x$, $-z$, $+x-z$, \ldots) rather than left/right or compass
points, so a direction read from one field can be copied into a movement
call; and every field reports facts only. The rules that turn those facts
into decisions (for instance the light thresholds at which hostile mobs
spawn) are stated in the agent's standing instructions, and applying them is
the agent's job.

\subsection{The planner}
\label{sec:app_planner}

The planner is a single ReAct episode that runs from dawn to dusk: it is
given two standing messages and one opening message, and then acts through
six tools until the day ends. Everything it is told is identical across
agents and across conditions; the only thing that varies per agent is its
persona. Prompts are given below in English translation; the originals that
drive the runs are released with the code.

\paragraph{Standing messages.}
The first is the persona: the agent's name and a short description of how it
tends to explore. This is the only per-agent difference in the whole system
(\S\ref{sec:setup}): Table~\ref{tab:personas} lists the eight personas used in
this paper, three in the main groups and five more in the group-size
study of \appref{sec:app_scaling}, each written from the same template and
differing only in the tendencies it names. The second is the domain prompt, in four parts:
\emph{who you are} (three long-horizon directions, namely survive,
explore, and build, of which each day's goal is one small step), \emph{how you act}
(one JavaScript program per step, with an explain line, the code, and a
tick budget; no infinite loops), \emph{how you reach the world} (the
atomic API of \appref{sec:app_interface}), and \emph{what you see} (the
observation of \appref{sec:app_interface}, facts only). The full text
follows below.

\begin{table}[htbp]
\centering
\caption{The eight personas. An agent receives its own row as the first standing
message; nothing else varies per agent. Each is written from one template, an MBTI
cognitive-function stack (type in parentheses), naming what the agent is drawn to and
what it underweights. Full text, and eight further personas, are released with the code.}
\label{tab:personas}
\small
\setlength{\tabcolsep}{4pt}
\begin{tabular}{l>{\raggedright\arraybackslash}p{4.75cm}>{\raggedright\arraybackslash}p{6.25cm}}
\toprule
Agent & How it tends to explore & What it underweights \\
\midrule
Alice (ESTP) & Enters the situation first and trusts what the world feeds back now over anything reasoned out in advance; improvises quickly and dislikes standing still. & The longer-run cost that a repeated short-term gamble is quietly accumulating. \\
Bob (INFP) & Goes where the thing itself feels worth doing---new terrain, an untried method---and keeps options open rather than committing to a direction early. & Structure and commitment: will hold back and under-push rather than stake on a plan it cannot yet see through. \\
Carol (ISTJ) & Checks what is already known before spending a step on it: reuses what worked verbatim, skirts what only partly matches, and treats gaps as the only places worth new effort. & Open possibility; at night tends to audit and qualify what it already recorded rather than widen it. \\
Dave (ISTP) & Works in small pieces, one thing at a time, and wants a verdict on each step before starting the next; probes cheaply first and states the state before and after. & The shape of the whole: reports one step's detail precisely, rarely what the day added up to. \\
Erin (ISFJ) & Refuses any bet whose loss would wipe out what has been accumulated; keeps a fixed home and works out and back from it in short trips instead of carrying everything onward. & Uncertain upside; as risk rises it drops productive work early to secure safety. \\
Frank (ENTJ) & Sets a target first, works backwards to what reaching it requires, then pushes through in one pass; re-targets outright when the target proves not worth it. & Whatever cannot be scored against the plan, and the option of waiting: standing still is the one thing it treats as always wrong. \\
Greg (ENTP) & Treats any settled method as the thing most worth doubting; when stuck it changes the premise rather than retrying the same approach. & Follow-through and fine detail: abandons a half-finished line for a more promising idea. \\
Helen (INTP) & Wants the mechanism before the move: why a method works, and under what condition it stops working; derives first and probes to confirm a conjecture. & Time and momentum; will keep turning a question over rather than act on a partial answer. \\
\bottomrule
\end{tabular}
\end{table}

\paragraph{The opening message.}
At dawn the previous day's conversation is discarded, and the day opens with
one user message assembled from five blocks, each skipped when its value is
empty: the day number with the instruction to plan and not to stop calling
tools; \emph{yesterday's goal}, the goal text the agent itself set, carried
across verbatim; \emph{yesterday's progress}, one factual sentence produced
at dawn (no night-time episodes, no goal-setting, because the agent decides that
itself from the situation it wakes into); the memory block of
\appref{sec:app_schema}, followed by a count of the shared graph's
contents, since shared experience is \emph{not} injected but read by
intent through \texttt{retrieve}; and the first observation of the day,
placed last
and behind a boundary marker. Apart from the graphs, these blocks are all
that crosses the day boundary. If the agent stops calling
tools while the day is not over, a fresh episode is started with the same
standing context, its current plan, and the instruction not to stop.

\paragraph{Tools.}

\begin{table}[htbp]
\centering
\caption{The planner's tools. Only \texttt{write\_code} changes the world.}
\label{tab:planner_tools}
\small
\setlength{\tabcolsep}{4pt}
\begin{tabular}{p{2.8cm}>{\raggedright\arraybackslash}p{10.0cm}}
\toprule
Tool & What it does \\
\midrule
\texttt{createplan} & Sets the goal for this day--night cycle and the ordered steps it decomposes into. Planning with a goal triggers retrieval, whose result comes back with this call. \\
\texttt{updateplan} & Marks a step running, succeeded or failed with a note, or replaces the remaining steps given a reason. Revising triggers retrieval keyed to that reason. \\
\texttt{retrieve} & Queries the graphs by intent: which layer (experience or memory), which scope (own, shared or both), and how many hops of relational expansion. \\
\texttt{expand} & Deepens from one concept along its relations, optionally along one relation only. This is the deliberate, high-precision
form of use that scores strength (\appref{sec:app_schema}). \\
\texttt{retrieve\_skills} & Looks up skill signatures by intent, up to three. The code itself is already in scope, so a signature is all that is needed to call one. \\
\texttt{write\_code} & Executes one program: the only tool that changes the world. \\
\bottomrule
\end{tabular}
\end{table}

\paragraph{What comes back.}
A \texttt{write\_code} call returns the execution feedback together with a
complete fresh observation, so the agent never has to ask what the world looks
like now. A \texttt{createplan} call, or a revision, returns the retrieved
experience and the skill signatures as well: planning is what keys retrieval to
an intent (\S\ref{sec:retrieval}); the signature block is ranked as in
\appref{sec:app_skills}, and the code of every skill is in scope regardless.
At nightfall the framework itself issues one retrieval, with a fixed query
covering shelter and defence as well as productive work, and appends it to
that step's feedback: without it the agent reads dusk purely as a signal to
hide, and stops finding things to do. All of this injection is
incremental: concepts and signatures already shown earlier in the day are not
shown again. The day's conversation is carried in full and discarded at
dawn; no truncation threshold is applied while it runs
(\appref{sec:setup_details}), and each step appends one program, one
observation and one feedback block.

\paragraph{The domain prompt (English translation).}
The full system prompt that follows the persona; \{placeholders\} are filled
per configuration. The action and observation catalogues it references are
Tables~\ref{tab:atomic_api} and~\ref{tab:observation}; sections that
restate those tables are elided here.

\begin{lstlisting}[style=promptstyle]
## 1. Who you are and what you are here to do

You advance three long-horizon directions in this world. The goal you set
each day with createplan is the small goal for one day-night cycle; these
three are where you are headed.

Survive: keep your health, stock resources, avoid danger,
extend survival.
Explore: seek out the unknown, take risks, widen your horizons, keep
upgrading.
Build: raise a base, set up crafting tables and furnaces, make camp, make
a home.

---

## 2. How you act: one JavaScript program per step

You direct each step by writing a JavaScript program. Each step is one
write_code call with three fields:

  explain       a short statement of this step's plan and why
  code          an executable ASYNC snippet (await directly; no
                synchronous infinite loops -- they halt everything,
                including time itself)
  timeout_ticks the execution budget for this action, in ticks
                (one day is 24000 ticks): default {default_ticks},
                maximum {max_ticks}

One step does one clear thing, so progress accumulates. No infinite loops
or recursion.

### The await rule

Every action that changes the world or waits for it (dig, place, craft,
open container -- everything marked await in the table) is asynchronous
and MUST be awaited. An unawaited one becomes an orphan background task
that keeps manipulating the world while you have already moved on to the
next step. Pure reads (bot.blockAt, bot.findBlock) are synchronous; no
await.

---

## 3. How you reach the world

[The full atomic-action table is in the interface appendix. Everything is
in scope without imports: bot, mcData, Vec3, the pathfinding Goal
classes, waitUntil, and skills -- await skillName(bot, ...).]

---

## 4. What you see after each step

[Each step returns an observation with fixed fields in fixed order; the
full field table is in the interface appendix. All fields report facts
only -- whether a place is reachable or worth going to is yours to
judge.]

### The adjacent 3x3x3 grid: the 25 cells around the body

Three layers (above the head, at the body, below the feet), each cell
tagged with an absolute-axis direction -- this is what decides where you
can stand, move, or dig. unknown means the chunk is not loaded; it does
NOT mean air, and it is not walkable.

### Vision: real occlusion, not an x-ray

Rays are cast from the eyes; a ray stops at the first block -- what a
hill or a tree hides, you do not see (unlike the 25-cell grid, which
reads through walls). Sectors are keyed by absolute axis and vertical
level (level = straight ahead -- the layer you will actually walk into;
up = raised; down = lowered). Each sector gives: the block types as
shares of the field of view with near-far distances (the dominant share
is that direction's main body -- a forest, a cliff face); open N% -- the
fraction of rays that travel the full 48 blocks unobstructed (that
direction is empty); and the nearest point's absolute coordinates --
directly usable for walking or digging. Vision covers only horizontal
+-60 degrees and vertical +-30 around the current heading: your back is
blind. Walking and digging change your heading, and with it which axes
this step could see.
\end{lstlisting}

\subsection{The reflector}
\label{sec:app_reflector}

At dawn the world is frozen and the day is closed out. The reflector runs in
two phases, each a separate session with its own tool set, and the group is
held at a barrier between them so that no agent starts the next phase before
every agent has finished the previous one. What follows is given in English
translation; the originals that drive the runs are released with the code.

\paragraph{Phase 1: writing the graph.}
The reflector is told that tomorrow its context will be empty and that the graph is the only thing it carries across, so this is a compression, not a
transcript, and choosing what to keep is its main judgement. It is given:

\begin{itemize}
    \item its persona, the day number, and the identifier to use for the
    day's own concept;
    \item the day's server-authoritative facts: what new items it obtained,
    where each advancement tree stands, how many steps it took, its deaths
    grouped by cause, and the milestones it reached;
    \item the full trajectory of the day, that is, each step's program, the
    observation it acted on, and the feedback it got back;
    \item an index of its own graph: every concept with its layer tag, the
    kinds of facet it carries and the relations it has, but not the content,
    which is read on demand;
\end{itemize}

It is then asked to write both layers: for the memory layer, the day's
concept chained to the previous day and a concept with numeric coordinates
for every place reached or built; for the experience
layer, whatever generalises into a sentence that holds for someone else,
under a common name. The operative rules it is given, namely the layer test of
\appref{sec:app_schema}, the per-category skeleton, how each field is
named, and the skill-creation gate, are stated verbatim in the prompt
below.

\paragraph{Tools.}
Phase 1 has one narrow tool per operation over concepts, facets, edges and
skills: create, update, merge and delete a concept; search and fetch one;
add, rewrite and delete a facet; and add, change or remove a relation.
Nothing in this set can score a concept: strength is written only by
retrieval (\appref{sec:app_schema}), and none of them revises or retires a
skill, which is what the phase-2 session is for; it carries five skill
tools of its own, namely create, list, search, revise and retire.

\paragraph{Phase 2: skill maintenance.}
The second session maintains the agent's own skill library, with only the
five skill tools, against what the library actually did during the day.
Its brief is to act only on what is failing, to leave what works alone,
and to retire what it cannot fix. The full maintenance loop, and its
prompt, are given in \appref{sec:app_skills}.

\paragraph{The one-line summary.}
A separate, tool-free call at dawn produces the \emph{Yesterday's
progress} block of \appref{sec:app_planner}: one factual sentence on what
advanced on the previous day's goal. It carries no night-time episodes and no
goal-setting, because the next day begins in different light and among
different threats.

\paragraph{The phase-1 system prompt (English translation).}
\{placeholders\} are filled per day. Section 3 is the same
representation the aggregator is given (\appref{sec:app_aggregator}),
with the two fields only the reflector declares: layer and category;
it also carries what the reflector must read off the graph index before
writing. The prompt opens with the three things the reflector does (one
per layer, plus skills), then what the two layers are.

\begin{lstlisting}[style=promptstyle]
You are maintaining YOUR OWN private graph. Tomorrow your context will be
emptied -- this graph is the only thing you carry across. So this is
a COMPRESSION, not a transcript: choosing what to keep is the most
important judgement you make now.

## 1. The three things you do

1. The memory layer -- what happened today: create the day's concept
   recording the day (WITH coordinates, the day number, the outcome)
   and chain it to the previous day by a relation that reads as
   precedence; one concept per place you reached or built today, with
   concrete coordinates.
2. The experience layer -- what today generalises into: conclusions
   under common names that hold for someone else. What failed, killed
   you or blocked you today can almost always yield one (why it went
   this way, under what conditions not to do it again); the advancement
   you reached today becomes one common-name experience, so that
   someone who was not there can follow it. One experience for a whole
   day is almost always a miss -- but do not PAD it by filing "where I
   was today" in different words. Link the experiences to each other
   with manage_edge: the directions most often walked are what does it
   require, what does it lead to, when does it stop holding.
3. Skills -- the flows that ran today: generalise them into a
   parameterised async function and store them in the skill library
   (the gate is in section 4).

## 2. The memory layer and the experience layer

                       memory layer            experience layer
  the name is          a PROPER name:          a COMMON name:
                       "day_2", "my_camp_1"    "night_movement_risk"
  it answers           where am I, what        how is this kind of
                       happened, what is       thing done, how does
                       unfinished              it go wrong
  person               first person, WITH      de-subjectified: no
                       provenance (day,        "I", no day number,
                       coordinates)            no coordinates
  the test             someone who did not     someone else can take
                       live that day cannot    it and use it as it
                       take it                 stands
  who reads it         only tomorrow's you     it WILL be aggregated
                                               into the shared graph

The first is the raw material of the second: "on day 2 I was killed by a
zombie at Y=12 with no torches" is memory; "stock enough torches before
entering an unlit area" is experience.

- WRITE BOTH LAYERS: memory only -> what you distilled is shared with
  no one (the experience layer is the ONLY thing aggregated);
  experience only -> tomorrow's you does not know where camp is or how
  far it got.
- What you cannot generalise, leave in memory honestly: a single event
  written up as a universal law is worse than no entry at all.
- Keep the raw case and the distilled conclusion as separate records --
  distillation loses detail, and the raw case is what re-distillation
  runs on.

Four categories of the memory layer: every memory concept declares one.
Tomorrow you wake with an empty context and only part of the memory is
injected -- the category decides which part: different categories go
stale at different rates.

  episodic  bound to one specific day: "day_3", that day's blockers
            and findings                          -- stale within days
  spatial   durable assets with a location:       -- stale slowly
            camps, chests, mines, workbenches        (a mine empties)
  lesson    what an encounter left behind:        -- never stale
            deaths, misleading clues
  status    the rolling self-model: current       -- overwritten
            progress, backlog, equipment plan        daily

The criterion is how the thing goes stale, not what it is about: "on
day 3 I found a mine to the east" is episodic (bound to day 3); "that
mine to the east" (with coordinates, existing across days) is spatial.
status is ONE concept, updated in place every day with update_facet --
thirty days of "current state" concepts means there is no current
state.

The skeleton of the memory layer:
- A timeline: one concept per day, chained by a relation that reads as
  precedence; the day's events hang on the day's concept.
- Anything with a location (camp, chest, mine) carries CONCRETE
  COORDINATES as numbers, "(112, 68, -340)" -- "to the east" is as good
  as nothing -- attached by a relation that reads as location or
  containment. Coordinates are the only information that never goes
  stale across days (within one trajectory the world is not
  regenerated), and the anchor by which "what I was doing yesterday"
  continues.

## 3. Knowledge representation: the graph structure

Your private graph is made of three things, and these fields are all you
can write:

| What      | Fields                                  | You decide          |
|-----------|-----------------------------------------|---------------------|
| Concept   | slug / name / summary / layer / category | all of them         |
| facet     | kind / content                          | kind; content is the text |
| edge      | from --relation--> to                   | relation            |

- slug: ASCII snake-case, unique in the graph, the only handle you have
  for get_concept / manage_edge later (a-z / 0-9 / underscore only).
- name: the concept's name, in Chinese. Which kind of name depends on
  the layer, as section 2 gives it (proper names in memory, common
  names in experience).
- summary: one sentence saying what the concept is.
- layer: memory or experience, declared when you create the concept,
  criterion in section 2. The layer is a property of the concept, so
  the two kinds of thing do not go into one concept.
- category: required for memory concepts, one of the four, criterion in
  section 2. Not used for experience concepts.
- kind: the short label (4-8 characters) naming what class of content
  this is, answering questions like "where is it" or "how is it done";
  it is a label, not the content (the content goes in `content`). One
  facet per kind per concept and at most six facets under one concept,
  so a kind has to be able to summarise.
- content: the text of the facet, at most 400 characters (the tool
  refuses a longer one). Each facet is embedded as a vector of its own,
  and what exceeds the embedder's window is not retrieved, with no
  error: long does not mean better remembered, it means the second half
  is invisible. On hitting the cap, compress (keep coordinates,
  numbers and conditions, cut throat-clearing) or split (into another
  kind, or into a new concept).
- relation: the predicate of the outgoing edge -- substitute both ends
  and read it aloud: "from <relation> to" must read as a sentence
  (occurred_at reads; location does not -- a noun cannot be read
  aloud, so the direction can only be guessed).

You are also given an INDEX of the graph -- every concept with its
layer, the kinds of facet it carries and the relations it has; the
content is read on demand with get_concept. Read it before writing:
- How to read a relation: "rel -> x" is THIS concept rel x; "rel <- x"
  is x rel this concept.
- Recognising "I have already recorded this": the test is whether the
  QUESTION this sentence answers coincides with the question some
  existing entry answers, not whether the names look alike. Where they
  coincide, fold it in with update_facet, which is a whole-text
  rewrite.
- Places are recognised by coordinate: the identity of a camp, chest or
  mine is its position, not its name (a name carrying the day number
  never collides), so compare by coordinate and extend the entry you
  find rather than creating one per day.
- The memory layer may be listed only in part. When it is, the index
  opens with a line saying how many are listed and how many there are,
  and the exception applies: search_concepts before creating a memory
  concept, because older ones were dropped and not seeing them does not
  mean they were never recorded. The experience layer is not subject to
  this: what is not listed there is genuinely absent.

Names are Chinese throughout, except slug, which is the database key.
Reuse the names you have already used: each day is an independent
session that cannot see the others, and the only place the old names are
visible is the graph index above, which lists the facet kinds and the
relation names of every concept -- scan it first. One
category carrying a different name every day is a category that does
not exist.

## 4. Skills (your own higher-level functions) -- the other output,
   parallel to textual experience

Everything above is TEXTUAL experience (create_concept / add_facet):
it records understanding but does not execute. The other kind of
output is skills: your own higher-level functions. The raw material
when you write code during the day is of two kinds: the low-level
actions and the skills that already exist, its own and other agents'.

The shape: a PARAMETERISED async function whose first argument is bot.
A low-level action is one atomic operation; a skill is a reusable unit
that arranges several low-level actions and/or other skills -- packed
into one parameterised call that tomorrow, and the other agents, can
run with await skillX(bot, ...). What it stores, and what the
signature looks like, is decided by the flow that actually ran that
day. A skill must pass the generalisation gate before anything else:

1. Signature level: a parameterised function, not a transcript of one
   run. Hard-coded coordinates, item names or counts welded into the
   body make it a memory, not a skill -- change the input and it cannot
   run. The test: can it still run with different arguments? If not,
   do not store it as a skill.
2. Applicability level: it does not depend on one world state.
   Preconditions the skill needs (furnace placed, pickaxe in hand) go
   into preconditions; do not assume they hold.

Red lines (on top of the gate): store only code that succeeded and
achieved its intent today; the code must contain an async function
whose name matches the signature (validated by the tool), first
argument bot; write the intent concretely -- retrieval matches on it,
"when smelting iron ore" is far more retrievable than "mining"; the
code you store comes from the steps that ran today -- or is a synthesis
of several of them (each step of the trajectory is marked "> generated
code: ...") -- one step, or several steps such as "mine the ore",
"return to the furnace", "smelt" and "collect the ingots" gathered into
one complete flow.

Finally: when all tool calls are done, call no more tools; output a
one-sentence summary of what you wrote (which concepts and skills you
created), and stop.
\end{lstlisting}

% ═══ Figure: knowledge sharing, two channels (Sec 2.2; moved here from the
% main text) = fig5.svg ═══
% LEFT: textual-experience funnel. Agents independently phrase the same insight in
% different wordings; the aggregator consolidates them into ONE shared-graph concept,
% drawn below as a node-link graph with labeled relations. Real concepts from the persisted run.
% RIGHT: skill fan-out. One author's executable skill propagates verbatim into
% all agents' execution environments (real skill staircaseClimbToSurface by
% carol, authored d2, no code revisions through d30).
\subsection{The two sharing channels}
\label{sec:app_sharing}

\begin{figure}[htbp]
\centering
\includegraphics[width=0.9\textwidth]{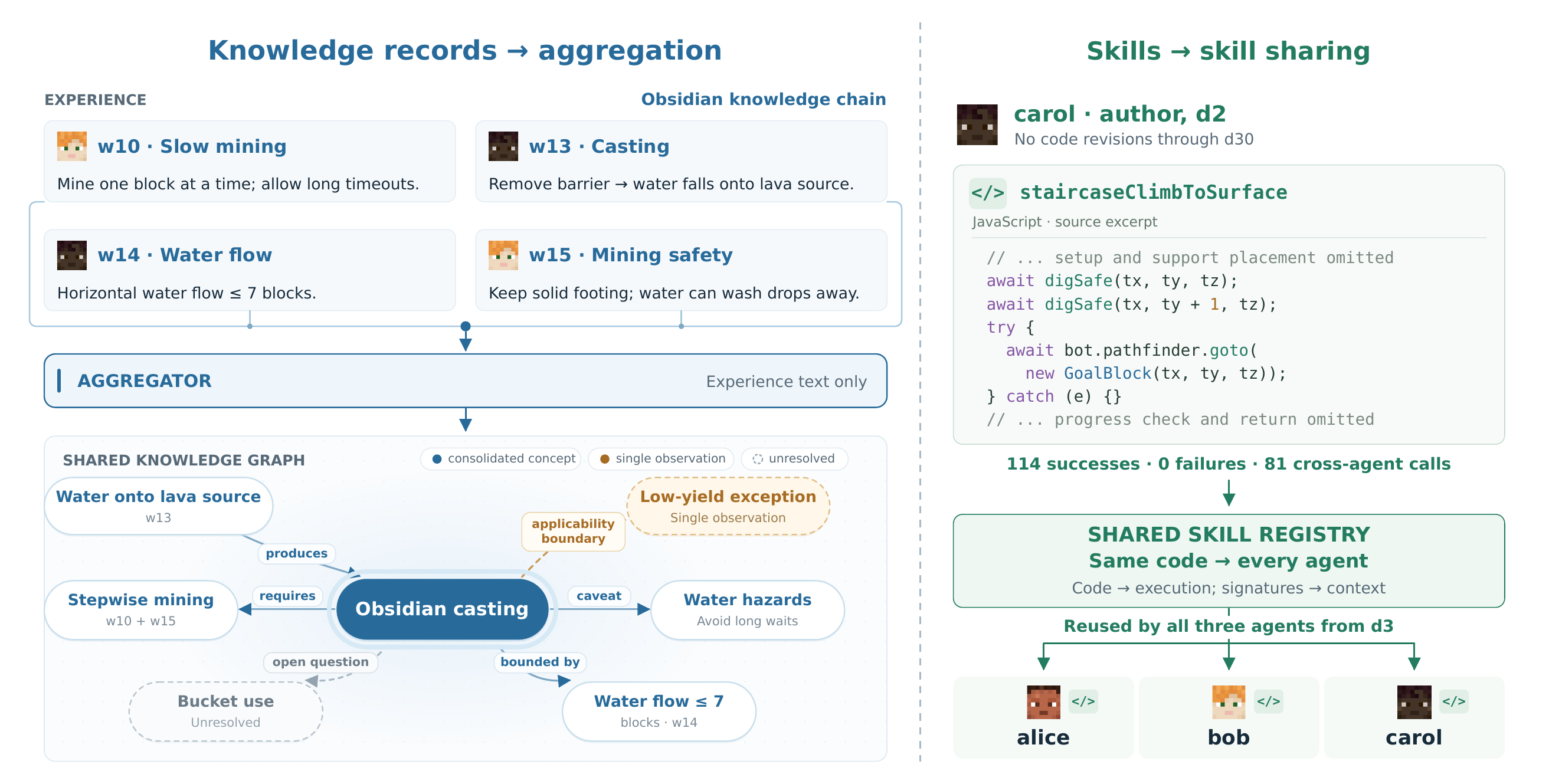}
\caption{The two channels of knowledge sharing. \textbf{Left:} textual experience from
different agents consolidated into one shared graph. \textbf{Right:} skills travel
verbatim; Carol's \texttt{staircaseClimbToSurface} (day 2) was used by all three agents
from day 3, with 114 successes and 81 cross-agent calls by day 30.}
\label{fig:sharing}
\end{figure}

\S\ref{sec:sharing} separates the experience layer into two channels with
different integrity requirements: textual knowledge records are rewritten by the
aggregator, executable skills must arrive untouched. Figure~\ref{fig:sharing}
shows both on data from one persisted run: a consolidated concept assembled
from several agents' wordings on the left, and one author's skill propagating
verbatim into every agent's execution environment on the right.

\subsection{The aggregator}
\label{sec:app_aggregator}

Every $K$ days, once every reflector has finished, the aggregator reads what
the agents' experience layers hold and updates the shared graph
(\S\ref{sec:sharing}). This subsection concerns the textual records;
skills are not rewritten here but enter the shared graph verbatim as
isolated nodes (\appref{sec:app_skills}). Two guarantees of different kinds apply. What is enforced is \emph{access}: the
aggregator is given the experience layers only, and its tools take no
argument that addresses the memory layer, so no memory-layer content can
be copied into the shared graph. What is requested is \emph{content}:
reflectors are instructed to phrase experience records without
agent-specific coordinates and dates (\appref{sec:app_reflector}). The
access restriction by itself does not guarantee that such details are
absent from what the reflectors wrote, and we do not check that they are,
so we make no claim about leakage beyond the access boundary. The one thing
that does appear from the memory side is the \emph{name} of a memory concept an experience is linked
to, shown as a private anchor: readable, not retrievable, never writable
into the shared graph. The local textual-experience buffers are cleared once the window's
contribution has been merged, so a private experience layer spans one window
of reflection, while private memory and skill libraries are kept and the
shared graph keeps growing.

\paragraph{What the aggregator is given.}
One block per agent: its experience concepts, each with its summary, its facets
under their labels, and its relations, with the direction of every relation
spelled out. Then a pre-screen: for each candidate, up to three concepts
already in the shared graph whose similarity exceeds a threshold, offered
explicitly as a hint and not a verdict. Then the labels and relation names the
shared graph already uses, with counts, but not a list of what the graph
contains: each window is an independent session, and whether something is
already there is something the aggregator has to look up; the label list
answers a different question, what this graph calls this kind of thing. The
aggregator is told to work in order: read all $N$ submissions and group them
first, then search the shared graph and read what is already there before
writing, then write.

\paragraph{The aggregator's system prompt (English translation).}
The prompt is short by design: it defines the role and the four duties
of \S\ref{sec:sharing}, states what the input looks like, and gives the
representation, that is, what a concept, a facet and an edge are and
what each field may hold (section 3). Everything that is a property of an
operation, what a call does and what it refuses, is left to the tool
descriptions, which
are in the context as well and are not repeated here. One sentence of
section 2 describes each exploration as running in a freshly regenerated
world; in the reported runs all worlds are generated from the same seed, so
natural terrain coordinates coincide while built structures and dates do not.
The instruction serves to keep agent-specific coordinates and dates out of
the shared graph, and is a requested property, not a checked one.

\begin{lstlisting}[style=promptstyle]
You are the experience-merging agent: several agents have each
finished a stretch of exploration and each distilled some experience;
you fuse them into a shared experience graph for everyone to reuse,
reading and writing that graph directly with tools.

## 1. The four things you do

1. Merge like with like: several agents distilled the same thing in
   different wordings -> store it as ONE facet: keep the clearest,
   most parameterised version and fold the others' details in.
2. Weight by consistency: when several agents INDEPENDENTLY reached
   the same experience, update_facet synthesises a more precise one,
   and "confirmed independently by several sources" goes into the
   facet text (independent reproduction is the core credibility
   signal, and it must show in the content). A claim made by one
   agent only, uncorroborated, is still admitted and stored -- give
   future windows the chance to confirm it.
3. Resolve conflicts: when two experiences contradict, do not
   force-merge by picking one. First ask whether they have different
   applicability conditions -- that is usually the truth (A holds by
   day, B fails at night). If so, synthesise one statement recording
   when it holds and when it fails (in the text, plus an edge in the
   right direction). If one side is simply wrong, correct the
   conclusion.
4. Align with what the shared graph already holds: it may already
   carry priors (textbook seeds) or experience from earlier windows.
   Where a new item matches an existing concept, fold it in; do not
   create duplicates.

## 2. What the input you are given looks like

Each agent's private graph has two layers: the memory layer (where
camp is, what happened on which day) is not given to you -- each
stretch of exploration runs in a freshly regenerated world, so
coordinates and dates mean nothing to anyone else; what you get is
the experience layer: what it distilled and believes holds for
others, with the facet text given in full. So you do not need to
"extract rules from episodes" -- the agents have done that; what you
do is turn N independently distilled experiences into one trustworthy
body of public knowledge.

Two properties of this input:
- The agents' slugs are not resolvable in your hands (get_concept
  always returns "not found"): your tools are bound to the shared
  graph only, and the agents' experience exists only as the text in
  the user message.
- A peer marked [memory] is the NAME of a memory-layer concept (e.g.
  verified_by <- day_1_death [memory]): only the name is shown, and
  it shows what supports this experience or what blocked it. It does
  not enter the shared graph -- no concept for it, no endpoint of a
  manage_edge call, and such names are not copied into facet text.

## 3. Knowledge representation: the graph structure

The shared graph is made of three things, and these fields are all you
can write:

| What            | Fields                    | You decide        |
|-----------------|---------------------------|-------------------|
| Concept         | slug / name / summary     | all three         |
| facet           | kind / content            | kind; content is the text |
| edge            | from --relation--> to     | relation          |

- slug: ASCII snake-case, unique in the graph, the only handle you
  have for get_concept / manage_edge later (a-z / 0-9 / underscore
  only).
- name: the concept's name, in Chinese. The shared graph holds common
  names only: a coordinate, "day N" or "my" in a name means an
  individual case leaked in -- rewrite it as a common name.
- summary: one sentence saying what the concept is.
- kind: the short label (4-8 characters) naming what class of content
  this is, answering questions like "how is it done" or "when does it
  fail"; it is a label, not the content (the content goes in
  `content`). One facet per kind per concept and at most six facets
  under one concept, so a kind has to be able to summarise.
- content: the text of the facet, at most 400 characters (the tool
  refuses a longer one). Each facet is embedded as a vector of its
  own, and what exceeds the embedder's window is not retrieved, with
  no error: long does not mean better remembered, it means the second
  half is invisible. On hitting the cap, compress (keep numbers and
  conditions, cut throat-clearing) or split (into another kind, or
  into a new concept).
- relation: the predicate of the outgoing edge -- substitute both
  endpoints and read it aloud: "from <relation> to" must read as a
  sentence (requires is fine; precondition is not -- a noun does not
  read as a sentence and the direction is anyone's guess; nominal
  relations submitted by agents are rewritten into predicates when
  merging).

Finally: when all tool calls are done, call no more tools; output a
one-sentence summary of which experiences you merged, and stop.
\end{lstlisting}

\subsection{Retrieval: implementation and hyperparameters}
\label{sec:app_retrieval}

This appendix completes \S\ref{sec:retrieval} with the implementation and
the hyperparameter values. All embedding goes through one asymmetric
instruction-tuned encoder (KaLM, 896-d) \citep{zhao2025kalm}, over two
indices keyed by slug, one for concepts and one for skills. A concept is
indexed as several shards---one identity shard over its name and summary
and one shard per facet---and a query scores a concept by fusing its shards
back into one cosine-scale score, so that a facet can bring back the
concept it belongs to and a short name cannot outvote a well-supported one.
Anchoring admits a concept by the relative gate of
\S\ref{sec:retrieval} alone, with no absolute floor: cosine values are not
comparable across encoders or corpora, whereas the relative gate is
portable. On expansion, query--relation similarities are $z$-standardized
within each entry's candidates before entering
Eq.~\eqref{eq:edge-score}, which restores their spread without changing
the ranking: every relation text shares the entry name as a prefix, so the
raw similarities sit in a band in which softmax would be nearly uniform.
The candidate set is the relation types actually present on the current
frontier. Recall returns every facet of every node reached, which is
bounded by the (concept, kind) uniqueness of \appref{sec:app_schema}, and
every injected block is deduplicated against the conversation so far, so
that a repeated injection returns only the increment. Retrieval defaults
to the experience layer, since the memory layer is already injected each
morning.

Table~\ref{tab:retrieval_hyper} lists the values used in every run; they
are identical across conditions.

\begin{table}[htbp]
\centering
\caption{Retrieval hyperparameters, identical across all conditions. The
temperature and $h$ govern only edge sampling; anchoring is
deterministic given the index.}
\label{tab:retrieval_hyper}
\small
\setlength{\tabcolsep}{6pt}
\begin{tabular}{llp{7.2cm}}
\toprule
Symbol & Value & Role \\
\midrule
$k$ & 4 & vector candidates considered for anchoring \\
$\rho$ & 0.80 & relative similarity gate: admit within $\rho$ of the best match \\
$h$ & 2 & hop cap; the agent chooses $1..h$ per query \\
$p$ & 0.95 & nucleus pruning of low-probability edges, after softmax \\
$\tau$ & 1.0 & softmax temperature over edge scores \\
$\Delta_{\mathrm{expand}}$ & 1.0 & strength gain when the agent expands a concept \\
$\Delta_{\mathrm{retrieve}}$ & 0.1 & strength gain when a query returns a concept \\
\bottomrule
\end{tabular}
\end{table}

\subsection{Skill lifecycle and the verbatim channel}
\label{sec:app_skills}

This appendix completes \S\ref{sec:sharing} on the skill side: the record
a skill carries, how it is created and maintained, and how it crosses
agent boundaries without ever being rewritten. The design question is the
mirror image of the textual channel's: there, knowledge is prose and
needs an LLM to merge it; here, knowledge is executable code, and what the
world reports back is unambiguous, namely whether a call resolved or
threw, so the machinery is measurement and bookkeeping, and the only LLM
judgement needed is ``which skill is failing and how do I fix mine''.
What that measurement records is \emph{execution completion}, not semantic
correctness: a call that resolves has run, which is a weaker claim than
having achieved what the caller intended, and no stage of the pipeline
checks the latter.

\paragraph{The record.}
A skill is one row in its author's library: an \texttt{intent} (the
natural-language condition under which it should be used, which is the
text the skill index embeds, so retrieval matches on \emph{when to use
it}, not on the code), a \texttt{signature} (what the context shows: the
call form, which must begin with the function name defined in the code,
since the acting agent calls by signature and a mismatch would make the
skill unreachable, which the creation tool rejects), the \texttt{js\_code} itself, optional \texttt{preconditions}, and a
\texttt{verified} flag. Beside these, the two count pairs and the
revision count of \appref{sec:app_schema}: the current-version pair is
the evaluation window, the all-time pair the skill's lifetime
reliability. One number answers ``did my fix work'', the other ``do I
trust this slot at all''.

\paragraph{Creation.}
Skills are written from programs that
executed successfully that day and whose outcome the reflector reads as
having achieved its intent; the creation prompt states this threshold and
the reflector applies it, and no independent test re-checks it. This
includes skills composed from several successful fragments of a day, in
which case the composed function is not separately executed before it is
registered. Skills are written by the
phase-1 reflector in the first place, and by the phase-2 maintenance
session when it adds one. A name collision is rejected outright: the JS function name and the slug
are unique across the whole group. A near-duplicate \emph{intent}, by
contrast, is only reported back to the author as an advisory list, since
whether two intents name the same job is a judgement the author makes and
not a fact the tool can settle.

\paragraph{Measurement: the verbatim channel.}
The code of every verified skill is injected wholesale into the execution
environment, in the same scope as the atomic API and available to every
\texttt{write\_code} step at zero context cost, while what enters the
context is only signatures: a base block of the top-20 by cumulative
success, supplemented by signatures retrieved for the current goal.
Listing all signatures instead fills the context with skills unrelated
to the current goal. Use is measured by the world, not inferred from text: each injected skill is
wrapped so that a call either resolves (recorded as an execution
completion) or throws (recorded as a failure, with the error message
kept). The counts are
routed back each night, deterministically, to the \emph{author's}
library, so the record a maintainer sees is the skill's true usage
across the whole group, and a user's failing calls adjust the author's
record, not the user's.

\paragraph{Maintenance and retirement.}
The phase-2 session (\appref{sec:app_reflector}) reads, for each of its
own skills: both count pairs, the revision count, and up to ten
deduplicated error messages from the current window. Its brief is to act
only on what is failing, with the fix driven by the error pattern; the
prompt below states it in full. Revising the code resets the evaluation
window while the lifetime totals keep accumulating. This is what makes
maintenance a feedback loop rather than guesswork; editing only the
intent or preconditions does not reset the window, as that is
documentation, not code. A skill that cannot be fixed is \emph{retired},
that is, demoted out of the verified set so that it is no longer injected
or retrieved. Retirement is reversible by rewriting. A skill nobody
ever calls is read as a retrieval failure of its intent, and is fixed by
making the intent more specific (``when smelting iron ore'' matches;
``mining'' does not). The author may revise or retire only its own
skills, and the barrier structure of \appref{sec:app_algorithm} ensures
that no agent starts a day calling a skill that was retired overnight.

\paragraph{Propagation.}
In the shared graph every verified skill is an isolated node that carries
its author's code verbatim. We implement these nodes as a read-only
registry over the per-agent libraries: retrieval fans out to every
library, merges by similarity with each hit tagged by author and counts,
and the code is neither duplicated nor rewritten in transit. The code that runs is the author's code, unmodified, and its revision
history lives in exactly one place.
The contrast with the textual channel (\S\ref{sec:sharing}) is what the
pair is for: prose has to be \emph{interpreted} to be merged, so it goes
through semantic consolidation, whereas code can be reused as it stands, so what
the group accumulates about it is measurement rather than an opinion.

\paragraph{The skill-maintenance prompt (English translation).}
The phase-2 prompt, abridged to its operative rules.

\begin{lstlisting}[style=promptstyle]
You are the maintainer of your own skill library. Each dawn, the skills
you wrote were called by many users during the past day; now you look at
how they performed, and whether to improve or retire them.

You will see all of your own skills, one line each: cumulative
successes/failures of the current code version since its last revision;
lifetime successes/failures over all versions, never reset; how many
times the code has been revised; and real error samples from this
window, deduplicated, at most ten per skill.

Judge each skill; act on the ones with problems. A skill with only
failures, or a high failure rate, has a bug, not bad luck: rewrite it
with update_skill, reading the errors---timeouts: split the work or
request a longer budget; missing inputs: check the inventory first;
pathfinding failures: aim at a looser target; wrong blocks: re-read the
world. One you cannot fix, demote for now: a retired skill stops being
injected, and stopping the loss beats letting it burn your step budget
again tomorrow. Skills that work are left alone. A skill nobody calls
means its intent is too broad or too narrow to be retrieved---that too
is fixed by rewriting.

Touch only your own skills. Revising code resets the counters: that is
your clean evaluation window, and tomorrow's counts tell you whether the
fix worked. Do not let a bad skill linger---it is injected every day and
wastes your budget until you fix it or retire it.

When you are done calling tools, output a one-sentence summary of what
you maintained, changed, or retired, and why.
\end{lstlisting}

\subsection{Relation to prior systems}
\label{sec:app_comparison}

\S\ref{sec:related} positions \SYSNAME{} against prior work argument by
argument; this appendix compresses the same comparison into a single view.
Table~\ref{tab:system_comparison} collects related agent-memory and
experience-sharing systems along the axes that the mechanism sections of
this paper isolate: who accumulates, what the artifact of accumulation is,
whether situated and transferable knowledge are separated, whether what is
shared gets consolidated, whether an accumulated store is transferred to a
new group, and how memory is given back at decision time. The rows are not
validated under one protocol: these systems were evaluated in different
environments, text games for AriGraph, coding benchmarks for GEA,
task-given embodied, game and question-answering benchmarks for G-Memory,
multi-agent task suites for LLMA-Mem, task-given decentralized control and
cooperative planning for SEPS, Policy retrieval and DAMCS, and Minecraft for
the remaining rows, so the table compares designs rather
than reported numbers.

\begin{table}[htbp]
\centering
\caption{Related agent-memory and experience-sharing systems. ``Unit''
is who accumulates; ``artifact'' is what accumulation produces;
``memory $\rightarrow$ context'' is how stored knowledge reaches the
acting agent. Rows are compared on design, not on reported scores.}
\label{tab:system_comparison}
\small
\setlength{\tabcolsep}{3.5pt}
\begin{tabular}{p{1.85cm}p{0.95cm}p{2.5cm}p{1.35cm}p{1.35cm}p{1.35cm}p{2.3cm}}
\toprule
System & Unit & Artifact & Situated / transferable split & Shared \& consolidated & Store moved to a new group & Memory $\rightarrow$ context \\
\midrule
Voyager & 1 & personal skill library (code) & --- (skills only) & --- & --- (one agent) & embedding top-$k$ over skill descriptions \\
GITM / JARVIS-1 & 1 & structured external knowledge + text memory & --- & --- & n.e. & retrieval over knowledge base \\
Optimus-1 & 1 & hybrid graph + experience pool & --- & --- & n.e. & graph query for planning \\
MrSteve & 1 & What-Where-When episodic trace & episodic only & --- & n.e. & spatiotemporal retrieval \\
AriGraph & 1 & KG world model, episodic + semantic & \cmark{} (two layers, one agent) & --- & n.e. & local subgraph exploration \\
ADAM & 1 & causal graph & --- & --- & n.e. & graph-guided reasoning \\
MARS & 1 & induced rule set & --- & --- & n.e. & rules in prompt \\
GEA & group & pooled evolutionary traces & --- & pooled, not consolidated & n.a. & reflector reads the pool \\
G-Memory & group & insight, query and interaction graphs & --- (levels of abstraction) & team store; insights summarized per task & n.e. & query-graph retrieval, role-filtered \\
LLMA-Mem & group & episodic, procedural and transactive stores & episodes local, procedures shared (hybrid topology) & procedures consolidated every $N$ tasks & n.e. & relevance--importance top-$k$ \\
SEPS & group (decentralized RL) & few high-priority transitions relayed between agents & --- & selected, not consolidated & n.e. & relayed transitions enter each learner's training data \\
Policy retrieval & group & policy modules (network parameters) & --- & retrieved and composed, not consolidated & n.e. & module chosen by task similarity \\
DAMCS & group & per-agent hierarchical KG; structured messages & --- & messages exchanged, no group store & n.e. & partner's message in prompt \\
\SYSNAME{} & group & two-layer graphs + skills; shared consolidated graph & \cmark{} (memory stays private) & \cmark{} (semantic aggregation) & \cmark{} (\S\ref{sec:transfer}) & intent-aware subgraph retrieval \\
\bottomrule
\end{tabular}

\vspace{2pt}
\raggedright\footnotesize \cmark{} = demonstrated for the object named in
the cell; --- = not applicable to the system as designed; n.e.\ = not
evaluated in the published description; n.a.\ = a transfer result is
reported for a different object, so the column does not apply. GEA
reports transfer of the evolved framework across coding models
\citep{weng2026gea}; our transfer experiment moves an accumulated
knowledge store, namely the shared graph and the skill library, into a
freshly initialized group, which is a different object and is not
comparable with that result.
\end{table}

Three rows deserve a sentence beyond the table, because the resemblance
is closest and the difference easiest to miss.

\paragraph{AriGraph.}
The nearest single-agent ancestor of our graph, evaluated on text games
\citep{anokhin2024arigraph}: an agent grows a
knowledge-graph world model with episodic and semantic layers, explores
it locally when deciding, and revises it from interaction. \SYSNAME{}
adopts the graph memory and the two-way split, and differs in what the
split is \emph{for}. In AriGraph both layers serve one agent's
understanding; in \SYSNAME{} the split is the sharing boundary. The
semantic layer is what may leave the agent, and the episodic layer never
does. A second mechanism (aggregation, \appref{sec:app_aggregator})
exists precisely to reconcile $N$ agents' semantic layers into one graph.
Retrieval also differs in kind: AriGraph's agent walks its own graph
interactively, whereas our retrieval is intent-keyed, scored, and
sampled, and runs over a graph the acting agent did not write.

\paragraph{Voyager.}
The source of our action space and of the skill artifact, and the row our
skill channel most resembles. The difference that matters most is where
goals come from. Voyager organizes exploration around tasks proposed by a
dedicated automatic curriculum module; \SYSNAME{} integrates goal
generation into the acting agent's ongoing interaction loop, in which the
agent selects and revises its own goals using its current state and its
accumulated knowledge, with no separate curriculum module. The distinction
concerns how goals are generated and how they govern interaction, not
whether human intervention is required: Voyager's curriculum is automatic,
and the system is evaluated on open-ended exploration, tech-tree mastery
and map coverage. Taking the group as given, three further differences
follow. Skills are
shared through a registry rather than living in one
library, so a skill's record accumulates every agent's calls, not just
its author's; maintenance is driven by that cross-agent record
(\appref{sec:app_skills}); and skills are the \emph{second} channel,
alongside the consolidated textual graph, which has no analogue in
Voyager. The ablations of \appref{sec:ablation_conditions} do not
separate the two channels, since each removal that affects sharing
(\abl{S}, \abl{E}) takes both away. The evaluated action interfaces also differ from Voyager's original design \citep{wang2023voyager}.
Its appendix lists expert-written control primitives under ``Control primitive APIs implemented by us'' separately from ``Control primitive APIs provided by Mineflayer''; the list includes
\texttt{exploreUntil}, \texttt{mineBlock}, \texttt{craftItem},
\texttt{smeltItem}, \texttt{killMob} and others. \texttt{mineBlock}, for
example, mines and collects a specified number of blocks within a 32-block
distance. These routines enter the prompt as expert-written programs. In
Voyager's own baseline comparisons (Table A.3), the listed differences are
the skill library, automatic curriculum and self-verification. We take the
essence of learning to be the induction of generalizable skills from many
trajectories of interaction; the upper bound of learning is such a set of
generalizable skills. A primitive packages resource search, navigation,
mining and collection into one long-horizon routine written in advance.
Adding it to the agent's initial capabilities raises its lower bound of
competence and shortens the interval between that lower bound and what
remains acquirable; starting from atomic actions preserves this interval,
with the upper bound unchanged. Our protocol therefore removes these
primitives to measure what can be acquired from atomic actions. Our
adaptation keeps the automatic curriculum, iterative code generation with
execution feedback, self-verification and per-agent skill library, and
starts every method from the same atomic API. Under this bare API,
Voyager's hard-coded first task can fail repeatedly;
\appref{sec:app_baselines} reports this as a concrete limitation of our
adaptation. These results characterize the adapted implementations under
our interface. The original end-to-end system lies outside this comparison.
The comparison remains a comparison of mechanisms: Voyager and Echo keep
their respective mechanisms while the supplied action capabilities are
held fixed.

\paragraph{GEA.}
The nearest group-level system, evaluated on coding benchmarks
\citep{weng2026gea}: the group is an evolutionary unit whose members'
traces are pooled and read back by a reflector. \SYSNAME{} also lets the
group read what every member produced, but the shared artifact is a graph
whose records are consolidated rather than a set of parallel traces, and
skills travel on a separate channel. The two also differ in what they
report as transfer, as the note under
Table~\ref{tab:system_comparison} states. We do not compare numbers with
GEA: its evaluation protocol and ours differ in environment, control
interface and budget (\appref{sec:app_baselines}).

\section{Experiments}
\label{sec:app_exp}

\subsection{Reproducibility configuration}
\label{sec:setup_details}

Table~\ref{tab:config} lists the settings that determine how a run behaves
and that a reader needs in order to rebuild one. Values are identical in
every reported condition unless a row says otherwise, and the
configuration file released with the code carries every one of them. Parameters
that are already tabulated elsewhere are not repeated here: retrieval
hyperparameters in Table~\ref{tab:retrieval_hyper}
(\appref{sec:app_retrieval}), memory injection budgets in
Table~\ref{tab:memory_budget} (\appref{sec:app_schema}), the action API
and observation fields in \appref{sec:app_interface}, and the personas in
Table~\ref{tab:personas} (\appref{sec:app_planner}).

\begin{table}[htbp]
\centering
\caption{Reproducibility configuration. Values are identical in every
reported condition unless a row says otherwise; the configuration file
released with the code carries every value, and each run records the
configuration it used in its own summary.}
\label{tab:config}
\small
\setlength{\tabcolsep}{4pt}
\begin{tabular}{p{2.4cm}p{10.4cm}}
\toprule
Group & Value \\
\midrule
Backbone & DeepSeek-V4.1-Flash (DS-V4.1F, default; also the backbone of the ablations and of both external baselines), GLM-5.3, hy4-preview and Kimi-K3; transfer target GLM-5.3-Flash; one backbone serves all three roles of a run \\
Temperature & 0.3 planner, 0.7 reflector, 0.7 aggregator \\
Top-$p$ & not set \\
Reasoning & \texttt{low} for all three roles \\
Output cap & 131{,}072 tokens per planner step, 65{,}536 per reflector and aggregator step, 8{,}192 for the night note \\
Step cap & 299 tool calls per day, 99 per reflector session, 99 per aggregation \\
Context & no truncation threshold: a day's conversation is carried in full and discarded at dawn \\
\midrule
Encoder & KaLM, asymmetric instruction-tuned, 896-d \citep{zhao2025kalm}, over two indices, one for concepts and one for skills \\
Query / document & an instruction prefix is prepended to queries only, never to indexed text \\
Truncation & 512-token encoder window; shards cut at 512 characters; facet content above 400 characters is rejected by the tool, not truncated \\
\midrule
Game & Minecraft Java Edition 1.21.11 (Fabric server), Mineflayer 4.37.1 client \\
World & seed 7 for every reported run, natural spawn point of that seed, difficulty hard, \texttt{keepInventory} on, one day--night cycle = 24{,}000 ticks, game time frozen during deliberation and reflection, game clock at $2.5\times$ \\
Agents & $N=3$ (Alice, Bob, Carol), each in a separate server instance generated from the same seed; $N=5$ and $N=8$ add the next personas in a fixed order \\
Horizon & 30 nights (60 for the time-horizon runs of \appref{sec:app_scaling}); aggregation every $K=2$ nights, i.e.\ 15 windows \\
\bottomrule
\end{tabular}
\end{table}

Two protocol choices are not configuration values and are stated here
because they affect how the numbers should be read. First, comparisons
are made within a backbone: \SYSNAME{} is paired with the planner-only
control on the same backbone, and absolute values are not compared across
backbones, which differ in survival by more than the effect we measure.
Second, the 30-day horizon is an evaluation
budget, not a stopping criterion of the framework: agents keep proposing
goals, accumulating experience and maintaining their knowledge with no
predefined terminal task, and nothing in the loop asks them to stop at
day 30. The transfer experiment of \S\ref{sec:transfer} uses
GLM-5.3-Flash, a smaller model in the same family as the GLM-5.3
backbone, as its target backbone; its source stores are the shared graph
and skill library of one completed DeepSeek-V4.1-Flash run at day 30, the
non-transfer control is initialized from empty stores, and both then
follow the identical 30-day protocol, three runs per group.

\paragraph*{Sources of randomness.}
The runs of a cell repeat one configuration; they are not different
random seeds. Fixed in every run: the world seed, the natural spawn point
of that seed (so the framework's spawn-point generator is never drawn),
difficulty, personas, prompts and all hyperparameters. Not fixed: (i) LLM
sampling, whose temperatures are in Table~\ref{tab:config} and which
exposes no controllable sampling seed; (ii) the edge sampling of relational
expansion (\appref{sec:app_retrieval}), which draws from an unseeded
generator, whereas anchoring is deterministic given the index; and (iii)
game dynamics, since mob spawns, combat outcomes and the timing of
asynchronous programs differ between runs on the same seed. A run's
number records launch order only and enters no random generator. The
spread across the runs of a cell therefore combines all three sources;
each cell reports ten runs.

\subsection{Metric definitions}
\label{sec:app_metrics}

All metrics come from the structured snapshots recorded at the end of each
day. Deaths is cumulative per agent, averaged over the agents of a cell.
XP/life is final-day cumulative XP divided by deaths $+1$, computed per
agent and then averaged; since XP is not lost on death, it measures what
an agent gains per life. Tool era is the highest technology tier
reached in the agent's equipment, iron or diamond in our runs, and read
from the tool tier itself rather than from obtaining obsidian, which is a
milestone with its own column. Each Tool era entry carries that tier, the
mean day on which the agents that reached it first did so, and their count
out of all agents of the cell. Obsidian is the day of first obtainment
and Nether the day of first entry, each
averaged over the agents that reached it and reported with that count in
parentheses. Counts are out of all agents of a cell (three per run, 30 per
cell). Repeat counts an agent's deaths that repeat one of its own earlier
deaths: same recorded cause, within 16 blocks of the earlier death's
position; it is a per-agent mean like Deaths. Weakest XP/life takes, in
each run, the lowest XP/life among its three agents and averages that over
the runs of a condition. Others' skills is the share of all skill calls in
a condition that invoke a skill whose author is another agent of the same
group, as recorded in the skill registry; it is zero by construction
wherever sharing is removed. Lower is better for Deaths, repeated
deaths and attainment days; higher is better for XP/life, the tool tier
and the counts in parentheses. All values are averages over the ten sampled runs of a cell; because
the agents of a run share one graph, the run is the sampling unit.

\subsection{Baselines}
\label{sec:app_baselines}

\paragraph*{The same-backbone baseline.}
The baseline is \SYSNAME{} with the knowledge machinery removed and nothing
else changed: same backbone, prompts, tool set, night count, group size,
world seed, and day--night protocol. The baseline agent keeps the planner
of \S\ref{sec:agents}. It still sets the day's goal and ordered steps
itself, still acts by writing one short program per step over the same
atomic API, and still reasons over the observations and feedback that come
back. What it loses is persistence. No local graph is written, no shared
graph is consolidated, no skill is registered or retrieved, and no memory
slice is injected in the morning; because in \SYSNAME{} the reflector is
the process that maintains an agent's graph and emits the one line that
crosses the day boundary, removing the graphs also removes the nightly
reflection, so no text at all is carried from one day to the next.
Retrieval has nothing to retrieve, and the planner meets each day with the
model, the observations, and nothing else. The residue is exactly the
standard single-agent LLM-agent stack
\citep{yao2023react,shinn2023reflexion,autogpt2023}, in the same family as
the planner-only control of \citet{wang2023voyager}; we do not claim the
two match item by item in configuration.
Relative to it, \SYSNAME{} adds persistent memory, nightly reflection,
the skill channel, intent-aware retrieval and group-level consolidation.
The gap between the two rows of a backbone is therefore the combined
effect of that knowledge stack under a matched agent--world interface: it
is not an estimate of what sharing alone contributes, and it does not
establish superiority over other memory-equipped agents.

\paragraph*{External baselines.}
Voyager \citep{wang2023voyager} and Echo \citep{li2026echo} are
re-implemented on the same server, world seed, spawn point, difficulty,
day--night protocol and 30-day horizon, with three isolated agents per
group that share nothing, and with DeepSeek-V4.1-Flash as the single
backbone of every module. Both act through the same bare Mineflayer API as
\SYSNAME{} instead of their original action layers, and both receive the
long-term objective of our protocol (survive, explore, build) as their
overall goal, so that what differs is the mechanism that turns that goal
into tasks and experience. Tasks continue across day boundaries, since
neither system has a day--night notion.
\begin{itemize}
\item \textbf{Voyager.} Automatic curriculum, iterative code generation
with execution feedback (up to four rounds per task), self-verification,
and a per-agent skill library, with the original temperatures and warm-up
schedule. The optional wiki question-answering context is omitted. Because
the hard-coded first task can fail repeatedly under the bare API, the
curriculum takes over after five consecutive failures of it.
\item \textbf{Echo.} The released code does not include the
runtime it depends on, so we port its mechanisms rather than run the
original system. Successful tasks are stored as five-dimensional
structured experience (attribute, structure, function, procedure,
interaction); the most recent success serves as the anchor, the top-3
other experiences are retrieved by the official equal-weight five-dimension
similarity, and in-context analogical learning induces the next task.
Code generation and self-verification use the same executor as Voyager.
The visual dimensions are written from text observations, since the agents
receive no images. Two guards are ours: a seed task is skipped after three
failures, and an induced task must advance beyond its anchor, because
without this rule agents fell into loops of same-tier material
substitutions whose value never rose.
\end{itemize}

\paragraph*{Scope of the comparison.}
The planner-only control measures the combined effect of the knowledge
stack under our protocol. Voyager and Echo are adapted to the same
low-level action interface, which fixes the action capabilities supplied
in advance. Voyager's expert-written control primitives are listed separately
from Mineflayer's API in its own appendix; our protocol studies the acquisition
of such composite capabilities. This design addresses acquisition from a
common interface and keeps every method on the same atomic API. Under this
bare API, Voyager's hard-coded first task can fail repeatedly;
\appref{sec:app_baselines} reports this as a concrete limitation of our
adaptation. The external comparisons characterize the adapted implementations
under our interface. Voyager keeps its automatic curriculum, iterative code
generation with execution feedback, self-verification and per-agent skill
library; Echo keeps its structured experience and analogy-induced tasks.
The comparison therefore remains a comparison of mechanisms. The no-sharing
condition \abl{S} is an internal control that retains each agent's private
graph, skills, and reflection while removing both sharing channels.
Table~\ref{tab:system_comparison} (\appref{sec:app_comparison}) places the
prior systems along the axes the mechanism sections isolate.

\subsection{Full main results}
\label{sec:app_full_results}

Tables~\ref{tab:main} and~\ref{tab:backbones} pool the runs of each cell;
Table~\ref{tab:app_per_run} lists those runs individually, numbered in
launch order. All runs use world seed 7 and the natural spawn point of
that seed; the rows of a cell differ only through the uncontrolled
randomness listed in \appref{sec:setup_details}, so the spread across them
is that cell's run-to-run spread.

{\scriptsize
\setlength{\tabcolsep}{1.4pt}
\begin{longtable}{llccccc}
\caption{Per-run values behind Tables~\ref{tab:main} and~\ref{tab:backbones}: one row
per run of three agents (ten per cell, in launch order; seed 7), with metrics as in
Table~\ref{tab:main} computed over the run's agents; Tool era gives the highest tier
reached in the run. A dash means no agent reached the milestone within 30
game days. Names following a milestone count are the personas of Table~\ref{tab:personas}.}\label{tab:app_per_run}\\
\toprule
Backbone & Run & Deaths\,$\downarrow$ & XP/life\,$\uparrow$ & Tool era & Obsidian & Nether \\
\midrule
\endfirsthead
\toprule
Backbone & Run & Deaths\,$\downarrow$ & XP/life\,$\uparrow$ & Tool era & Obsidian & Nether \\
\midrule
\endhead
\bottomrule
\endfoot
DS-V4.1F & Voyager-1 & $325.0$ & $0.7$ & $iron\,(2.3\,\mathrm{day},\,3/3)$ & --- & --- \\
 & Voyager-2 & $412.0$ & $0.5$ & $iron\,(2.5\,\mathrm{day},\,2/3)$ & --- & --- \\
 & Voyager-3 & $167.7$ & $1.7$ & $dmd.\,(3.0\,\mathrm{day},\,1/3)$ & --- & --- \\
 & Voyager-4 & $280.7$ & $0.6$ & $iron\,(2.3\,\mathrm{day},\,3/3)$ & --- & --- \\
 & Voyager-5 & $205.3$ & $0.9$ & $iron\,(2.3\,\mathrm{day},\,3/3)$ & --- & --- \\
 & Voyager-6 & $208.3$ & $1.1$ & $iron\,(3.7\,\mathrm{day},\,3/3)$ & --- & --- \\
 & Voyager-7 & $213.7$ & $0.8$ & $iron\,(2.0\,\mathrm{day},\,3/3)$ & --- & --- \\
 & Voyager-8 & $162.0$ & $0.8$ & $iron\,(2.0\,\mathrm{day},\,2/3)$ & --- & --- \\
 & Voyager-9 & $121.7$ & $1.7$ & $iron\,(1.7\,\mathrm{day},\,3/3)$ & --- & --- \\
 & Voyager-10 & $193.0$ & $1.3$ & $iron\,(2.3\,\mathrm{day},\,3/3)$ & --- & --- \\
 & Echo-1 & $87.0$ & $3.4$ & $dmd.\,(10.7\,\mathrm{day},\,3/3)$ & --- & --- \\
 & Echo-2 & $79.7$ & $2.4$ & $dmd.\,(25.5\,\mathrm{day},\,2/3)$ & --- & --- \\
 & Echo-3 & $105.7$ & $2.5$ & $dmd.\,(17.0\,\mathrm{day},\,3/3)$ & --- & --- \\
 & Echo-4 & $88.0$ & $4.8$ & $dmd.\,(12.7\,\mathrm{day},\,3/3)$ & --- & --- \\
 & Echo-5 & $35.0$ & $8.9$ & $dmd.\,(24.0\,\mathrm{day},\,3/3)$ & --- & --- \\
 & Echo-6 & $46.3$ & $4.7$ & $dmd.\,(4.0\,\mathrm{day},\,1/3)$ & --- & --- \\
 & Echo-7 & $78.3$ & $3.5$ & $dmd.\,(15.3\,\mathrm{day},\,3/3)$ & --- & --- \\
 & Echo-8 & $83.7$ & $3.7$ & $dmd.\,(18.7\,\mathrm{day},\,3/3)$ & --- & --- \\
 & Echo-9 & $74.3$ & $3.2$ & $dmd.\,(9.5\,\mathrm{day},\,2/3)$ & --- & --- \\
 & Echo-10 & $57.7$ & $5.1$ & $dmd.\,(14.3\,\mathrm{day},\,3/3)$ & --- & --- \\
 & Planner-only-1 & $20.7$ & $22.8$ & $dmd.\,(15.5\,\mathrm{day},\,2/3)$ & --- & --- \\
 & Planner-only-2 & $21.7$ & $16.6$ & $dmd.\,(10.0\,\mathrm{day},\,2/3)$ & --- & --- \\
 & Planner-only-3 & $28.7$ & $12.9$ & $dmd.\,(19.0\,\mathrm{day},\,2/3)$ & --- & --- \\
 & Planner-only-4 & $26.3$ & $15.3$ & $dmd.\,(13.3\,\mathrm{day},\,3/3)$ & --- & --- \\
 & Planner-only-5 & $41.3$ & $7.8$ & $dmd.\,(26.0\,\mathrm{day},\,1/3)$ & --- & --- \\
 & Planner-only-6 & $19.0$ & $16.1$ & $dmd.\,(14.0\,\mathrm{day},\,2/3)$ & --- & --- \\
 & Planner-only-7 & $28.0$ & $10.9$ & $dmd.\,(8.0\,\mathrm{day},\,1/3)$ & --- & --- \\
 & Planner-only-8 & $26.7$ & $14.8$ & $dmd.\,(15.0\,\mathrm{day},\,2/3)$ & --- & --- \\
 & Planner-only-9 & $39.7$ & $11.4$ & $dmd.\,(22.0\,\mathrm{day},\,1/3)$ & --- & --- \\
 & Planner-only-10 & $27.0$ & $11.4$ & $iron\,(2.7\,\mathrm{day},\,3/3)$ & --- & --- \\
 & OpenKAGE-1 & $26.0$ & $15.4$ & $dmd.\,(8.0\,\mathrm{day},\,3/3)$ & $17.0\,\mathrm{day}\,(1/3;\,\textnormal{Carol})$ & $23.0\,\mathrm{day}\,(1/3;\,\textnormal{Carol})$ \\
 & OpenKAGE-2 & $19.7$ & $10.9$ & $dmd.\,(11.0\,\mathrm{day},\,3/3)$ & --- & --- \\
 & OpenKAGE-3 & $17.7$ & $7.8$ & $dmd.\,(20.0\,\mathrm{day},\,2/3)$ & --- & --- \\
 & OpenKAGE-4 & $13.0$ & $24.0$ & $dmd.\,(16.7\,\mathrm{day},\,3/3)$ & $27.5\,\mathrm{day}\,(2/3;\,\textnormal{Alice, Carol})$ & --- \\
 & OpenKAGE-5 & $10.7$ & $13.0$ & $dmd.\,(8.0\,\mathrm{day},\,1/3)$ & --- & --- \\
 & OpenKAGE-6 & $9.7$ & $27.7$ & $dmd.\,(16.0\,\mathrm{day},\,3/3)$ & $20.0\,\mathrm{day}\,(1/3;\,\textnormal{Bob})$ & $23.0\,\mathrm{day}\,(1/3;\,\textnormal{Bob})$ \\
 & OpenKAGE-7 & $17.7$ & $10.6$ & $dmd.\,(12.7\,\mathrm{day},\,3/3)$ & $22.5\,\mathrm{day}\,(2/3;\,\textnormal{Alice, Carol})$ & $24.5\,\mathrm{day}\,(2/3;\,\textnormal{Alice, Carol})$ \\
 & OpenKAGE-8 & $12.0$ & $27.8$ & $dmd.\,(9.0\,\mathrm{day},\,1/3)$ & --- & --- \\
 & OpenKAGE-9 & $25.7$ & $6.1$ & $dmd.\,(17.5\,\mathrm{day},\,2/3)$ & --- & --- \\
 & OpenKAGE-10 & $10.0$ & $38.8$ & $dmd.\,(14.3\,\mathrm{day},\,3/3)$ & --- & --- \\
\midrule
GLM-5.3 & Planner-only-1 & $30.7$ & $16.0$ & $dmd.\,(7.7\,\mathrm{day},\,3/3)$ & --- & --- \\
 & Planner-only-2 & $42.0$ & $6.7$ & $dmd.\,(7.5\,\mathrm{day},\,2/3)$ & --- & --- \\
 & Planner-only-3 & $42.3$ & $5.2$ & $dmd.\,(15.5\,\mathrm{day},\,2/3)$ & --- & --- \\
 & Planner-only-4 & $41.7$ & $4.5$ & $dmd.\,(28.5\,\mathrm{day},\,2/3)$ & --- & --- \\
 & Planner-only-5 & $17.7$ & $15.1$ & $dmd.\,(13.0\,\mathrm{day},\,3/3)$ & --- & --- \\
 & Planner-only-6 & $34.3$ & $7.2$ & $dmd.\,(12.0\,\mathrm{day},\,1/3)$ & --- & --- \\
 & Planner-only-7 & $37.3$ & $6.5$ & $dmd.\,(19.0\,\mathrm{day},\,1/3)$ & --- & --- \\
 & Planner-only-8 & $29.7$ & $6.0$ & $dmd.\,(19.0\,\mathrm{day},\,1/3)$ & --- & --- \\
 & Planner-only-9 & $27.3$ & $8.1$ & $dmd.\,(19.7\,\mathrm{day},\,3/3)$ & --- & --- \\
 & Planner-only-10 & $31.0$ & $6.1$ & $dmd.\,(15.0\,\mathrm{day},\,3/3)$ & --- & --- \\
 & OpenKAGE-1 & $30.7$ & $22.5$ & $dmd.\,(12.7\,\mathrm{day},\,3/3)$ & $24.5\,\mathrm{day}\,(2/3;\,\textnormal{Alice, Carol})$ & --- \\
 & OpenKAGE-2 & $50.7$ & $4.8$ & $dmd.\,(12.0\,\mathrm{day},\,3/3)$ & $23.5\,\mathrm{day}\,(2/3;\,\textnormal{Bob, Carol})$ & --- \\
 & OpenKAGE-3 & $21.7$ & $23.7$ & $dmd.\,(12.0\,\mathrm{day},\,3/3)$ & --- & --- \\
 & OpenKAGE-4 & $53.0$ & $9.9$ & $dmd.\,(14.3\,\mathrm{day},\,3/3)$ & $24.0\,\mathrm{day}\,(2/3;\,\textnormal{Alice, Bob})$ & $28.0\,\mathrm{day}\,(1/3;\,\textnormal{Alice})$ \\
 & OpenKAGE-5 & $51.7$ & $24.3$ & $dmd.\,(14.3\,\mathrm{day},\,3/3)$ & --- & --- \\
 & OpenKAGE-6 & $25.0$ & $9.9$ & $dmd.\,(14.5\,\mathrm{day},\,2/3)$ & $25.0\,\mathrm{day}\,(2/3;\,\textnormal{Alice, Bob})$ & --- \\
 & OpenKAGE-7 & $9.7$ & $62.5$ & $dmd.\,(15.3\,\mathrm{day},\,3/3)$ & $29.0\,\mathrm{day}\,(1/3;\,\textnormal{Alice})$ & --- \\
 & OpenKAGE-8 & $17.3$ & $24.0$ & $dmd.\,(16.5\,\mathrm{day},\,2/3)$ & --- & --- \\
 & OpenKAGE-9 & $14.3$ & $33.8$ & $dmd.\,(14.3\,\mathrm{day},\,3/3)$ & --- & --- \\
 & OpenKAGE-10 & $12.0$ & $25.4$ & $dmd.\,(12.7\,\mathrm{day},\,3/3)$ & $16.0\,\mathrm{day}\,(1/3;\,\textnormal{Bob})$ & $26.0\,\mathrm{day}\,(1/3;\,\textnormal{Bob})$ \\
\midrule
hy4-preview & Planner-only-1 & $37.7$ & $9.7$ & $dmd.\,(15.0\,\mathrm{day},\,3/3)$ & --- & --- \\
 & Planner-only-2 & $45.0$ & $9.0$ & $dmd.\,(18.0\,\mathrm{day},\,3/3)$ & --- & --- \\
 & Planner-only-3 & $65.0$ & $5.0$ & $dmd.\,(19.3\,\mathrm{day},\,3/3)$ & --- & --- \\
 & Planner-only-4 & $66.7$ & $5.9$ & $dmd.\,(16.0\,\mathrm{day},\,3/3)$ & --- & --- \\
 & Planner-only-5 & $37.7$ & $7.9$ & $dmd.\,(14.3\,\mathrm{day},\,3/3)$ & --- & --- \\
 & Planner-only-6 & $61.0$ & $5.2$ & $dmd.\,(9.3\,\mathrm{day},\,3/3)$ & --- & --- \\
 & Planner-only-7 & $35.7$ & $11.8$ & $dmd.\,(9.3\,\mathrm{day},\,3/3)$ & --- & --- \\
 & Planner-only-8 & $46.3$ & $10.4$ & $dmd.\,(21.7\,\mathrm{day},\,3/3)$ & --- & --- \\
 & Planner-only-9 & $47.0$ & $6.6$ & $dmd.\,(23.0\,\mathrm{day},\,3/3)$ & --- & --- \\
 & Planner-only-10 & $28.0$ & $10.9$ & $dmd.\,(18.0\,\mathrm{day},\,2/3)$ & --- & --- \\
 & OpenKAGE-1 & $18.7$ & $17.1$ & $dmd.\,(25.0\,\mathrm{day},\,3/3)$ & --- & --- \\
 & OpenKAGE-2 & $5.3$ & $78.0$ & $iron\,(3.0\,\mathrm{day},\,3/3)$ & --- & --- \\
 & OpenKAGE-3 & $15.7$ & $65.8$ & $dmd.\,(13.0\,\mathrm{day},\,2/3)$ & --- & --- \\
 & OpenKAGE-4 & $15.3$ & $12.2$ & $dmd.\,(12.0\,\mathrm{day},\,2/3)$ & $28.0\,\mathrm{day}\,(1/3;\,\textnormal{Carol})$ & $29.0\,\mathrm{day}\,(1/3;\,\textnormal{Carol})$ \\
 & OpenKAGE-5 & $16.0$ & $8.9$ & $iron\,(4.7\,\mathrm{day},\,3/3)$ & --- & --- \\
 & OpenKAGE-6 & $20.3$ & $15.1$ & $dmd.\,(17.0\,\mathrm{day},\,1/3)$ & --- & --- \\
 & OpenKAGE-7 & $9.7$ & $42.2$ & $dmd.\,(19.5\,\mathrm{day},\,2/3)$ & --- & --- \\
 & OpenKAGE-8 & $18.0$ & $12.9$ & $dmd.\,(25.0\,\mathrm{day},\,2/3)$ & --- & --- \\
 & OpenKAGE-9 & $23.0$ & $25.9$ & $dmd.\,(13.0\,\mathrm{day},\,1/3)$ & --- & --- \\
 & OpenKAGE-10 & $14.3$ & $45.4$ & $iron\,(3.0\,\mathrm{day},\,3/3)$ & --- & --- \\
\midrule
Kimi-K3 & Planner-only-1 & $49.3$ & $6.1$ & $dmd.\,(24.5\,\mathrm{day},\,2/3)$ & --- & --- \\
 & Planner-only-2 & $73.7$ & $4.0$ & $dmd.\,(22.5\,\mathrm{day},\,2/3)$ & --- & --- \\
 & Planner-only-3 & $71.7$ & $10.2$ & $dmd.\,(26.0\,\mathrm{day},\,1/3)$ & --- & --- \\
 & Planner-only-4 & $93.7$ & $7.5$ & $dmd.\,(15.5\,\mathrm{day},\,2/3)$ & --- & --- \\
 & Planner-only-5 & $53.3$ & $8.5$ & $dmd.\,(30.0\,\mathrm{day},\,1/3)$ & --- & --- \\
 & Planner-only-6 & $54.7$ & $7.0$ & $dmd.\,(11.5\,\mathrm{day},\,2/3)$ & --- & --- \\
 & Planner-only-7 & $101.0$ & $4.5$ & $dmd.\,(24.0\,\mathrm{day},\,1/3)$ & --- & --- \\
 & Planner-only-8 & $41.0$ & $7.7$ & $dmd.\,(19.0\,\mathrm{day},\,2/3)$ & --- & --- \\
 & Planner-only-9 & $39.7$ & $7.8$ & $dmd.\,(30.0\,\mathrm{day},\,2/3)$ & --- & --- \\
 & Planner-only-10 & $71.3$ & $6.7$ & $iron\,(2.7\,\mathrm{day},\,3/3)$ & --- & --- \\
 & OpenKAGE-1 & $41.7$ & $5.3$ & $dmd.\,(20.0\,\mathrm{day},\,1/3)$ & --- & --- \\
 & OpenKAGE-2 & $39.0$ & $4.6$ & $dmd.\,(12.0\,\mathrm{day},\,1/3)$ & --- & --- \\
 & OpenKAGE-3 & $41.3$ & $6.2$ & $dmd.\,(15.0\,\mathrm{day},\,1/3)$ & $20.0\,\mathrm{day}\,(1/3;\,\textnormal{Alice})$ & --- \\
 & OpenKAGE-4 & $32.3$ & $4.7$ & $iron\,(2.0\,\mathrm{day},\,3/3)$ & --- & --- \\
 & OpenKAGE-5 & $59.0$ & $5.4$ & $iron\,(3.7\,\mathrm{day},\,3/3)$ & --- & --- \\
 & OpenKAGE-6 & $24.7$ & $16.1$ & $dmd.\,(20.5\,\mathrm{day},\,2/3)$ & --- & --- \\
 & OpenKAGE-7 & $56.7$ & $7.3$ & $dmd.\,(18.5\,\mathrm{day},\,2/3)$ & $24.0\,\mathrm{day}\,(1/3;\,\textnormal{Alice})$ & --- \\
 & OpenKAGE-8 & $27.7$ & $7.5$ & $dmd.\,(22.0\,\mathrm{day},\,1/3)$ & --- & --- \\
 & OpenKAGE-9 & $34.3$ & $4.5$ & $dmd.\,(27.0\,\mathrm{day},\,1/3)$ & --- & --- \\
 & OpenKAGE-10 & $64.0$ & $3.7$ & $dmd.\,(17.0\,\mathrm{day},\,1/3)$ & --- & --- \\

\end{longtable}
}

\begin{figure}[htbp]
\centering
\includegraphics[width=\linewidth]{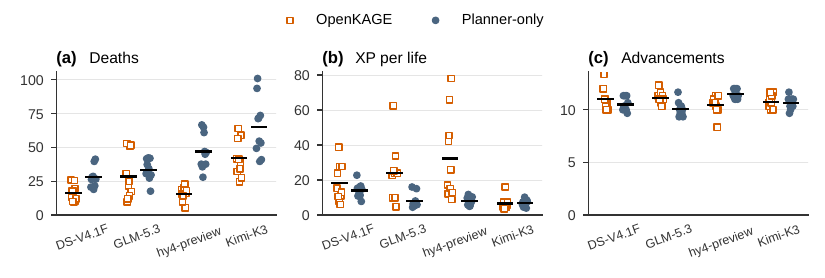}
\caption{Run-level values behind Table~\ref{tab:backbones}: one dot per run
(per-agent mean over its three agents), with the cell mean as a horizontal
bar. (a)~Deaths. (b)~XP per life. (c)~Milestones unlocked.}
\label{fig:runs}
\end{figure}

\paragraph{Run-level distributions.}
Figure~\ref{fig:runs} shows every run behind Table~\ref{tab:backbones}
rather than the cell means. The separation in deaths is not driven by a few
runs: on hy4-preview and DS-V4.1F all ten \SYSNAME{} runs die less than the median
planner-only run, and on Kimi-K3 nine of ten do. GLM-5.3 is the
exception: seven of its \SYSNAME{} runs lie below the planner-only median,
but three runs with more than $50$ deaths per agent pull its mean up, which
is why its mean difference in deaths is small. XP
per life shows the reverse picture on GLM-5.3, where \SYSNAME{} runs spread
upwards and the control's stay low.

\begin{figure}[htbp]
\centering
\includegraphics[width=\linewidth]{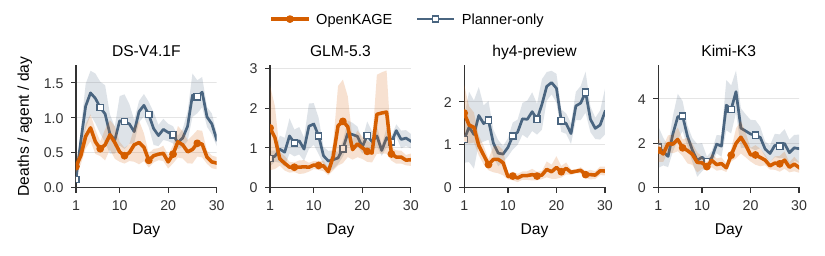}
\caption{Deaths per agent per day (three-day moving average) for
\SYSNAME{} and its planner-only control on each backbone: mean over 10 runs
of three agents, with a band of $\pm 1$ standard error across runs. Each
panel has its own vertical scale.}
\label{fig:death_rate}
\end{figure}

\paragraph{Deaths over time.}
Figure~\ref{fig:death_rate} plots deaths per agent per day for \SYSNAME{}
and its planner-only control on each backbone. \SYSNAME{} dies less on
most days, $21$ to $29$ of the $30$ depending on the backbone (on the
smoothed curves of Figure~\ref{fig:death_rate}). On
hy4-preview and Kimi-K3 its daily rate also falls over the 30 days, while
the control's stays level or rises; on hy4-preview the change from the
first to the last ten days differs between the two ($-5.1$ vs.\ $+4.8$
deaths per agent). On DS-V4.1F both
curves are roughly flat, and on GLM-5.3 the rate of \SYSNAME{} rises over
the horizon while staying below the control's, so on these two backbones
the reduction is a lower level rather than a steeper decline.

\begin{figure}[htbp]
\centering
\includegraphics[width=\linewidth]{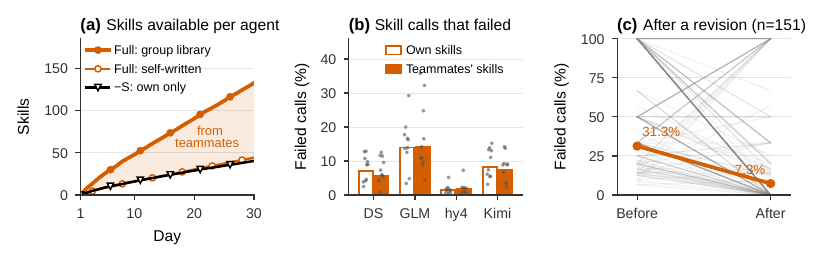}
\caption{Growth and reuse of the shared skill library (10 runs of three agents per
curve). (a)~Skills available to an agent on DeepSeek-V4.1-Flash: under Full the whole
group library, of which the agent wrote only the self-written part and the shaded gap
comes from teammates; under \abl{S} only the skills it wrote. (b)~Share of skill calls
that raised an error, for own and teammates' skills on each backbone (bars pool ten runs,
dots are single runs). (c)~DS-V4.1F: each thin line is one of the $151$ revisions after
which the skill was called again, joining its failure share before and after; the thick
line pools all.}
\label{fig:knowledge}
\end{figure}

\paragraph{Growth and reuse of the skill library.}
On the default backbone, the skill library available to each agent grows
almost linearly over the 30 days, from $47.8$ skills on day 10 to $90.3$ on
day 20 and $132.5$ on day 30 (Figure~\ref{fig:knowledge}a). Each agent
writes about $44$ of these skills itself, close to an agent that
shares nothing ($40.3$ under \abl{S}), so the shaded gap in
Figure~\ref{fig:knowledge}a, roughly two thirds of the library, comes from
teammates. On every backbone, $62$ to $65\%$ of all skill calls invoke a
skill written by another agent of the group. How large the library grows
depends on the backbone, from about $35$ to $40$ skills per agent on GLM-5.3 and
Kimi-K3 to about $130$ to $140$ on DS-V4.1F and hy4-preview, in line with the
backbones' differences in milestone attainment. On DS-V4.1F, the ten groups
called $462$ distinct skills, and $358$ of these were called at least once
by an agent other than their author.

A call counts as failed when the skill raised an error. The failure rate
differs between backbones, from about $1.5\%$ of calls on hy4-preview to
$14\%$ on GLM-5.3, but on none of them does it differ by more than about one
and a half percentage points between an agent's own skills and its teammates'
(Figure~\ref{fig:knowledge}b); on DS-V4.1F it is $5.6\%$ of $3{,}526$
calls for teammates' skills against $6.9\%$ of $2{,}206$ for the agent's
own. A skill written by one agent therefore fails about as often in the hands
of another as in its author's. These rates measure execution errors only; a
call that returned without error need not have reached its goal.

\paragraph{Revision of skills after failed calls.}
Skills are not written once and left. Each night the author of a skill sees
its accumulated successes and failures, with samples of the error messages,
and may rewrite its code or take it out of circulation. On DS-V4.1F, authors
revised the code of $160$ skills in $232$ revisions (one revision for $117$
skills, two for $26$, three to nine for the other $17$) and took skills
out of circulation $50$ times. In $225$ of the $232$ revisions the skill had at least
one failed call since it was written or last revised. For the $151$ of these
revisions after which the skill was called again, the failure rate of its
calls fell from $31.3\%$ of $706$ calls before the revision to $7.3\%$
of $1{,}624$ calls after it (Figure~\ref{fig:knowledge}c). Across the $151$
revisions the failure rate fell for $109$, was unchanged for $13$ and rose
for $29$, and was zero after the revision for $81$. This is not a causal
estimate: skills enter the comparison because they failed, so part of the
fall is regression to the mean, and the windows are counted by day, not by
call. The rate after revision remains above the $6.1\%$ failure rate of all
calls. Revisions do not always succeed, and a revision can trade one error
for another; of the $29$ rises, $23$ rest on three or fewer calls after the
revision. Table~\ref{tab:app_revisions} lists four revisions with the
error that triggered each.

\begin{table}[htbp]
\centering
\caption{Examples of skill revisions triggered by failed calls (DS-V4.1F). Calls are
counted as successes/failures before the revision (since the skill was written or last
revised) and after it. One example per distinct error message is taken in run order,
among revisions after which the skill did not fail again.}
\label{tab:app_revisions}
\small
\setlength{\tabcolsep}{3.5pt}
\begin{tabular}{@{}p{2.9cm}p{3.4cm}p{4.9cm}cc@{}}
\toprule
Skill & Error before & Change & Before & After \\
\midrule
\texttt{build\_\allowbreak box\_\allowbreak hut\_\allowbreak with\_\allowbreak door} & \texttt{doorCell is not iterable} & Reads the door cell as an array and guards a missing door & $1/1$ & $1/0$ \\
\texttt{seal\_\allowbreak side\_\allowbreak shelter\_\allowbreak and\_\allowbreak torch} & \texttt{origin is not iterable} & Normalises the origin argument to an array before use & $0/2$ & $3/0$ \\
\texttt{place\_\allowbreak block\_\allowbreak clear\_\allowbreak of\_\allowbreak self} & no support block below target & Checks the block below the target and returns early when absent & $3/1$ & $14/0$ \\
\texttt{dig\_\allowbreak and\_\allowbreak seal\_\allowbreak pit} & timeout after 5100\,ms & Replaces tick waits with a timed sleep & $2/1$ & $1/0$ \\
\bottomrule
\end{tabular}
\end{table}

\subsection{Ablation conditions}
\label{sec:ablation_conditions}

What each condition of \S\ref{sec:ablation} leaves intact matters for
reading the gaps, so it is stated per condition.

\begin{itemize}
\item \textbf{Raw. Direct sharing (no consolidation).} No component is
removed: private graphs, subgraph retrieval, the memory layer, the shared
graph and cross-agent skill reuse all stay as in Full. Only the aggregation
step changes: at each window the three agents' experience-layer concepts,
facets and edges are appended to the shared graph verbatim, with
window--arm prefixes on slugs, no merging or selection, and no edges
crossing into the memory layer; the aggregation agent is not called.
Retrieval budgets, skill sharing and layer clearing are unchanged; its
comparison with Full is reported below.
\item \textbf{\abl{S}. No sharing (individual memory).} Private graphs and
subgraph retrieval stay; both sharing channels go: no aggregation (the
shared graph stays empty, retrieval sees only the agent's own graph) and
no cross-agent skill reuse (skills are read from the author's library
only).
\item \textbf{\abl{R}. Flat retrieval (RAG-style).} The full knowledge
pipeline stays, including writing, aggregation, sharing, skill retrieval and
the vector anchoring step with the same relative gate. Retrieval then stops at
the anchors: it returns their facets as a flat list, with no relational
expansion, no edge structure in the result, and the agent's expand tool is
removed. The condition applies to both the private and the shared graph.
The three removals are bundled, so the comparison with Full does not
separate multi-hop expansion from seeing the edges, and \abl{R} also reads
less retrieved text when the plan is set (\appref{sec:app_cost}). On the
default backbone the agents of Full rarely call retrieve themselves (67
calls in 10 runs, 57 of them at one hop) and never call expand, so the
difference arises mostly in the retrieval that setting a plan triggers.
Only retrieval varies, though the knowledge agents subsequently act on may
diverge as they take different actions.
\item \textbf{\abl{M}. No memory layer.} The experience layer,
aggregation, the shared graph, and skills all keep operating; reflection
no longer writes memory concepts and the morning injection disappears.
Unlike the others, this one varies \emph{which layer} stores knowledge,
not what is stored.
\item \textbf{\abl{E}. No experience layer.} Reflection writes only to
the memory layer: no textual experience record is written and the skill
channel is off, so no skill is created, maintained, injected or reused
across agents. The aggregator receives nothing, the shared graph stays empty, and
experience retrieval returns nothing.
\item \textbf{B. No graph (bare model).} Removes every graph, private
and shared: no memory, no aggregation, and no skills. This is the no-graph
control of the main experiment, here also the floor of the suite.
\end{itemize}

Two caveats follow from the design itself. First, \abl{S} and \abl{E}
both leave the shared graph empty; they differ in whether the agent's
private experience is still written. It is under \abl{S} but not under
\abl{E}, so their gaps overlap by construction and must not be read as
additive independent contributions. Second, the run-to-run spread within
a condition is wide (Table~\ref{tab:app_ablation_per_run}), so small
differences in Deaths and XP/life between conditions are read as lying within
that spread.

Table~\ref{tab:app_ablation_per_run} lists the ten runs of each ablation
condition, including Raw, in the conventions of Table~\ref{tab:main}.
Full and B are the DS-V4.1F \SYSNAME{} and planner-only rows of
Table~\ref{tab:app_per_run}.

{\scriptsize
\setlength{\tabcolsep}{1.9pt}
\begin{longtable}{lccccc}
\caption{Per-run values of the ablated conditions (DeepSeek-V4.1-Flash,
30 game days, world seed 7), one row per run of three agents (including
the Raw condition), in launch order and in the conventions of
Table~\ref{tab:app_per_run}; milestone counts name the agents as in Table~\ref{tab:app_per_run} (personas in Table~\ref{tab:personas}).}\label{tab:app_ablation_per_run}\\
\toprule
Run & Deaths\,$\downarrow$ & XP/life\,$\uparrow$ & Tool era & Obsidian & Nether \\
\midrule
\endfirsthead
\toprule
Run & Deaths\,$\downarrow$ & XP/life\,$\uparrow$ & Tool era & Obsidian & Nether \\
\midrule
\endhead
\bottomrule
\endfoot
Raw-1 & $17.3$ & $35.2$ & $dmd.\,(16.5\,\mathrm{day},\,2/3)$ & --- & --- \\
Raw-2 & $8.0$ & $29.1$ & $dmd.\,(22.0\,\mathrm{day},\,1/3)$ & --- & --- \\
Raw-3 & $10.7$ & $35.6$ & $dmd.\,(9.0\,\mathrm{day},\,2/3)$ & --- & --- \\
Raw-4 & $32.0$ & $16.5$ & $dmd.\,(11.0\,\mathrm{day},\,3/3)$ & --- & --- \\
Raw-5 & $15.7$ & $9.6$ & $dmd.\,(15.5\,\mathrm{day},\,2/3)$ & --- & --- \\
Raw-6 & $16.3$ & $19.5$ & $dmd.\,(9.0\,\mathrm{day},\,2/3)$ & --- & --- \\
Raw-7 & $7.3$ & $29.3$ & $iron\,(2.3\,\mathrm{day},\,3/3)$ & --- & --- \\
Raw-8 & $13.3$ & $12.8$ & $iron\,(2.3\,\mathrm{day},\,3/3)$ & --- & --- \\
Raw-9 & $39.0$ & $19.6$ & $dmd.\,(13.5\,\mathrm{day},\,2/3)$ & --- & --- \\
Raw-10 & $16.7$ & $8.3$ & $iron\,(2.7\,\mathrm{day},\,3/3)$ & --- & --- \\
\midrule
\abl{S}-1 & $27.0$ & $15.1$ & $dmd.\,(9.0\,\mathrm{day},\,2/3)$ & --- & --- \\
\abl{S}-2 & $20.7$ & $16.1$ & $dmd.\,(18.3\,\mathrm{day},\,3/3)$ & $14.0\,\mathrm{day}\,(1/3;\,\textnormal{Carol})$ & $21.0\,\mathrm{day}\,(1/3;\,\textnormal{Carol})$ \\
\abl{S}-3 & $13.7$ & $14.5$ & $dmd.\,(8.0\,\mathrm{day},\,2/3)$ & --- & --- \\
\abl{S}-4 & $18.7$ & $10.8$ & $dmd.\,(9.0\,\mathrm{day},\,3/3)$ & --- & --- \\
\abl{S}-5 & $15.7$ & $8.4$ & $dmd.\,(16.0\,\mathrm{day},\,2/3)$ & --- & --- \\
\abl{S}-6 & $24.7$ & $6.3$ & $dmd.\,(17.5\,\mathrm{day},\,2/3)$ & --- & --- \\
\abl{S}-7 & $14.7$ & $21.5$ & $dmd.\,(13.0\,\mathrm{day},\,3/3)$ & $29.0\,\mathrm{day}\,(1/3;\,\textnormal{Bob})$ & --- \\
\abl{S}-8 & $13.7$ & $15.9$ & $dmd.\,(14.3\,\mathrm{day},\,3/3)$ & --- & --- \\
\abl{S}-9 & $10.7$ & $46.7$ & $dmd.\,(14.0\,\mathrm{day},\,1/3)$ & --- & --- \\
\abl{S}-10 & $14.7$ & $6.3$ & $iron\,(5.0\,\mathrm{day},\,3/3)$ & --- & --- \\
\midrule
\abl{R}-1 & $19.7$ & $8.3$ & $dmd.\,(17.0\,\mathrm{day},\,1/3)$ & --- & --- \\
\abl{R}-2 & $17.0$ & $14.1$ & $dmd.\,(16.0\,\mathrm{day},\,2/3)$ & --- & --- \\
\abl{R}-3 & $22.0$ & $7.9$ & $dmd.\,(10.0\,\mathrm{day},\,2/3)$ & --- & --- \\
\abl{R}-4 & $12.7$ & $19.1$ & $dmd.\,(7.0\,\mathrm{day},\,2/3)$ & --- & --- \\
\abl{R}-5 & $15.7$ & $9.0$ & $dmd.\,(17.0\,\mathrm{day},\,1/3)$ & --- & --- \\
\abl{R}-6 & $17.0$ & $7.2$ & $iron\,(4.3\,\mathrm{day},\,3/3)$ & --- & --- \\
\abl{R}-7 & $13.7$ & $12.1$ & $iron\,(4.0\,\mathrm{day},\,3/3)$ & --- & --- \\
\abl{R}-8 & $8.3$ & $28.7$ & $dmd.\,(20.0\,\mathrm{day},\,2/3)$ & --- & --- \\
\abl{R}-9 & $19.7$ & $7.5$ & $iron\,(4.7\,\mathrm{day},\,3/3)$ & --- & --- \\
\abl{R}-10 & $10.7$ & $18.6$ & $dmd.\,(10.0\,\mathrm{day},\,2/3)$ & --- & --- \\
\midrule
\abl{M}-1 & $8.3$ & $33.4$ & $dmd.\,(12.5\,\mathrm{day},\,2/3)$ & --- & --- \\
\abl{M}-2 & $5.0$ & $103.1$ & $dmd.\,(21.7\,\mathrm{day},\,3/3)$ & --- & --- \\
\abl{M}-3 & $6.3$ & $58.8$ & $dmd.\,(22.0\,\mathrm{day},\,1/3)$ & --- & --- \\
\abl{M}-4 & $16.0$ & $11.5$ & $iron\,(2.7\,\mathrm{day},\,3/3)$ & --- & --- \\
\abl{M}-5 & $17.3$ & $16.5$ & $dmd.\,(21.0\,\mathrm{day},\,2/3)$ & --- & --- \\
\abl{M}-6 & $13.3$ & $19.1$ & $dmd.\,(20.5\,\mathrm{day},\,2/3)$ & --- & --- \\
\abl{M}-7 & $9.3$ & $18.8$ & $dmd.\,(13.0\,\mathrm{day},\,1/3)$ & --- & --- \\
\abl{M}-8 & $14.0$ & $28.5$ & $dmd.\,(18.5\,\mathrm{day},\,2/3)$ & --- & --- \\
\abl{M}-9 & $8.7$ & $57.6$ & $dmd.\,(14.7\,\mathrm{day},\,3/3)$ & --- & --- \\
\abl{M}-10 & $17.7$ & $30.1$ & $dmd.\,(17.3\,\mathrm{day},\,3/3)$ & --- & --- \\
\midrule
\abl{E}-1 & $6.3$ & $37.6$ & $dmd.\,(4.0\,\mathrm{day},\,1/3)$ & --- & --- \\
\abl{E}-2 & $8.7$ & $14.1$ & $dmd.\,(29.0\,\mathrm{day},\,1/3)$ & --- & --- \\
\abl{E}-3 & $16.7$ & $44.0$ & $dmd.\,(12.5\,\mathrm{day},\,2/3)$ & --- & --- \\
\abl{E}-4 & $18.7$ & $26.6$ & $dmd.\,(16.5\,\mathrm{day},\,2/3)$ & --- & --- \\
\abl{E}-5 & $16.7$ & $15.9$ & $dmd.\,(12.5\,\mathrm{day},\,2/3)$ & --- & --- \\
\abl{E}-6 & $26.3$ & $6.6$ & $dmd.\,(18.0\,\mathrm{day},\,1/3)$ & --- & --- \\
\abl{E}-7 & $15.7$ & $15.6$ & $dmd.\,(8.0\,\mathrm{day},\,1/3)$ & --- & --- \\
\abl{E}-8 & $13.7$ & $25.9$ & $dmd.\,(14.7\,\mathrm{day},\,3/3)$ & --- & --- \\
\abl{E}-9 & $10.7$ & $12.8$ & $dmd.\,(17.0\,\mathrm{day},\,1/3)$ & --- & --- \\
\abl{E}-10 & $11.3$ & $17.9$ & $dmd.\,(7.0\,\mathrm{day},\,1/3)$ & --- & --- \\

\end{longtable}
}

\paragraph{Raw: consolidation versus direct sharing.}\label{sec:app_raw_sharing}
Full and Raw share the horizon, group size, world seed and hyperparameters,
and the same records reach the shared graph in both; only their organisation
differs. Table~\ref{tab:ablation} summarizes both cells (ten runs each) and
Table~\ref{tab:app_ablation_per_run} lists the Raw runs.

The mean differences are small at this sample size: repeated deaths are
$7.4$ vs.\ $10.8$, the weakest agent's XP/life $7.9$ vs.\ $5.9$, mean deaths
$16.2$ vs.\ $17.6$ and mean XP/life $18.2$ vs.\ $21.6$; obsidian is $6/30$
vs.\ $0/30$ and the Nether $4/30$ vs.\ $0/30$. All three agents of a group
obtain a diamond pickaxe in $6$ of $10$ Full runs and $1$ of $10$ Raw runs.
With 10 runs per condition, the means of Raw and Full lie within each other's
run-to-run spread, and the deep-milestone counts are small for both. The sharing
channel is used almost equally in both conditions ($62\%$ vs.\ $60\%$
of skill calls invoke another agent's skill, $94\%$ and $95\%$ success), so
the comparison varies how the shared records are organised while the sharing
channel is active. The link from the structure of the shared graph to
behaviour is descriptive here.

\subsection{Scaling: long-horizon accumulation and group size}
\label{sec:app_scaling}

Sections~\ref{sec:results}--\ref{sec:transfer} show that \SYSNAME{}
reaches milestones the baseline does not reach, and that the knowledge it
accumulates transfers to a fresh group on another backbone. Two questions
remain. Does per-agent progress grow when the group grows? Does progress keep
accumulating once the horizon is doubled? The following runs examine both
axes, group size and time horizon; they are supplementary: each curve is one
run, so they support observations rather than conclusions.

\begin{figure}[htbp]
\centering
\includegraphics[width=\linewidth]{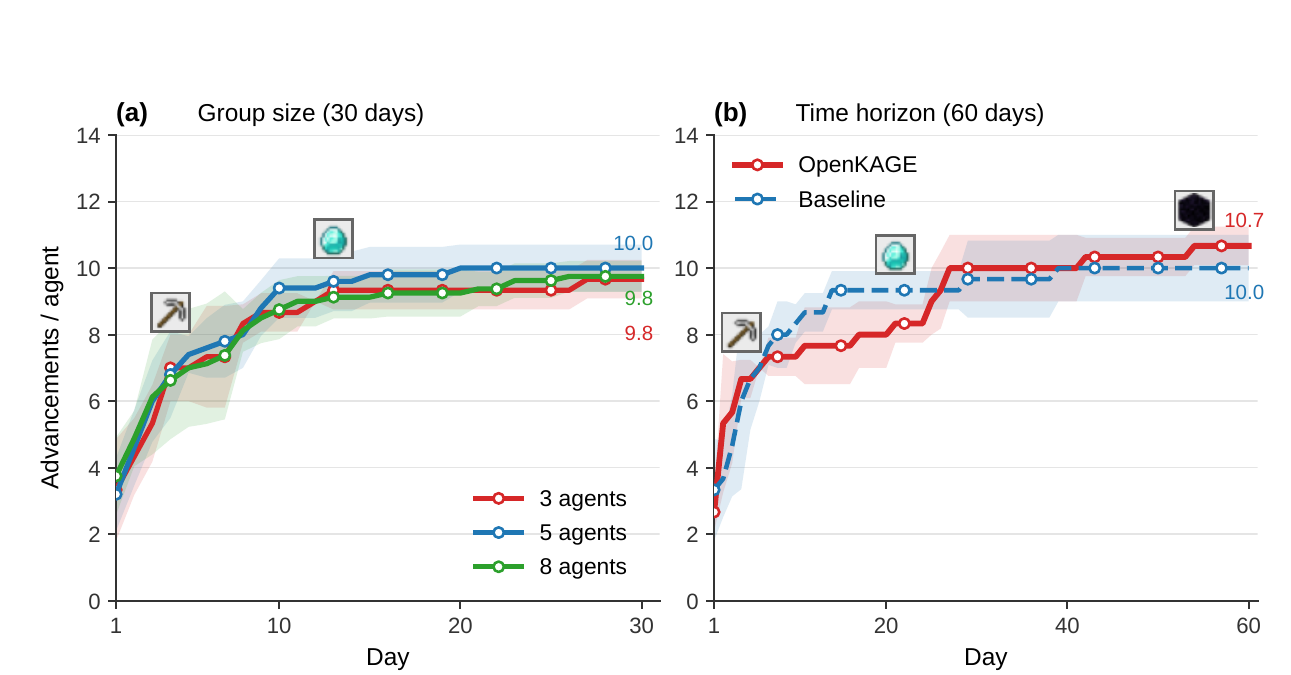}
\caption{Scaling in group size and time horizon (GLM-5.3-Flash). Mean milestones per
agent, bands $\pm 1$ sd across agents; each curve is a single run. (a)~$N{=}3$, $5$, $8$
agents, 30 game days. (b)~\SYSNAME{} and the baseline over 60 game days at $N{=}3$.}
\label{fig:scaling}
\end{figure}

\paragraph{Similar endpoints across evaluated group sizes.}
Groups of $N{=}3$, $5$ and $8$ agents running
30 game days each reach comparable endpoints; the gaps between them are
smaller than the spread across the agents of any one group
(Figure~\ref{fig:scaling}(a)). With one run per group size, this shows no
group-size effect within the observed spread, not that none exists.

\paragraph{Time scale: a longer horizon reaches the next milestone.}
At $N{=}3$, extending the protocol from 30 to 60 game days lets \SYSNAME{}
reach obsidian on day 54 ($1$ of $3$ agents). The baseline, given the same
60 days, stalls at a bottleneck from day 40 on, and no agent reaches
obsidian (Figure~\ref{fig:scaling}(b)).

\paragraph{GLM-5.3-Flash is a weak backbone: deep milestones stay out of reach.}
\label{sec:app_flash}
The target backbone of the transfer experiment (\S\ref{sec:transfer}) is a
small model, and its own ceiling bounds what that experiment can show. The
scaling runs above, all on GLM-5.3-Flash and starting from empty stores, give
that ceiling. Enlarging the group to $N{=}5$ or $N{=}8$ agents does not carry
any of them to obsidian or the Nether within 30 game days
(Figure~\ref{fig:scaling}(a)); even doubling the horizon to 60 game days
yields obsidian for only $1$ of $3$ agents, on day 54, and no agent enters the
Nether (Figure~\ref{fig:scaling}(b)). In the transfer experiment, by contrast,
one of nine agents obtains obsidian within 30 days. Thus on this weak backbone
the accumulated knowledge reaches, within 30 days, a milestone that groups
starting from empty stores do not reach with more agents or with the same
30-day horizon.

\subsection{Inference cost}
\label{sec:app_cost}

Game time is paused during deliberation, so a fixed number of game days
does not fix the inference budget. Table~\ref{tab:app_cost} counts what each
condition spends per game night, and Table~\ref{tab:app_text} splits
\SYSNAME{}'s share by component and reports the wall-clock hours of a
complete run for each condition.

\begin{table}[htbp]
\centering
\caption{Inference cost per game night (30 game days, ten runs of
each cell). Calls are LLM calls of the whole group of three agents; steps
are executed environment steps per agent.}
\label{tab:app_cost}
\small
\setlength{\tabcolsep}{5pt}
\begin{tabular}{llcc}
\toprule
Backbone & Method & LLM calls & Steps / agent \\
\midrule
DS-V4.1F & Voyager & $439$ & $66.8$ \\
 & Echo & $336$ & $56.7$ \\
 & Planner-only & $136$ & $39.3$ \\
 & \SYSNAME{} & $164$ & $32.5$ \\
 & Raw & $179$ & $36.6$ \\
 & \abl{S} & $150$ & $30.1$ \\
 & \abl{R} & $172$ & $32.3$ \\
 & \abl{M} & $183$ & $37.9$ \\
 & \abl{E} & $175$ & $41.8$ \\
\midrule
GLM-5.3 & Planner-only & $87$ & $27.3$ \\
 & \SYSNAME{} & $136$ & $31.4$ \\
\midrule
hy4-preview & Planner-only & $160$ & $46.9$ \\
 & \SYSNAME{} & $219$ & $40.3$ \\
\midrule
Kimi-K3 & Planner-only & $115$ & $42.2$ \\
 & \SYSNAME{} & $127$ & $35.8$ \\
\bottomrule
\end{tabular}
\end{table}

On the default backbone \SYSNAME{} makes $21\%$ more LLM calls than the
planner-only control ($164$ vs.\ $136$ per night), while executing slightly
fewer environment steps per agent
($32$ vs.\ $39$). The two external baselines make more calls than \SYSNAME{}
($336$ and $439$) and take more steps per agent ($57$ and $67$). Reflection
and aggregation account for $53$ of \SYSNAME{}'s $164$ calls per night
($32\%$); on the other backbones the extra calls over the planner-only
control range from $12$ to $59$ per night. All cells are counted in the
same way.

\begin{table}[htbp]
\centering
\caption{Inference cost on DS-V4.1F (ten runs per condition, three agents): calls and
saved text volume per game night by component, and mean wall-clock hours of a complete
30-night run. Text volume is saved message characters divided by $1.7$, an estimate of
tokens; each message is counted once, so it is not billed tokens, which also include
re-sent history. The runs carry no token log.}
\label{tab:app_text}
\small
\setlength{\tabcolsep}{6pt}
\begin{tabular}{lcccc}
\toprule
 & \multicolumn{2}{c}{Planner-only} & \multicolumn{2}{c}{\SYSNAME{}} \\
\cmidrule(lr){2-3}\cmidrule(lr){4-5}
Component & Calls & Text (k) & Calls & Text (k) \\
\midrule
Planning and execution & $136.0$ & $98$ & $110.1$ & $155$ \\
Reflection & --- & --- & $46.4$ & $221$ \\
Aggregation & --- & --- & $7.5$ & $38$ \\
\midrule
Total & $136$ & $98$ & $164$ & $414$ \\
\midrule
Wall-clock hours per run & \multicolumn{2}{c}{$7.6$} & \multicolumn{2}{c}{$12.1$} \\
\bottomrule
\end{tabular}
\end{table}

By this measure \SYSNAME{} processes about $4.2$ times as much text per
night as the planner-only control (per-run range $389$--$449$k vs.\
$91$--$107$k): $62\%$ of it is reflection and aggregation, which read each
agent's day, and planning itself carries $1.6$ times the text because
retrieved knowledge and the morning injection enter the prompts, even
though it makes fewer calls. The extra cost of the knowledge components is
therefore larger in text than in calls, and we make no claim about
computational efficiency; the results establish progress and survival
within a fixed game horizon.

Full and \abl{R} are not at equal retrieval budget (ten runs and all
three agents per condition, mean over the 30 days; ranges are across runs).
The user prompt that opens a day has $13{,}295$ characters under Full
($11{,}416$--$15{,}722$) and $14{,}800$ under \abl{R} ($12{,}495$--$17{,}435$).
It carries the agent's memory layer, injected each morning and untouched
by \abl{R}, so its length reflects how much the agents of each run wrote,
not the retrieval mechanism. The block of experience
and skills returned when the agent sets its plan is longer under Full
($14{,}788$ characters, $13{,}284$--$16{,}124$) than under \abl{R}
($11{,}901$, $10{,}795$--$14{,}568$), about $24\%$ more retrieved text at that
point. Over a whole run Full processes about $6\%$ more text than \abl{R}
($414$k vs.\ $390$k per night) with a similar number of LLM calls ($164$ vs.\
$172$), so the difference in retrieval is real but small next to the text
spent on reflection and aggregation, which both conditions keep.
The other ablations, which remove memory, experience or sharing, change
what is written and read, so their costs in Table~\ref{tab:app_cost}
describe different amounts of work, not the same work done more cheaply.

\end{document}